\documentclass{article}

\usepackage{arxiv}

\usepackage[utf8]{inputenc} 
\usepackage[T1]{fontenc}    
\usepackage{hyperref}       
\usepackage{url}            
\usepackage{booktabs}       
\usepackage{amsfonts}       
\usepackage{nicefrac}       
\usepackage{microtype}      

\usepackage{footmisc}
\usepackage{tablefootnote}
\usepackage{todonotes}
\usepackage{scrextend}
\usepackage{changepage} 
\usepackage{subcaption}
\newcounter{CQscounter}
\newcommand{\incrementCQscounter}{\protect\stepcounter{CQscounter}\theCQscounter}
\usepackage{csquotes}
\usepackage{cleveref}
\usepackage{supertabular}
\usepackage{listings}
\newcommand{\smalltt}[1]{{\small\texttt{#1}}}

\title{Pattern-based design applied to cultural heritage knowledge graphs}

\author{Valentina Anita Carriero\\
Department of Computer Science and Engineering \\
University of Bologna \\
Mura Anteo Zamboni 7, 40126 Bologna, Italy \\ 
\texttt{valentina.carriero3@unibo.it} \And
Aldo Gangemi\\
Digital Humanities Advanced Research Centre \\
Department of Classical Philology and Italian Studies \\
University of Bologna \\
Via Zamboni 32, 40126 Bologna, Italy \\
\texttt{aldo.gangemi@unibo.it} \And
Maria Letizia Mancinelli \\
Central Institute for Cataloguing and Documentation \\
Ministry of Cultural Heritage and Activities \\
Via di San Michele 18, 00153 Roma \\
\texttt{marialetizia.mancinelli@beniculturali.it} \And 
Andrea Giovanni Nuzzolese \\
Semantic Technologies Laboratory \\
Institute of Cognitive Sciences and Technologies \\
Italian National Research Council \\
Via San Martino della Battaglia 44, 00185 Rome, Italy \\ 
\texttt{andreagiovanni.nuzzolese@.cnr.it} \And 
Valentina Presutti \\
Department of Modern Languages, Literatures, and Culture \\
University of Bologna \\
Via Cartoleria 5, 40124 Bologna, Italy \\ 
\texttt{valentina.presutti@unibo.it} \And 
Chiara Veninata \\
Central Institute for Cataloguing and Documentation \\
Ministry of Cultural Heritage and Activities \\
Via di San Michele 18, 00153 Roma \\
\texttt{chiara.veninata@beniculturali.it}
}

\begin{document}
\maketitle

\setcounter{CQscounter}{1}


\begin{abstract}
Ontology Design Patterns (ODPs) have become an established and recognised practice for guaranteeing good quality ontology engineering. There are several ODP repositories where ODPs are shared as well as ontology design methodologies recommending their reuse. Performing rigorous testing is recommended as well for supporting ontology maintenance and validating the resulting resource against its motivating requirements. Nevertheless, it is less than straightforward to find guidelines on \emph{how} to apply such methodologies for developing domain-specific knowledge graphs. ArCo is the knowledge graph of Italian Cultural Heritage and has been developed by using eXtreme Design (XD), an ODP- and test-driven methodology. During its development, XD has been adapted to the need of the CH domain e.g. gathering requirements from an open, diverse community of consumers, a new ODP has been defined and many have been specialised to address specific CH requirements. This paper presents ArCo and describes \emph{how} to apply XD to the development and validation of a CH knowledge graph, also detailing the (intellectual) process implemented for matching the encountered modelling problems to ODPs. Relevant contributions also include a novel web tool for supporting unit-testing of knowledge graphs, a rigorous evaluation of ArCo, and a discussion of methodological lessons learned during ArCo development.  
\end{abstract}




\section{Introduction}
\label{sec:intro}
Museums, libraries, archives, private collections and other cultural institutions have the essential mission to preserve the cultural objects they collect. Hence, data about these objects is of utmost importance, since it allows to keep memory of them, their life cycle as well as their artistic, social, and historical context. If data are shared, they can be used as a means of enhancing cultural properties, by spreading knowledge on cultural heritage, and widening its potential consumers. Cultural Heritage (CH) data can have various types of consumers such as citizens, students, scholars, scientists, managers, public administrations and companies. Consequently, it can impact on different domains such as tourism, research, management, teaching, etc. Moreover, cultural institutions and research organisations can mutually benefit from the data they publish, especially by creating connections between their knowledge bases. The Linked Data paradigm has shown its effectiveness in supporting this practice~\cite{Bizer2009}, and its adoption in the Cultural Heritage domain is leading to a significant transformation in the management of CH data~\cite{DBLP:journals/semweb/DijkshoornJAOSW18,DBLP:journals/semweb/BoerWGOHIOS13,DBLP:series/ihis/Hyvonen09,DBLP:journals/semweb/IsaacH13,DBLP:journals/jocch/DaquinoMPTV17}. 

The Italian Cultural Heritage is an important part of the world's CH\footnote{According to UNESCO, Italy is the country with the highest heritage sites in the world \cite{Carriero2019}.}, and a great resource for Italy from aesthetic, social, historical, cognitive and economic points of view. More and more cultural institutions are publishing their data, often as open data, in order to allow for interchange, interlinking and mutual enrichment.

In \cite{Carriero2019} we introduce ArCo\footnote{Architecture of Knowledge, from Italian \emph{Architettura della Conoscenza}.}, a resource that contributes to this vision by publishing a knowledge graph (KG), consisting of a network of ontologies that model the CH domain and a Linked Open Data (LOD) dataset of $\sim$172.5M triples about Italian cultural properties, along with documentation and software artefacts. 

ArCo KG (composed of ArCo ontology network and LOD data) is available at the MiBAC's official SPARQL endpoint\footnote{\url{http://dati.beniculturali.it/sparql}}. The endpoint is based on the Open Source version of Virtuoso\footnote{https://github.com/openlink/virtuoso-opensource}, which is used by MiBAC for its liked data projects. ArCo KG is also released as part of a package, which consists of a \emph{docker} container available on GitHub\footnote{\url{https://github.com/ICCD-MiBACT/ArCo}} - allowing you to have everything on your own PC - and its running instance online\footnote{\url{https://w3id.org/arco}} - both English and Italian versions. This package includes: documentation, user guides and diagrams; the source code and a human-readable HTML documentation of the ontologies\footnote{Created with LODE: \url{http://www.essepuntato.it/lode}}; a SPARQL endpoint; examples of Competency Questions and their corresponding SPARQL queries; RDFizer\footnote{\url{https://github.com/ICCD-MiBACT/ArCo/tree/master/ArCo-release/rdfizer}}, a software for converting XML data represented according to ICCD cataloguing standards\footnote{\label{ref:normative}\url{http://www.iccd.beniculturali.it/it/normative}} to RDF compliant to ArCo ontologies. The docker release of ArCo can be extended and customised in order to use alternative triplestores or graph databases.


Besides the relevance of the produced resource, described in \cite{Carriero2019}, ArCo's project contributes to push the state of the art in knowledge graph engineering, with special focus on the CH domain, by sharing its ``behind the scenes'', i.e. the intellectual and methodological processes performed, the adopted design principles and the lessons learned, all of which constitute the main focus of this paper. ArCo KG development follows  
a pattern-based ontology design methodology named eXtreme Design (XD)~\cite{Blomqvist2010,Blomqvist2016}, and has contributed to extend and improve it, as discussed in Section \ref{sec:method}. 

ArCo KG is an evolving creature, so is the methodology it follows i.e. XD. New requirements are continuously collected, incremental versions are regularly released, and its methodological approach is discussed with the community, and possibly refined and evolved\footnote{ArCo's implementation of XD is discussed on a dedicated mailing list \url{arco-project@googlegroups.com} as well as during webinars and meetups.}.
The current version of ArCo KG mainly derives from the General Catalogue of Italian Cultural Heritage\footnote{\label{ref:catalogo}\url{http://www.catalogo.beniculturali.it}} (GC), which is maintained by the Central Institute for Catalogue and Documentation (ICCD) of the Ministry of Cultural Heritage and Activities\footnote{\label{ref:iccd}\url{http://www.beniculturali.it}} (MiBAC). ICCD coordinates the cataloguing activities by collecting and integrating data coming from diverse institutions all over Italy with the help of a collaborative platform named SIGECweb\footnote{\label{ref:sigec}\url{http://www.iccd.beniculturali.it/it/sigec-web}}. The General Catalogue data are finally stored in a relational database (and encoded in XML). In order to convert such data into a knowledge graph, the conceptual model behind such database must be formalised in a reference ontology.

There are several, valuable existing models for representing Cultural Heritage data and publishing them as LOD. The Europeana Data Model (EDM)~\cite{DBLP:journals/semweb/IsaacH13} and CIDOC Conceptual Reference Model (CRM)~\cite{DBLP:journals/aim/Doerr03} are two prominent examples. EDM defines a basic set of classes and properties for describing cultural objects, which are used to aggregate CH data into the Europeana portal\footnote{\label{ref:europeana}\url{https://www.europeana.eu/en}}. CIDOC CRM is an international standard for representing the CH domain, supporting the exchange of information between museums, archives and libraries.
Both models focus on supporting a linked data encoding of \emph{metadata} that can be extracted from catalogue records. They successfully support two main use cases: feeding cultural heritage data aggregators such as Europeana, and enabling data interchange between cultural institutions. 

As compared to EDM and CIDOC CRM, ArCo KG aims at modelling the Cultural Heritage universe of discourse with a much finer grain and by addressing a wider variety of concepts, ranging from cultural properties' metadata (e.g. authors, creation date, current location, style) to research findings and theories (e.g. scientific processes performed for analysing a cultural property, theories and foundations about possible former settlements in an archaeological site). 

By formalising the semantics of cultural properties, the events they participate in, the types of places they are located in, the processes they are involved in, etc. ArCo KG provides the CH and the Semantic Web communities with a set of ontology patterns to encode CH knowledge graphs. The ultimate goal is to enable researchers and scholars to make new findings about cultural entities, and to develop new theories based on observations performed on knowledge graphs modelled by means of ArCo ontologies. 

ArCo ontologies are aligned to EDM and CIDOC CRM, in order to facilitate linking and reuse by aggregators. Alignment and differences between ArCo ontologies, EDM and CIDOC CRM are discussed in detail in Section \ref{sec:situation} and in Section \ref{sec:cidoc-edm}. Nevertheless ArCo ontologies take a different foundational commitment than CIDOC CRM and EDM. The foundations of ArCo KG are: (i) the theory of Constructive Descriptions and Situations (cDnS) \cite{DBLP:journals/aamas/Gangemi08} and (ii) the reflection of an epistemological perspective on cultural properties. These are discussed in detail in Section \ref{sec:situation} and Section \ref{sec:odps}. Informally, ArCo KG is \emph{situation-centric}, meaning that all facts in its universe of discourse are modelled as \emph{situations}: occurrences of relational contexts involving objects, that can be temporally and spatially indexed (e.g. where a cultural property is located, why and when; which author is attributed to a cultural property, based on which criteria, and when). The epistemological stance substantiates in distinguishing facts (e.g. the creation date of a cultural property; the author of a cultural property) from interpretations (e.g. dating estimation, authorship attribution) about a cultural property as well as from entities that document them (e.g. catalogue records) along with their evolving content (e.g. versions). 

\paragraph{Contribution}
This paper extends \cite{Carriero2019} by providing an in-depth analysis of ArCo KG (including examples) and its development context. Novel contributions can be summarised as follows:
\begin{itemize}
\item an extension of the eXtreme Design methodology for dealing with Cultural Heritage ontology projects (or for knowledge domains with characteristics similar to CH)
\item an architectural ontology pattern for implementing large ontology networks
\item a detailed explanation of the foundations of Arco ontologies
\item a detailed description of the main modelling issues addressed by ArCo and the related implemented ODPs    
\item a formal evaluation of ArCo ontologies based  both established structural metrics~\cite{Tartir2010,Yao2005,Gangemi2006,Orme2006,Schlicht06,dAquin09,Khan16}, and XD-based unit testing
\item a tool (TESTaLOD) for supporting XD-based regression tests
\item a thorough description of the experience in applying XD to the development of ArCo KG

\end{itemize}

After Section \ref{sec:catalogue}, which describes the General Catalogue of Italian Cultural Heritage, 
Section \ref{sec:method} describes the eXtreme Design methodology and discusses how we applied and extended it in the context of the ArCo project. Section \ref{sec:situation} provide details about the foundations of ArCo ontologies and poses the basis to discuss, in Section \ref{sec:odps}, the main modelling issues addressed in ArCo ontologies and how they have been matched to existing Ontology Design Patterns.
Section \ref{sec:evaluation} evaluated ArCo knowledge graph. Section \ref{sec:related} discusses relevant related work and Section \ref{sec:lessons} summarises the lessons learned from the experience of developing ArCo, so far. Finally, Section \ref{sec:conclusion} wraps up the paper and points out some ongoing and future work.

\section{The journey through semi-structured data on Italian Cultural Heritage}
\label{sec:catalogue}
Building a knowledge graph and its reference ontology network requires to understand the domain and the ontological commitment that its conceptualisation conveys, and to transform the available data into linked entities that comply with the resulting ontologies. There may be different scenarios in terms of what is available at the beginning of a knowledge graph project, but one of the most common situations is having a (set of) database(s) where the data are stored and maintained. Along with a continuous interaction with the administrators of the databases and the domain experts, these resources are to be analysed in order to extract the (often implicit) conceptual model of the domain that they encode. ArCo KG main datasources have been an XML database of catalogue records and a set of pdf documents describing catalogue standards. 

Cataloguing cultural heritage is the process of identifying and describing, through metadata, entities that are considered cultural properties, by virtue of their historic, artistic, archaeological and ethnoanthropological interest. In Italy, the Italian Ministry of Cultural Heritage and Activities\footref{ref:iccd} (MiBAC), regions and local agencies are in charge of cooperatively cataloguing Italian cultural heritage they own, aiming at safeguarding, enhancing and making publicly available data on cultural heritage. 

\subsection{The General Catalogue of Italian Cultural Heritage}
ICCD coordinates these cataloguing activities by maintaining the General Catalogue of Italian Cultural Heritage\footref{ref:catalogo} (GC), which is the official institutional database of Italian cultural heritage, promoting integrated management of data coming from all over Italy and from diverse institutions and local contexts. 

\begin{figure}[!ht]
\centering
	\includegraphics[scale=0.18]{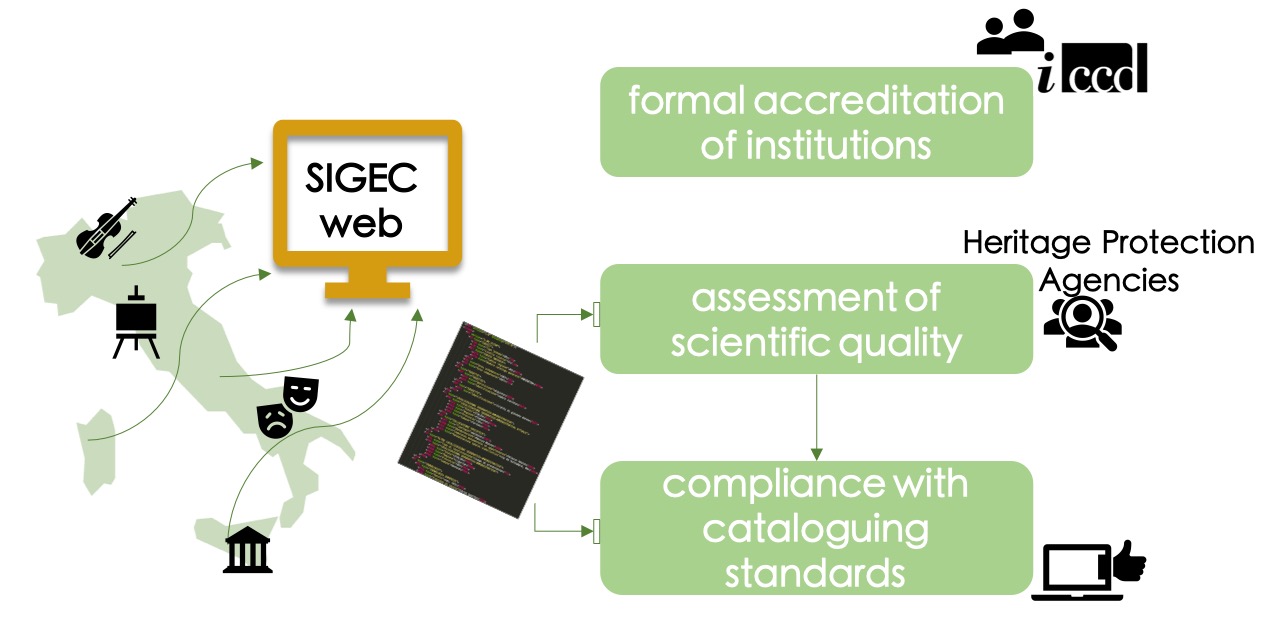}
	\caption{Accreditation and validation process for contributing to SIGECweb.}
	\label{img:sigec}
\end{figure}

The General Catalogue is built upon a collaborative platform, named SIGECweb\footref{ref:sigec}, to which national or regional, public or private, institutional organisations that administer cultural properties can submit their catalogue records, i.e. files containing data on cultural properties and compliant with predetermined standards and guidelines (see Subsection \ref{par:cataloguing-standards}). Only users from institutions that are formally authorised by ICCD can access and contribute to SIGECweb, with specific profiles (e.g. administrator, cataloguer). 
The quality of the database and its highly reliable provenance is guaranteed by (i) an accreditation process that allows only authorised entities to contribute to the platform, (ii) a data validation phase performed by heritage protection agencies that assess the scientific quality of catalogue records, and (iii) an automatic data validation phase based on compliance with specific cataloguing standards (see Figure \ref{img:sigec}).

SIGECweb currently contains 2,735,343 catalogue records, 831,114 of which are publicly accessible through the General Catalogue. The privacy level associated with the remaining records prevents them to be openly published, since they refer to properties either private, or being at stake (e.g. items in unguarded buildings), or still requiring a scientific assessment by accounted institutions.

\subsection{Italian Cataloguing Standards}
\label{par:cataloguing-standards}
In order to guarantee high quality, consistency and interoperability between data accessible through the General Catalogue, ICCD defines a set of standards (\emph{normative})\footref{ref:normative} for encoding catalogue records (\emph{schede di catalogo}), which provides a template for collecting and organising data on different types of cultural properties and a methodological base for cataloguing.
Thus, these standards are part of ArCo KG input, along with data contained in the GC catalogue records.

\begin{figure*}[!ht]
\centering
	\includegraphics[width=0.7\textwidth]{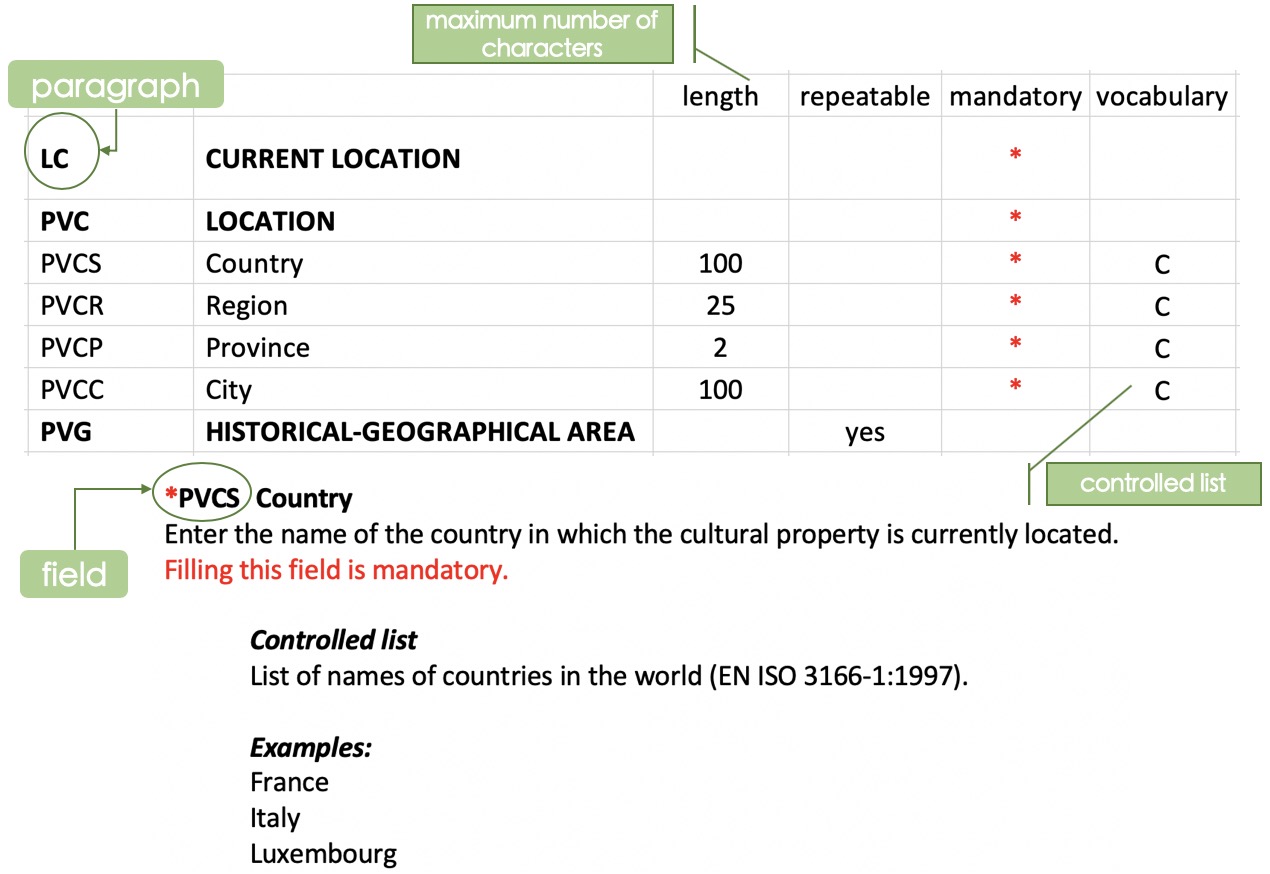}
	\caption{An example of the structure of an ICCD cataloguing standard.}
	\label{img:cataloguing-standard}
\end{figure*}

Each cataloguing standard consists of a PDF document\footnote{These standards are gradually being published also in XML Schema Definition (XSD) format.} that contains, as shown in Figure \ref{img:cataloguing-standard}: a table listing all fields 
for collecting data about a cultural property, and the respective rules for compilation. These fields are grouped into \emph{paragraphs} with regard to the topic (e.g. geographical information), following a hierarchical structure. To each paragraph and field in the hierarchy are associated a tag and generic instructions: maximum number of characters allowed, whether that field is repeatable,
whether its compilation is mandatory,
etc. Moreover, all fields are accompanied by rules for compilation, such as syntactic rules (e.g. the date format), and useful examples.

\subsubsection{30 types of cultural property, 30 cataloguing standards}
ICCD collects catalogue records about 9 categories of cultural properties, which generalise over 30 different more specific types: archaeological, architectural and landscape, demo-ethno-anthropological, photographic, musical, natural, numismatic, scientific and technological, historical and artistic properties. For each of the 30 typologies, a specific cataloguing standard has been defined, while a \emph{cross} cataloguing standard (\emph{Normativa Trasversale}) groups and defines the field common to all kinds of cultural property.
Nevertheless, the particular features of each cultural property type require specific standards for defining additional paragraphs and fields, e.g. a paragraph for describing possible accessories, such as the instrument case, of a musical instrument. 
Moreover, some fields are associated with controlled lists, i.e. lists of non-overlapping terms used to control terminology.
In many cases, these controlled lists differ, partially or completely, depending on the cultural property that is being catalogued: for example, according to the standard for photographs, the list associated with the field \emph{type of measurement} contains values such as \enquote{height x length} or \enquote{height x length x thickness}, while, when cataloguing technological heritage, examples of valid values are \enquote{weight} and \enquote{volume}.

Currently, an effort is being made by ICCD in publishing on GitHub\footnote{\url{https://github.com/ICCD-MiBACT/Standard-catalografici/tree/master/strumenti-terminologici}} many of these controlled lists in RDF using SKOS.

\subsubsection{One cataloguing standard, different versions over time} 
\begin{figure}[!ht]
\centering
	\includegraphics[scale=0.18]{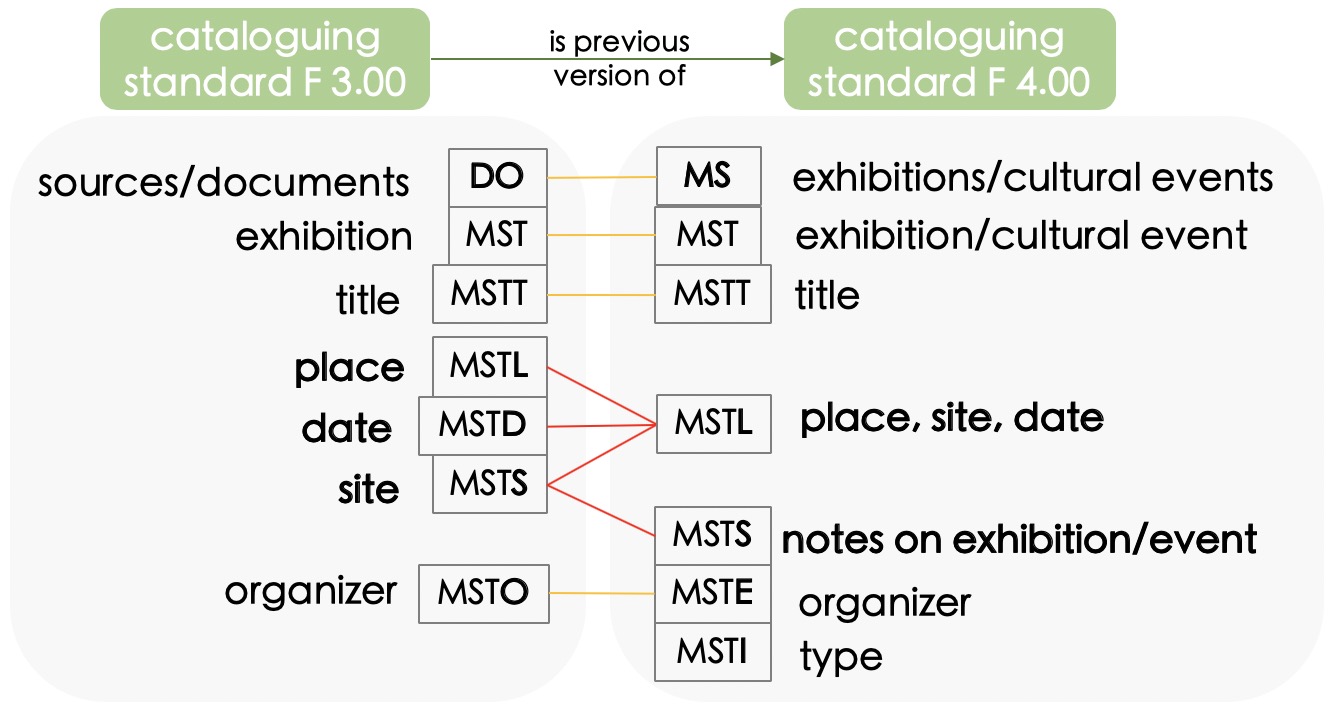}
	\caption{The ICCD cataloguing standard \emph{F} for photographs: difference and mapping between version 3.00 and version 4.00.}
	\label{img:standard-versions}
\end{figure}

ICCD has been sharing cataloguing standards since 1990: they have undergone changes and updates, regarding both their structure and rules for compilation\footnote{Previous and current versions include: 1.00 and 2.00 (1990-2000), 3.00 (2002-2004), 3.01 (2005-2010), 4.00 (since 2015).}. 
As a result, the GC contains heterogeneous catalogue records, following different versions, thus requiring expensive and time-consuming mapping activities: indeed, in moving from a version to the next one, ICCD did not systematically keep track of changes.
While in some cases this mapping is straightforward, in other cases differences over data due to different versions can be significant. 
Let us consider an example, depicted in Figure \ref{img:standard-versions}, of the standard F (for cataloguing photographs). In version 3.00 there are three separate fields to indicate \emph{place, site} and \emph{date} of a photograph (\texttt{MSTL}, \texttt{MSTD}, \texttt{MSTS}, respectively), while in version 4.00, they are all encoded in a single field (\texttt{MSTL}).

\subsection{A closer look at actual catalogue records}

Although catalogue records submitted to SIGECweb are subject to a validation process, being collaborative in nature means that catalogue records are not error-free.
There are cases of: mandatory fields that are not properly filled, thus producing an error code; 
catalogue records containing values alternative to those provided by controlled lists, hence undermining data homogeneity; use of non-standard formats (e.g. for dates); minor bugs and typing errors.
ICCD is continuously working for improving the collecting process, in order to minimise these situations.

Moreover, catalogue standards themselves could be improved in their structure, in order to maximize data mining from catalogue records: there are still many fields allowing for long descriptive texts, from which structured high-quality information could be derived and extracted. 
%
\section{Applying eXtreme Design principles to model the Cultural Heritage domain}
\label{sec:method}
In order to develop ArCo ontologies, which cope with a huge and complex domain such as Cultural Heritage's, we use ontology design patterns (ODPs) \cite{DBLP:series/ihis/GangemiP09,DBLP:books/ios/HGJKP2016}. Ontology patterns provide solutions to recurrent modelling issues. Their adoption guarantees a high level of the overall ontology quality, and favour its re-usability \cite{Blomqvist2009Experiments}.

The use of design patterns in ontology engineering is less evident than in software engineering. Software design patterns are such a standard practice that many programming languages have built-in types implementing, or inspired by, them e.g. Observer in Java and Iterable both in Java and in Python. ODPs instead, although recognised as good practices in general, are yet distant to achieve a clear standard reference in the knowledge engineering community, and very far to be common practice in the Linked Data community. Their introduction in the Semantic Web is relatively recent~\cite{DBLP:conf/semweb/Gangemi05} and to date, there is still lack of tooling to ease their adoption. A very recent and promising contribution to fill this gap is CoModIDE~\cite{DBLP:conf/esws/ShimizuHH20}, a Prot\'eg\'e\footnote{\url{https://protege.stanford.edu/}} plugin supporting pattern-based design\footnote{At the time of ArCo development the tool was not available, we plan to test and use it in future developments.}. The design of this tool is inspired by the combination of two paradigms: modular ontology modelling~\cite{DBLP:journals/corr/abs-1808-08433} and eXtreme Design (XD)~\cite{Blomqvist2010,Blomqvist2016}. We follow (and extend) the XD methodology for the design of ArCo ontologies.

\subsection{eXtreme Design methodology}
\label{sec:extreme-design}
XD is an ontology design methodology that puts the reuse of ODPs at its core both as a principle and as an explicit activity. It provides guidelines for such activity. Experiments have proved its positive impact on ontology engineering and ontology quality~\cite{Blomqvist2010,DBLP:conf/esws/ShimizuHH20}. 

XD is partly inspired by eXtreme Programming (XP)~\cite{ShoreWarden07}, an agile software development methodology that aims at minimizing the impact of changes at any stage of the development, and producing incremental releases based on customer requirements and their prioritization. Although the two approaches have similarities, they diverge towards different focuses mainly due to the core differences between software systems and knowledge bases. Where XP diminishes the value of careful design, this is exactly where XD has its main focus. XD is test-driven, and applies the divide-and-conquer approach as well as XP does. Also, XD adopts pair design (as opposed to pair programming). The intensive use of ODPs, modular design, and collaborative approach are the main characterising principles of the method.
Further details on the relation between XP and XD, and a thorough description of XD are given in~\cite{DBLP:conf/semweb/PresuttiDGB09}.

\begin{figure*}[!ht]
\centering
	\includegraphics[width=0.7\textwidth]{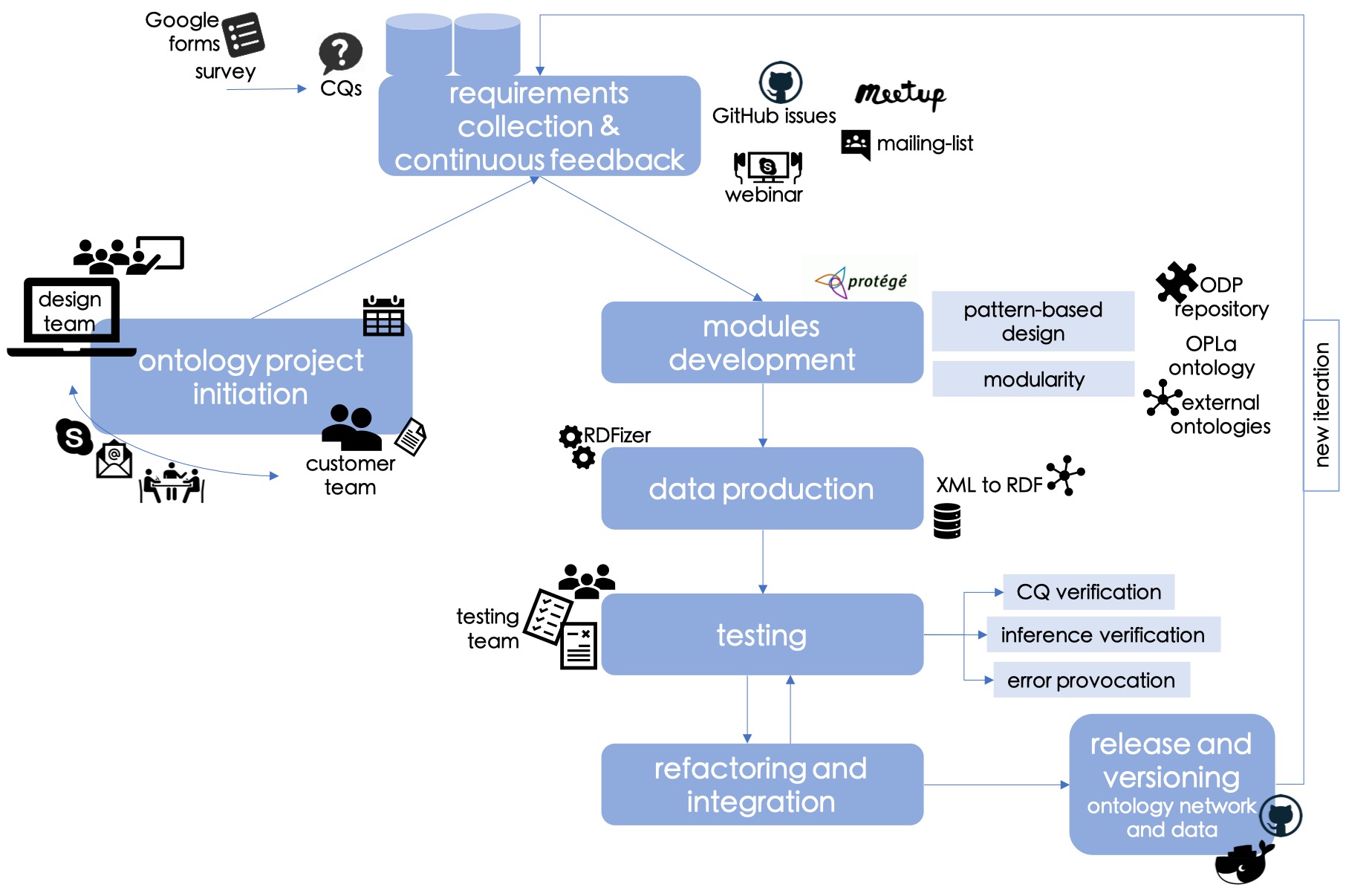}
	\caption{The XD methodology as implemented for the ArCo knowledge graph.}
	\label{img:methodology}
\end{figure*}

\begin{figure}[!ht]
\centering
	\includegraphics[scale=0.19]{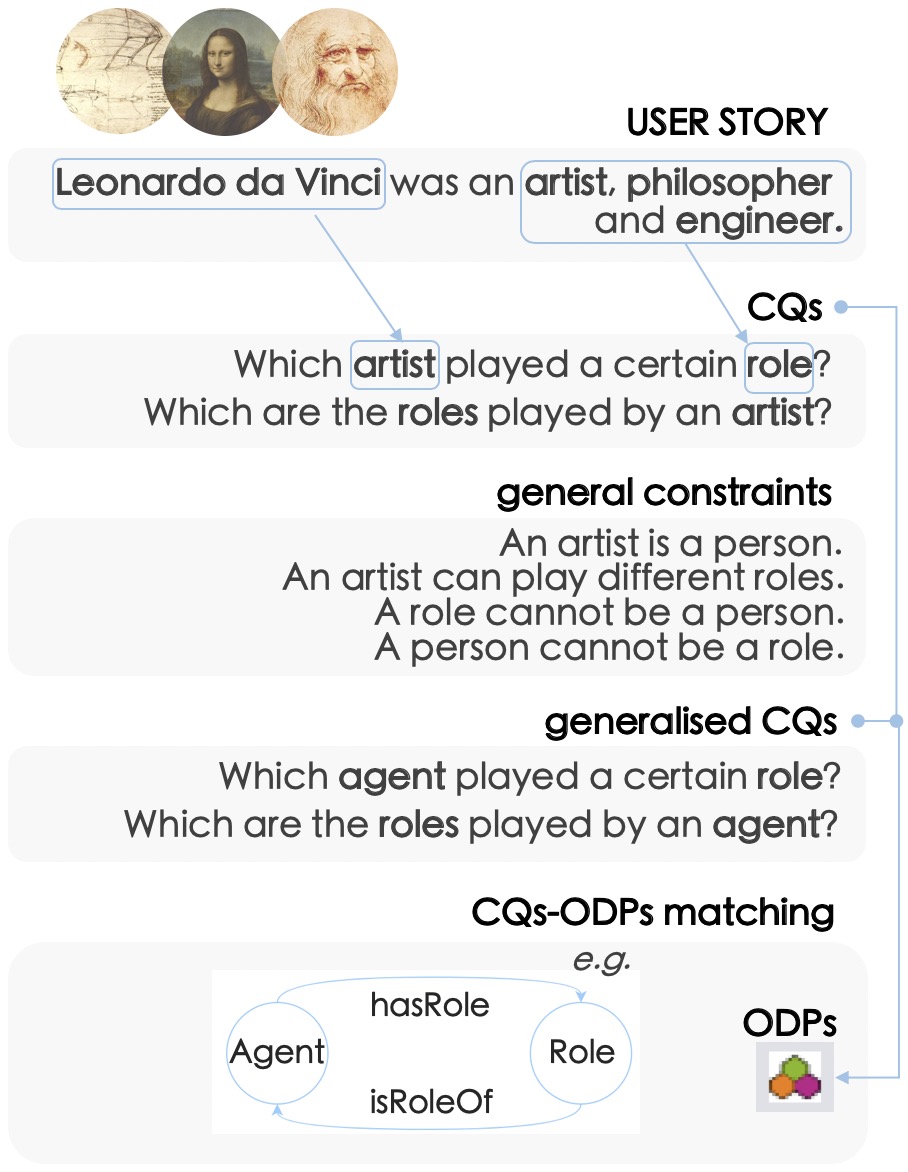}
	\caption{An example of a user story translated into CQs, and of the matching between a CQ and an ODP.}
	\label{img:user-story}
\end{figure}

As depicted in Figure \ref{img:methodology}, after the project initiation, XD is executed by iterating a set of steps, each involving one or more \emph{teams}: a \emph{customer team}, which elicits the requirements that guide the design and testing process; a \emph{design team}, which is in charge of identifying and implementing the ODPs that best address the given requirements; a \emph{testing team}, which performs testing and validation of the produced ontology components; an \emph{integration team}, which takes care of integrating the different components. In XD, the design team works in parallel and interactively with the testing team: the same requirements are used as input by the design team, for producing the ontology, and by the testing team, for translating them into unit tests.

Figure \ref{img:user-story} depicts a simple example that will be used to illustrate the main steps of the methodology.

\paragraph{Requirements engineering.} 
A fundamental step is to collect requirements and to engineer them. 
Requirements are collected in the form of \emph{user stories}, which are provided by the customer team. A user story is a set of sentences, which describe by example the kind of facts that the resulting knowledge graph is required to encode. The length of a user story is limited to favour keeping them focused; its maximum length is decided by the design team. The customer team is instructed to break possible complex stories into smaller and simpler ones. An example of simple user story is (cf. Figure \ref{img:user-story}): ``Leonardo da Vinci was an artist, a philosopher and an engineer''. 
User stories may be associated with a priority level, a title, and an ID\footnote{Each ontology project will define its own conventions.}. The title or ID can be used to express possible dependencies of a story on others, meaning that it cannot be analysed until the other stories have been treated. The priority level influences the order in which stories are treated. Dependencies and priorities are assigned by the design team considering input from the customer team. 

\paragraph{Competency Questions.} One or more competency questions (CQs)~\cite{GruningerFox94} are derived from a generalisation of the user stories. CQs are the natural language counterpart of structured queries that we want to enable against the resulting knowledge graph. Generalising a user story means to identify the main concepts that they exemplify. This generalisation is carried out by the design team, in collaboration with the customer team. For example, considering the previous user story, two CQs may be derived from it: ``Which are the roles played by an artist?'' and ``Which artists played a certain role?'' (cf. Figure \ref{img:user-story}). In this case, the design and customer teams have identified that "Leonardo da Vinci" is an example of \emph{Artist}, which is a relevant concept in the domain, and that "artist", "philosopher", and "engineer" are examples of \emph{Roles} that an artist can play, another relevant concept to include in the ontology. 

In addition to deriving CQs, the design team interacts with the customer team in order to identify possible general constraints that may accompany them. General constraints express possible inferences or other rules that apply to the concepts that the story involves. In the example of Figure \ref{img:user-story}, the following general constraints can be drawn: ``An artist is a person'' and ``A person cannot be a role''. General constraints are the natural language counterpart of axioms that will be formalised in the ontology. The CQs and the general constraints define the ontology requirements, hence contributing to assess the ontological commitment, as far as the ontology domain tasks and scope are concerned. From the XD perspective, the ontological commitment includes both meta-level aspects such as adopting a 3d or 4d view, as well as functional/application aspects which draw the boundaries of the scope of the ontology. ArCo ontological foundations are discussed in Section \ref{sec:situation}.
 
 
 
 

\paragraph{Matching CQs to ODPs.} CQs guide the selection of ODP: a key process in XD, and in general in pattern-based design, is to \emph{match} CQs to ODPs. At each iteration, a coherent set of CQs is selected, i.e. CQs dealing with same modelling issues (e.g. roles played by agents). Possible existing solutions (ODPs) are analysed in order to find the most suitable one to be implemented in the ontology. This is a complex cognitive task that, currently, lack proper tool support. Hence, it is better performed by whom has previous knowledge about existing ODPs. Nevertheless, if one is not familiar with ODPs, they can be found on catalogues, such as the catalogue maintained by the University of Manchester\footnote{\url{http://www.gong.manchester.ac.uk/odp/html/index.html}} and the ODP portal\footnote{\url{http://www.ontologydesignpatterns.org}\label{fn:odp:portal}}, as well as in reference literature, such as the Workshop on Ontology Design and Patterns series\footnote{\url{http://ontologydesignpatterns.org/wiki/WOP:Main}} and~\cite{DBLP:books/ios/HGJKP2016}. When an ODP is properly defined and documented, it thoroughly describes the modelling issue that it addresses, by providing its related competency questions. A designer is able to assess whether an ODP matches the CQs that she has at hand by comparing them with the ODP's CQs. Often, ODP's CQs are more general than the domain-specific CQs of an ontology project. In such case, the designer will generalise her CQs further, to understand whether the candidate ODP can be reused, given a specialisation of its vocabulary.
For example the CQ in Figure \ref{img:user-story}: ``Which artist plays a certain role?'' can be generalised into ``Which agent plays a certain role?''. If we consider this generalised version of the CQ, we can easily match it to the \emph{AgentRole}\footnote{\url{http://www.ontologydesignpatterns.org/cp/owl/agentrole.owl}} ODP, available and documented on the ODP portal\footref{fn:odp:portal}. 
We describe this process, with real examples from the ArCo project in Section \ref{sec:odps}.

\paragraph{Testing and integration.} XD is test-driven and follows a unit testing approach as described in~\cite{DBLP:conf/semweb/PresuttiDGB09,Blomqvist2012}. The CQs and general constraints defined by the design team are shared with the testing team. While the design team uses them for producing a piece of the ontology, the testing team uses them for designing unit tests: CQs are translated into possible SPARQL queries, while general constraints are used to create sample triples, based on the user stories, that are expected to provoke either consistency/coherence errors or inferences. The testing team will use a draft terminology based on the CQs. In other words, unit tests are sample OWL/RDF files (encoding the user story) and SPARQL queries, annotated with their corresponding expected results. When the testing team receives an ontology piece from the design team, it firstly attempts to replace the corresponding unit test (draft) terminology with the provided ontology vocabulary. At this stage possible missing concepts can be spotted, which would make the test positive\footnote{A positive test means that some error has  raised.} and cause a feedback to the design team asking them to fix and resubmit. Once the terminological coverage is successful, the testing team proceeds with (i) verifying whether the queries produce the expected results and (ii) checking consistency and coherence of the ontology piece.
%
With reference to the example of Figure \ref{img:user-story}, the testing team will design and check the result of SPARQL queries aimed at retrieving the roles of an artist, against a sample OWL/RDF file that encodes triples about Leonardo da Vinci being an artist, a scientist and an engineer. Furthermore, they will check the ontology model e.g. as for the disjointness between the classes \smalltt{:Artist} and \smalltt{:Role}: this can be done by creating a sample OWL/RDF files with the following statements: 

\smalltt{:LeonardoDaVinci rdf:type :Artist .} \\
\smalltt{:LeonardoDaVinci rdf:type :Role .}

Integration tests follow a similar process, but focus on running regression tests\footnote{All unit tests passed so far.} and possible additional unit tests, on the whole ontology, after integrating the new piece. Integration tests may be performed by a dedicated team or by the testing team, depending on the size of the project and available resources.
The results of the tests are reported to the design team, which will fix and resubmit in case some of the tests are positive. 

\smallskip
So far, we have described eXtreme Design (XD) according to~\cite{DBLP:conf/semweb/PresuttiDGB09,Blomqvist2010,Blomqvist2016}. In the context of the ArCo's project we faced a number of challenges that are not explicitly tackled by XD, hence we contributed to extend it. Specifically, one challenge concerns how to design the architecture of an ontology network. As we deal with a wide and complex domain such as Cultural Heritage, we would have benefit from clear guidelines. There is some preliminary work described in~\cite{DBLP:series/ihis/GangemiP09} about architectural patterns for ontologies, defined as patterns that \begin{quote}
[...] affect the overall shape of the ontology, and dictate `how the ontology should look like'. [...] An ontology that has a simple modular architecture is composed of a set of ontologies, called modules, plus one ontology that imports all the modules.
\end{quote} To the best of our knowledge there are no catalogues nor empirical studies reported about these types of patterns.

Another challenge concerns the collection of requirements. Cultural heritage data are relevant for potentially many diverse consumers and applications. Our primary source of data is a catalogue, however our ultimate goal is to conceptualise the Cultural Heritage domain at large, going far beyond the cataloguing perspective. This means that ArCo needs to draw its requirements by a plethora of diverse potential consumers, which can enter the process at any time, posing new requirements. XD is adequate to support evolving requirements, but how to gather requirements from an evolving community is unclear.

A third challenge concerns testing. In this case, given the dimension and complexity of ArCo ontologies we soon realised that systematic testing, as recommended by XD, needed proper tool support, which was unavailable to the best of our knowledge. We have developed TESTaLOD to serve this purpose, which is presented in Section \ref{sec:testalod}.
\subsection{Collecting requirements from an evolving heterogeneous community}
\label{sec:customer}
%
When the project started, its main \emph{customer} was ICCD, i.e. the institute in charge of collecting and preserving the data of the General Catalogue, and of releasing updated cataloguing standards.
ICCD domain experts formed the customer team and provided indications to the design team for the selection and prioritisation of requirements. They also supported the design team in gaining a good comprehension of the cataloguing standards. ArCo ontologies reflect (hence are compatible to) ICCD standards. However, they are not committed exclusively to their interpretation of the Cultural Heritage domain, which is limited to the point of view of cataloguing practices. The situation we faced is that ICCD wanted to address the need of diverse communities of potential consumers of CH data, while keeping an institutional management of the development process. 
We opted for a twofold approach: (i) we opened the process of requirements collection in the style of open-source projects, and (ii) we have launched an ``Early Adoption Program'' aimed at engaging a number of representatives of potential consumers, who would provide both requirements and validation. This approach has favoured a relatively quick creation of an open community as well as widening the scope of the collected requirements.
%
%
Early Adopters (EAs) are given assistance and support, and the fulfillment of their issues and requirements are put high in priority. In order to guarantee a lively interaction within the project community, regular meetings (e.g. webinars or meetups) are held and issues are discussed in an open mailing list. Furthermore, proposals for improvement and bugs can be submitted GitHub issues\footnote{\url{https://github.com/ICCD-MiBACT/ArCo/issues}}. 
%
%

With this approach, the customer team became an evolving creature, which over time will extend by involving all representatives of potential producers and consumers of CH data. At the moment, ArCo is collecting requirements from private companies, public administrations, researchers and creative developers.
These requirements are collected in the form of small stories (according to XD). A story is a non-structured text of maximum 250 characters,
exemplifying some scenario or reporting real use cases. They are submitted by the customer team to a Google Form\footnote{\url{https://goo.gl/forms/zCixt3B1ABYbj9JS2}}. The form requires to associate each story with one of three categories expressing the type of project motivating it: (i) publishing CH data, (ii) linking existing LOD to ArCo KG, (iii) feeding some applications with ArCo KG or providing services based on it. More stories can be associated with a reference custom project name, and additional material can be uploaded (e.g. a sample of data in the original format).

Requirements coming from user stories, as well those extracted from ICCD standards, are translated into Competency Questions. All CQs, and related SPARQL queries, that so far guided ArCo KG design and testing are available online\footnote{
\url{https://github.com/ICCD-MiBACT/ArCo/blob/master/ArCo-release/test/CQ/CQs-SPARQLqueries.txt}}.

As an example, we report one of the stories collected through the Google form:

\begin{adjustwidth}{0.5cm}{}
\emph{Type: Linking my data to ArCo data}
\end{adjustwidth}
\begin{adjustwidth}{0.5cm}{}
\emph{Title: Cultural heritage and residential property}
\end{adjustwidth}
\begin{adjustwidth}{0.5cm}{}
\emph{Story: I am looking for a residential property to buy, and I want to filter the results based on the type of cultural heritage nearby.}
\end{adjustwidth}

This story requires to address the following CQs: ``What are the types of cultural properties located in a certain area?'', ``Which is the current location of a cultural property?'', ``Which are the geographic coordinates of the current location of a cultural property?'', ``Which is the type of a cultural property?''.
Other examples of stories concerned: linking  cultural properties to multimedia resources, such as photographic documentation; describing specific attributes of drawings or music heritage; tracking over time the availability of cultural properties that have been confiscated from organised crime; relating catalogue records to heritage protection agencies.

\subsection{An architectural pattern for large ontology networks}
\label{sec:modules}
Handling large ontologies is a non-trivial challenge for ontology engineers, reasoners and users. A modular approach, as opposed to a monolithic design, i.e. one ontology module addressing all CQs, favours readability, reusability and maintainability of an ontology \cite{modular-StuckenschmidtK07,ParentSpaccapietra09}.
Ontology modules are meant to identify conceptually coherent subparts of the domain. In this respect, XD lacks explicit guidelines on how to approach a modular design of potentially large, networked ontologies. 
%
Based on our experience in designing the architecture of the ArCo ontology network, we provide a set of guidelines as well as an architectural pattern that can be applied in other contexts with similar characteristics as the ArCo's project. %
%

\paragraph{The root-thematic-foundations architectural pattern.} 

\begin{figure}[!ht]
\centering
	\includegraphics[scale=0.23]{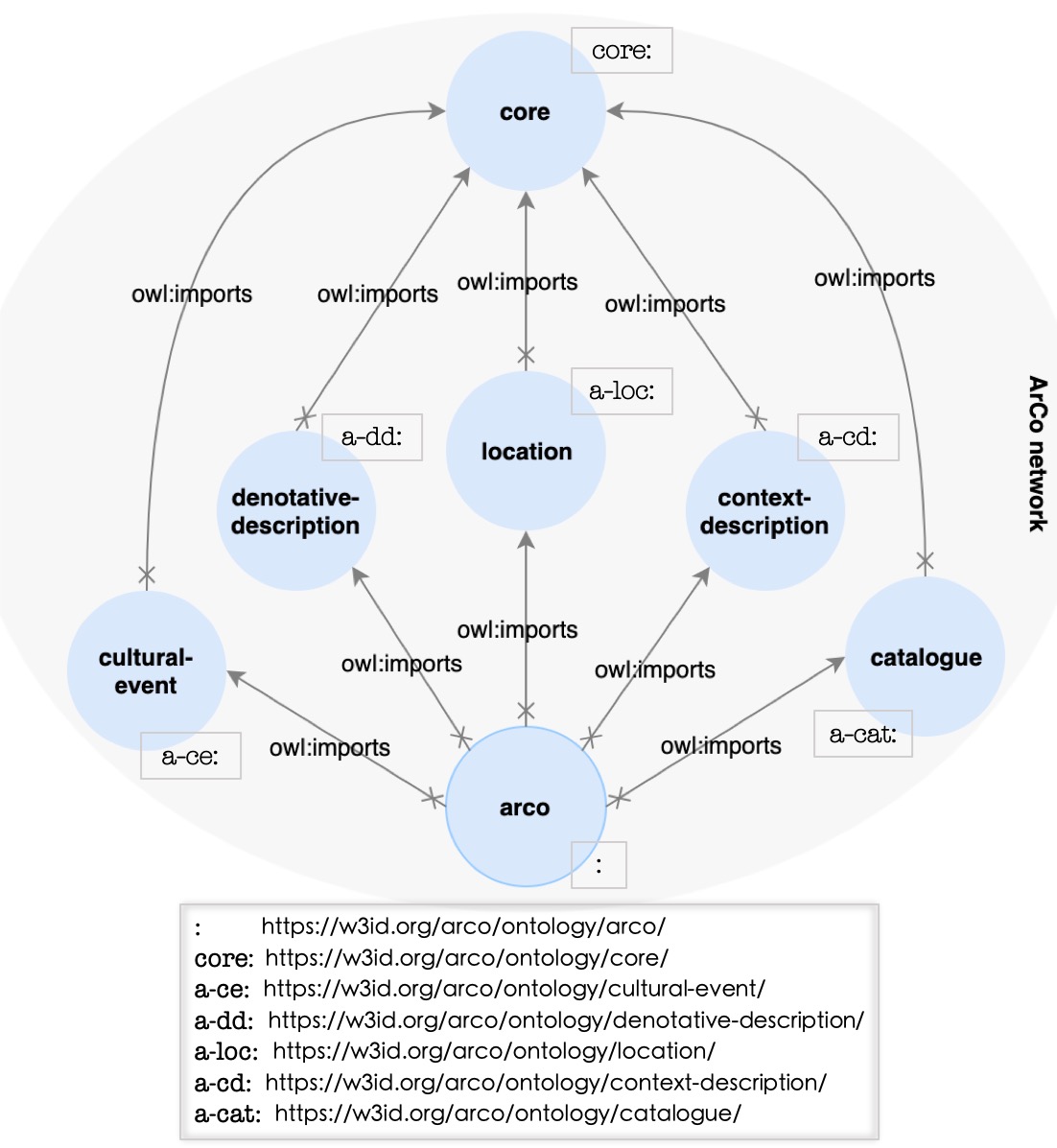}
	\caption{ArCo ontology network, currently including seven modules: \emph{arco} is the root node of the network, while \emph{core} is reused by all other modules, where concepts related to cultural properties and catalogue records are represented.
	}
	\label{img:network}
\end{figure}

We name the architectural pattern implemented by the ArCo ontology network: \emph{root-thematic-foundations} (Figure \ref{img:network}). It can be described as follows:
\begin{itemize}
    \item a root module acts as the \emph{entry point} of the network, i.e. it causes the whole network to be loaded by importing all main \emph{thematic modules}. In ArCo this is the \emph{arco} module, and it also contains the ontology top-level hierarchy of classes, with \smalltt{:CulturalProperty} as its root class. The root module may also contain ontology alignments. Alternatively, a separate module, i.e. alignment module, may be dedicated to this function and import the root module. With this configuration the alignment module acts as an alternative entry point to the network. 
    \item a second layer of the network is composed of the main \emph{thematic modules}, which are all imported by the root module. These modules may import, in turn, secondary thematic modules, that depend on them (which may form additional layers in the network). ArCo ontology network currently contains five main thematic modules: \emph{cultural-event}, \emph{denotative-description}, \emph{location}, \emph{context-description}, and \emph{catalogue}.
    \item a leaf module contains foundational concepts such as the part-whole relation, agent, physical object, role, etc. i.e. which are not domain-specific. This module is imported by all main thematic modules. In ArCo this is the \emph{core} module.
\end{itemize}

The implementation of the \emph{root-thematic-foundations} pattern requires a conceptual organisation of the domain into separate coherent subdomains. This can be achieved by clustering the requirements, based on thematic areas. The criteria for identifying thematic areas, and their granularity, can vary depending on the project's commitment, design choices and the size of the domain. As ontologies are evolving objects, new (main) thematic modules may be added over time in case future requirements identify new subdomains. 

In the context of ArCo, at a very early stage of development, we could leverage and analyse the ICCD catalogue, its data and standards, as well as the user stories provided by the customer team. In agreement with ICCD domain experts, we first focused on the \emph{cross} cataloguing standard and related user stories, which address concepts that are relevant for all types of cultural properties. Our hypothesis is that the \emph{cross} standard can give us a plausible overview of the CH domain. In all cataloguing standards, metadata are grouped into paragraphs, each containing different fields. We performed a manual clustering of these fields. This activity allowed us to identify five topics that could characterise the main thematic modules of the ArCo ontology network.

A first observation is that catalogue records contain: (i) data directly describing a cultural property and its contexts (e.g. techniques and materials, related exhibitions, surveys); (ii) data about catalogue records themselves (e.g. when they were created, by whom, their version, etc.); (iii) data about other entities referring to cultural properties (e.g. inventories, documentation, bibliography).
%
Based on this observation, a main thematic module (\emph{catalogue}\footnote{\smalltt{a-cat:} \url{https://w3id.org/arco/ontology/catalogue/}}) is dedicated to the ICCD General Catalogue (GC), and in particular to its catalogue records, their attributes and evolving process. 

Cultural properties, which are the main subjects of study of the CH domain, are described by means of  measurable, intrinsic aspects such as length, weight, materials, conservation status, as well as properties deriving from an interpretation process, such as  authorship attribution, dating. This conceptual distinction suggested us to define two additional thematic modules of the network: \emph{denotative description}\footnote{\smalltt{a-dd:} \url{https://w3id.org/arco/ontology/denotative-description/}} as for capturing descriptions of the first type, and \emph{context description}\footnote{\smalltt{a-cd:} \url{https://w3id.org/arco/ontology/context-description/}} that encodes interpretation situations and their related objects (e.g. inventories).

Finally, it results fairly evident that the \emph{locations} associated with a cultural property and the \emph{cultural events} in which it participates in, are two major components of its lifecycle. As a consequence, the ArCo ontology network includes the two thematic modules \emph{location}\footnote{\smalltt{a-loc:} \url{https://w3id.org/arco/ontology/location/}} and 
a \emph{cultural event}\footnote{\smalltt{a-ce:} \url{https://w3id.org/arco/ontology/cultural-event/}}. The module \emph{location} is dedicated to the different types of locations of a cultural property (e.g. current location, where it was found, where it was exhibited, where it was created, where it was stored, etc.), and to represent physical sites (e.g. a palace), geometrical features and related cadastral entities such as cadastral maps. The thematic module \emph{cultural event} is dedicated to classes and properties representing attributes of cultural events, including events that recur over time (cf. Section \ref{sec:recurrent-events}), e.g. festivals, recurrent exhibitions, festivities. 

The foundational concepts captured by the  \emph{core}\footnote{\smalltt{core:} \url{https://w3id.org/arco/ontology/core/}} module, and reused in all thematic modules, are described in detail in Section \ref{sec:situation}.

Finally, the \emph{arco}\footnote{\label{ref:arco-module}\smalltt{:} \url{https://w3id.org/arco/ontology/arco/}} module is the entry point of the network and defines the top-level hierarchy of CH concepts in ArCo, which is described in details in Section \ref{sec:situation}. 


\begin{table*}[htp]
\begin{center}
	\caption{Representative competency questions answered by ArCo ontology network.}
	\label{tab:arco-cq}
	\begin{tabular}{p{0.7cm}|p{6cm}||p{0.7cm}|p{6cm}}
		{ \bf ID } & { \bf Competency question } & { \bf ID } & { \bf Competency question }\\\hline 
		{ \bf } & { \bf ArCo module } & { \bf } & { \bf Cultural Event module }\\\hline
		CQ\theCQscounter{} & Is a cultural property tangible or intangible? & CQ18 & In which cultural events and exhibitions a cultural property has been involved?\\\hline
		CQ\incrementCQscounter & Is a cultural property movable or immovable? & CQ19 & Which cultural properties have been involved in a cultural event? \\\hline
		CQ\incrementCQscounter & Which are the cultural properties of a given type? & CQ20 & Which are the events of a recurrent event series? \\\hline
		CQ\incrementCQscounter & Which are the components of a complex cultural property? & CQ21 & Which is the time period elapsing between two events of a recurrent event series? \\\hline
		CQ\incrementCQscounter & Which is/are the residual(s) of a cultural property? & CQ22 & Which are the unifying criteria shared by all the events in a recurrent event series? \\\hline
		{ \bf } & { \bf Location module } & { \bf } & { \bf Denotative description module}\\\hline
		CQ\incrementCQscounter & Which are all the places where a cultural property has been located? Which are their types? & CQ23 & Which is the conservation status of a cultural property at a certain time?\\\hline
		CQ\incrementCQscounter & When has a cultural property been located in a place? & CQ24 & What is the technical status of a cultural property at a certain time?\\\hline
		CQ\incrementCQscounter & Which are the geographical coordinates of a cultural property? & CQ25 & Which are the measurements of a cultural property? \\\hline
		CQ\incrementCQscounter & Which are the cadastral data associated to the cultural property location? & CQ26 & Which are the elements, e.g. inscriptions, affixed on a cultural property? \\\hline
		{ \bf } & { \bf Context description module} & { \bf } & { \bf Catalogue module}\\\hline
		CQ\incrementCQscounter & Which are the authors attributed to a cultural property based on an interpretation criterion? & CQ27 & Which is the level of detail of the catalogue record?\\\hline
		CQ\incrementCQscounter & When has a cultural property been created? & CQ28 & When was a catalogue record created or updated?\\\hline
		CQ\incrementCQscounter & Which is the subject represented on a cultural property? & CQ29 & Which are all the versions of a catalogue record?\\\hline
		CQ\incrementCQscounter & Which are the current and/or previous owners of a cultural property? & CQ30 & Which is the (immediate) previous catalogue record version of a catalogue record version? And which is the (immediate) next one?\\\hline
		CQ\incrementCQscounter & Who commissioned a cultural property at a certain time? & CQ31 & Who, and playing which role, was responsible for creating, editing and updating a catalogue record?\\\hline
		CQ\incrementCQscounter & Which are the bibliography and documentation related to a cultural property? & CQ32 & Which is the catalogue record describing a cultural property? \\\hline
		CQ\incrementCQscounter & Which interventions and surveys have been carried out on a cultural property at a certain time? & CQ33 & Which is, and for what reason, the level of privacy of a catalogue record? \\\hline
		CQ\incrementCQscounter & Which collection a cultural property is member of?  \\\cline{1-2}
	\end{tabular}
	\end{center}
\end{table*}
\paragraph{Competency Questions.} Each module of the ontology network addresses a subset of the Competency Questions elicited by the customer team. Table \ref{tab:arco-cq} lists some representative CQs for each module, except the \emph{core} module, which is specialised by the other modules.
\section{Ontological foundations in ArCo}
\label{sec:situation}
Ontology Design Patterns (ODPs) are established solutions to modelling problems (i.e. requirements) that emerge from, and evolve through, applied and theoretical results. ODPs can have relations among them, including subsumption, overlap, merge, etc. (cf. \cite{DBLP:series/ihis/GangemiP09}). ODP subsumption is at the core of many formal ontology issues concerning the usefulness of foundational ontologies \cite{DBLP:journals/aim/GangemiGMO03,DBLP:conf/semweb/Gangemi05}. For example, the competency question \textit{Where is a cultural property currently located?} could be subsumed by a general one: \textit{What is the location of something at some time?}, and if a solution is not available for the specific requirement, we can reuse one that is good for the general requirement. In practice, when all the predicates from a requirement are stripped out of specificity, we get a foundational requirement, and if that requirement has a known solution (e.g. the \emph{Time Indexed Location} ODP), we can apply it directly, by specialising the predicates as expressed in the specific requirement (\textit{cultural property}, \textit{current}). 

When dealing with an ontology project as complex as ArCo, we need a good deal of generalisation that provides a shared modeling style to its data. The details of the ontological choices made against requirements are presented in Section \ref{sec:odps}.

\subsection{Foundational commitment in ArCo: DOLCE-Zero}

The ODPs implemented in ArCo ontologies are mostly taken from a set of interrelated foundational ODPs, inspired by DOLCE UltraLite+DnS (DUL)\footnote{\smalltt{dul:}, \url{http://www.ontologydesignpatterns.org/ont/dul/DUL.owl}} \cite{DBLP:books/ios/p/PresuttiG16}, and DOLCE-Zero (d0)\footnote{\smalltt{d0:}, \url{http://www.ontologydesignpatterns.org/ont/dul/d0.owl}} \cite{DBLP:conf/semweb/PaulheimG15}. 

DUL is a commonly used foundational ontology that commits to (i) DOLCE \cite{DBLP:journals/aim/GangemiGMO03} distinctions: objects vs. events vs. qualities (specific attributes of objects and events) vs. qualia (dimensional representations of qualities), and to (ii) D\&S \cite{MasoloKR04,DBLP:journals/aamas/Gangemi08} distinctions for second-order entities: situations vs. descriptions vs. concepts (see Section \ref{cdns} for details), including e.g. types, topics, roles, tasks, quality types, parameters, reified relations and classes, etc.

DOLCE-Zero contains a small set of classes on top of DUL, relaxing ambiguity resolution when needed. In particular, it introduces four ``union classes'':\\ \smalltt{d0:Characteristic}, \smalltt{d0:Eventuality}, \smalltt{d0:\-Ac\-tiv\-i\-ty}, and \smalltt{d0:Location}\footnote{d0 unions are formalised as follows:\\ 
$\texttt{d0:Characteristic} \equiv \texttt{(dul:Quality} \sqcup \texttt{dul:Region)}$\\
$\texttt{d0:Eventuality} \equiv \texttt{(dul:Event} \sqcup \texttt{dul:EventType)}$\\
$\texttt{d0:Activity} \equiv \texttt{(dul:Action} \sqcup \texttt{dul:Task)}$\\
$\texttt{d0:Location} \equiv \texttt{(dul:SpaceRegion} \sqcup \texttt{dul:PhysicalPlace} \sqcup \texttt{dul:[Social]Place)}$
}
that generalize some disjoint classes from DUL that are sometimes considered too ``strict'', e.g. qualities vs. dimensional regions, events vs. situations, actions vs. tasks, space regions vs. physical locations, etc. A widespread case is co-predication of physical objects, locations, and organisations.  For example, the Uffizi in Florence can be categorised as a Building (physical object), a Museum (a social object), and a relative Location (a spatial region), with experts understanding the Uffizi as a complex entity, whose heterogeneous features are not supposed to be analysed into three different categories, since they emerge out of \textit{co-predication} \cite{Pustejovski:95}. 

However, DOLCE (and its OWL implementation in DUL) has disjoint classes for physical vs. social objects vs. spatial regions), so inducing inconsistencies in the Uffizi knowledge graph. In practice, those distinctions are seldom represented in lightweight ontologies and natural language lexicons, originating debatable inconsistencies, as argued in \cite{DBLP:conf/semweb/PaulheimG15}, which reports a large-scale experiment that uses D0 to detect millions of inconsistencies in the DBpedia knowledge graph (without the relaxation provided by d0 many more would have been detected). 



ArCo foundational distinctions have a similar foundational commitment as DOLCE-Zero, so using more specific DUL's distinctions only when necessary. This commitment is implemented in \emph{Level 0}\footnote{\smalltt{l0:},  \url{https://w3id.org/italia/onto/l0/}}, which is part of the OntoPiA ontology network\footnote{\label{ref:ontopia}\url{https://w3id.org/italia/onto/FULL/}}, a standard reference for the Italian Public Administration. Furthermore, ArCo makes a general constructive commitment, embracing a broader, cognitive notion of \textit{situations} compared to the one defined in DUL: the ArCo \smalltt{core:Situation} class is made equivalent to the DOLCE-Zero \smalltt{d0:Eventuality} class. This decision  generalises over different event-like notions \textit{by design}, not only as a means to relax ambiguity, as explained in the next section.

\subsection{Constructive stance in ArCo}
\label{cdns}

A cluster of foundational requirements in ArCo is about events, states, actions, and their expressions, types and interpretations. Literature on these notions is heterogeneous \cite{casati1997fifty}, applying pragmatical, logical, and philosophical criteria, often mingled, to draw distinctions. As an example of the underlying problems, we can distinguish (i) a \textit{restoration} event, $e$ on a cultural property $c$, (ii) the \textit{restored} state $s$ of $c$, (iii) their types or categorizations (e.g. procedures, phases, tasks, roles) $t_{1...m}$ of $e$ or $s$, (iv) the propositions $p_{1...n}$ that \textit{report} $e$ or $s$, (v) the relationship $r$ holding between $c$, the participants in $e$ and $s$, and the propositions $p_{1...n}$, as well as (vi) the interpretation relation $i$ functioning as the intensional counterpart to $r$. In ArCo, we have requirements for most of those distinctions: sometimes we need to talk about events, states, their types, their reports, the relationships between the participants, and even interpretations of those relationships. 

A close, only superficially unrelated problem is that ArCo requirements (as most ontology design projects for the Semantic Web) need to represent n-ary relations (with $n>2$) in a logical language (OWL) that only can express unary or binary predicates. Time indexing is a major driver of n-ary relations. For example, the location of a cultural property can be true at a certain time, but not another, its official name can change, its physical structure can change due to restoration or natural events, the location itself or some of its properties, e.g. its name, can be indexed in time because of political or social changes.

While apparently this is a representation problem, rather than an ontological one, there is ample evidence (see e.g. \cite{davidson67lf,gentner13,DBLP:books/ios/p/GangemiP16,DBLP:journals/semweb/Gangemi20}) that the same cognitive constructions apply to events, states, actions, event types, action schemas, frames (in the sense of \cite{fillmore1976frame}), and relations as first-order objects. 

Based on this assumption, a useful generalisation can be applied, treating those entities as either (i) reified (extensional) relationships, e.g. the situation of Canova's Venus Victrix being located at Galleria Borghese in Rome since 1838, or (ii) reified (intensional) relations, e.g. the Attribution frame representing  the authorship attribution to cultural properties, as made by an interpreter based on some criteria.

That generalisation is applied quite often in ontology design. ODPs such as \emph{situation}, \emph{description}, \emph{descriptionsituation}, \emph{planexecution}, etc. (cf. \cite{DBLP:books/ios/p/PresuttiG16}), inspired by the theory of Constructive Descriptions and Situations (cDnS) \cite{DBLP:journals/aamas/Gangemi08}, provide a framework to systematically relate frames (a.k.a. intensional relations, schemas, descriptions), and frame occurrences (a.k.a. extensional relations, situations, states of affairs). 

The generalisation implements the cognitive assumption of extensional relations in the world being ``framed'' or ``schematised'' through observation, interpretation, diagnosis, norm, expectation, etc. In other words, framing applies a conceptual construction to a set of sensory perceptions, given data, reported facts, etc. The correspondence between frames and their occurrences leads to assigning contextual roles to participants (e.g. being a \textit{restored} object in a Restoration frame occurrence), tasks/types to actions (e.g. being a \textit{completion phase} in a Restoration frame occurrence), parameters to data values (e.g. being a \textit{reliable dating} in an occurrence of a frame for evaluating reliability of Carbon-14 dating of an archaeological object). Examples of application of those ODPs can be found in multiple domains \cite{GCB04,GangemiLegal2005,Oberle06,DBLP:journals/aamas/Gangemi08,DBLP:conf/kcap/ScherpFSS09,DBLP:conf/ekaw/GangemiAAPR16}.


ArCo adopts the framing patterns to the representation of cultural properties, using the class \smalltt{core:\-Sit\-ua\-tion} for frame occurrences. See Sections \ref{par:dynamics}, \ref{par:situations} and \ref{sec:recurrent-events} for the situation types modelled in ArCo. 

\subsection{Relations as graph patterns vs. triples}

While situations allow to generalise over any relational concept, regardless of their arity, semantic web practices seem to prefer binary predicates (called \textit{object} or \textit{datatype properties}) whenever useful. In order to accommodate the constructive stance of ArCo, which requires graphs including multiple triples for each relation, with the practical benefit of having simpler graphs, with one triple representing each relationship, the same relational concepts are represented as both $\ge2$-ary (situation classes) and projected binary relations (OWL properties). This redundancy supports both high-level modelling needs (time indexing, evolution, changes), and lightweight modelling. 
Where possible, an OWL property chain abridges a situational graph representing a $\ge2$-ary relation to its binary projection(s).

For instance, agencies related to a cultural property (e.g. a \textit{cataloguing agency}) can be represented in ArCo as either: (1) a situation class (\smalltt{core:\-Agent\-Role}) that includes the cultural property, the agency, and the role it played with respect to that cultural property, and (2) an OWL object property (\smalltt{:has\-Related\-Agency}), which directly links the cultural property to the agency, and is subsumed by the property chain  \smalltt{[core:has\-Agent\-Ro\-le O core:\-has\-Agent]}.

\subsection{Other foundational event-like notions}
\label{other-event-notions}

Other event-like notions in foundational and cultural ontologies can be aligned as subclasses of \smalltt{core:\-Si\-t\-ua\-tion}. Two examples are provided here. 

CIDOC CRM E5 Event, subclass of E4 Period, is defined as \textit{... changes of states in cultural, social or physical systems, regardless of scale, brought about by a series or group of coherent physical, cultural, technological or legal phenomena ... The distinction between an E5 Event and an E4 Period is partly a question of the scale of observation. Viewed at a coarse level of detail, an E5 Event is an `instantaneous' change of state}. CIDOC events are then considered as a granularity (or ``aspectual'') distinction within the larger class of periods, which seem to encompass also  other spatio-temporal phenomena, except E3 Condition State, which is defined as a particular state of an entity. Both E4 Period and E3 Condition State are subclasses of E2 Temporal Entity, which ``encompasses all phenomena''. Orthogonally, ArCo situations focus on the role structure of event-, state-, and relation-like entities. All CIDOC temporal entities can be considered situations in ArCo, while their aspectual distinctions are not directly addressed in ArCo, since they could be defined as subclasses of \smalltt{core:\-Sit\-ua\-tion} if needed in a cultural requirement. 
We remark that in CIDOC CRM participation (and therefore the possible roles of participants) is only defined for E5 Event, while all ArCo situations can have participants with a role.

As a second example, DOLCE \cite{DBLP:journals/aim/GangemiGMO03} notion of Event (a.k.a. Perdurant or Occurrent) is defined axiomatically as the class of entities that ``happen in time'', i.e. some of their proper parts/phases may not be present at each time they are present (extreme cases include instantaneous and stationary states). Participation is defined in DOLCE for all events. An extension of DOLCE \cite{MasoloKR04} axiomatises also intensional relations (called \textit{descriptions}) as schemas, with proper roles, for events or other entities. The notion of frame/description/intensional relation used in ArCo is compatible with that of \textit{description} in \cite{MasoloKR04}, while ArCo situations do not commit to the distinction between objects and events as applied in DOLCE (and CIDOC CRM), since a situation is defined as the occurrence of a description as observed, diagnosed, aggregated, invented, etc. In this sense, the \textit{constructive} approach inherited from cDnS \cite{DBLP:journals/aamas/Gangemi08} departs from other foundational ontologies, which focus on the 3D/4D distinction as primary to declare their commitments. The constructive stance prioritises the dependence of an event-like entity on its framing, so that a requirement for ontology design can be directly matched by a frame. 

This stance, besides cognitive results, is also inspired by Davidson \cite{davidson67lf}, which provides a solid ground to events as first-order entities corresponding to (reified) relationships. 

Finally, as discussed in Section \ref{epist}, a constructive stance is immediately applicable to the \textit{interpreted} nature of cultural entities: historical, anthropological or archaeological events are always dependent on some interpretation for their cultural identity to be established, and cultural properties may change their meaning when interpretation conditions change.

\subsection{Epistemological stance in ArCo}
\label{epist}

Related to the previous section, ArCo applies a distinction between three epistemological levels: \textit{factual}, \textit{interpretation}, and \textit{reporting} situations, which constitute the core of cultural data dynamics. For example, \textit{having} physical size, constitution, unique qualities, or authorship are \textit{factual} situations for a cultural property $c$, while \textit{establishing} constitution via Carbon-14 or \textit{attributing} authorship for $c$ are \textit{interpretation} situations. Finally, providing data and content about $c$ is a \textit{reporting} situation. The cultural heritage scientist typically establishes factual situations based both on direct or sensor-based observation data, and on documented evidence, mediated by interpretation activities. Data provided by cultural agencies, researchers, or citizens do not necessarily contain the full story of how facts have been established/reported in precise, extended causal chains. In ArCo, we need to live with incomplete epistemological foundations, trying to associate situations with their epistemological level.

The epistemological stance in ArCo involves the distinction between: (i) catalogue versions, e.g. when someone adds data in a catalog record; (ii)  interpretations, e.g. when a catalogue record reports an attribution, and (iii) current factual data, e.g. the authoritative data for a cultural property at the current state of the art. When the catalogue record is the only source for a knowledge graph, we depend on its versions to reconstruct the epistemological trajectory of $c$, with its interpretations, updates, and the current state of its factual data. Section \ref{par:dynamics} describes the predicates used for relating cultural property situations and where they are reported e.g. catalogue records.

However, the vision of ArCo goes well beyond reengineering traditional catalogues, and its epistemological stance accommodates for a more complex knowledge graph hosting interpretations with different provenance and reliability, different reporting sources, and potentially conflicting factual data, so enabling cultural knowledge graphs as investigation tools for researchers.

\subsection{ArCo top level hierarchy and distinctions}

The most general Cultural Heritage concept modelled in ArCo is \smalltt{:Cultural\-Property}. This class includes all tangible and intangible entities that \emph{have been recognised} to be part of our cultural heritage. Once an entity is recognised as being part of cultural heritage, it never stops being a \smalltt{:Cultural\-Property}. 
For example, a commissioned artwork is not an instance of ArCo's  \smalltt{:Cultural\-Property}, unless or until it is officially recognised as such. Hence, according to the definition by \cite{guarino2000formal}, \textit{being a cultural property} is an \emph{essential characteristic} of all instances of \smalltt{:Cultural\-Property}, i.e. it is a rigid property in all possible worlds for ArCo's universe of discourse: the Cultural Heritage domain.

\begin{figure*}[!ht]
\centering
	\includegraphics[width=0.75\textwidth]{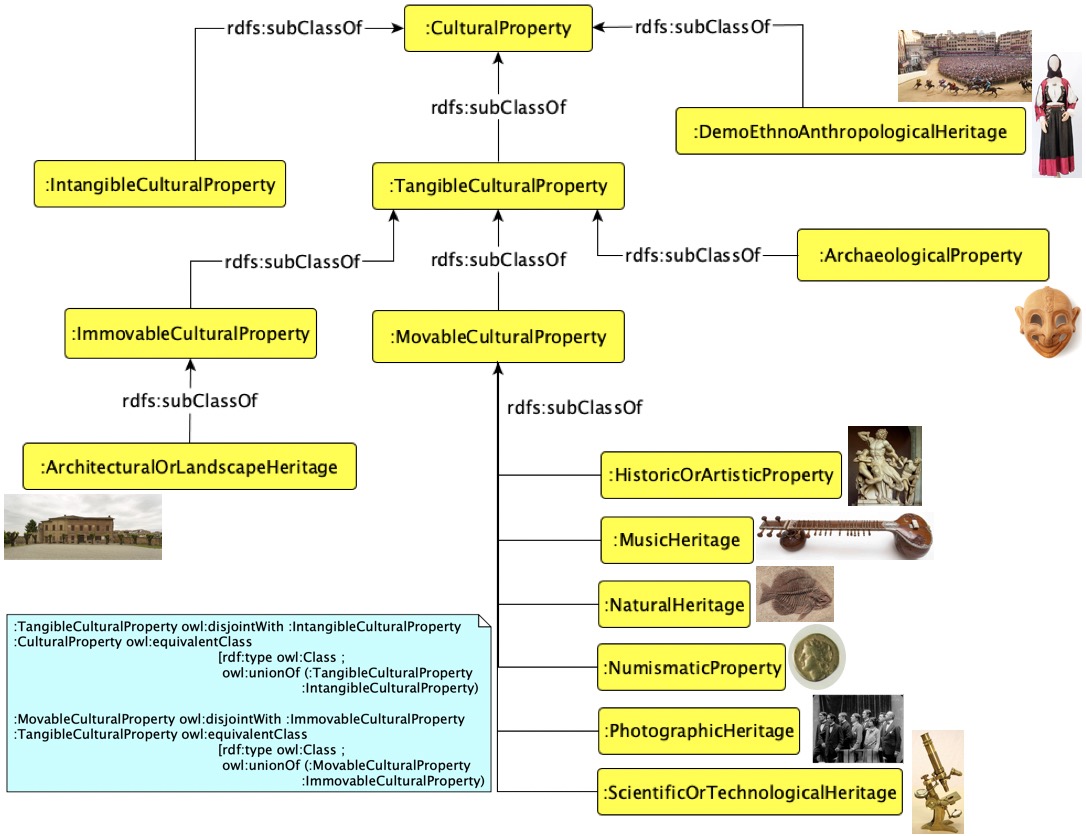}
	\caption{The taxonomy of cultural properties.}
	\label{img:taxonomy}
\end{figure*}

The root of ArCo's hierarchy (depicted in Figure \ref{img:taxonomy}) is the class \smalltt{:Cul\-tur\-al\-Pro\-per\-ty}.
In accordance to the main distinctions in Cultural Heritage identified by UNESCO\footnote{\url{http://www.unesco.org/new/en/culture/themes/illicit-trafficking-of-cultural-property/unesco-database-of-national-cultural-heritage-laws/frequently-asked-questions/definition-of-the-cultural-heritage/}}, \smalltt{:Cul\-tur\-al\-Pro\-per\-ty} is modelled as a \emph{partition} of two classes: \smalltt{:Tan\-gible\-Cul\-tu\-ral\-Pro\-per\-ty} and \smalltt{:In\-tan\-gi\-ble\-Cul\-tur\-al\-Pro\-per\-ty}. A tangible cultural property is a physical object, e.g. a photograph, an amphitheater, ancient garments. Intangible cultural properties are defined as including ephemeral performances, practices, skills or knowledge: oral literature, musical, choral or theatrical performances, handcrafted techniques, designs, algorithms. Intangible cultural heritage may be documented by text as well as audio and/or video recording. 

\noindent\smalltt{:Tan\-gi\-ble\-Cu\-ltur\-al\-Pro\-per\-ty} is further specialized in \smalltt{:Mo\-vable\-Cul\-tu\-ral\-Pro\-per\-ty}, i.e. objects that can be handled and moved by nature (e.g. a painting, a musical instrument), and \smalltt{:Im\-mo\-va\-ble\-Cul\-tu\-ral\-Pro\-per\-ty}, i.e. objects fixed or incorporated into the ground, which generally occupy a large area (e.g. an archaeological site, a palace and its gardens).

\subsubsection{Further specialisations based on Italian standards}
The ICCD standards extensively address different aspects of the cultural heritage domain in order to define the structure and content of catalogue records.  
Catalogue records can describe 30 types of cultural properties (cf. Section \ref{sec:catalogue}), each showing distinguishing features. ArCo's further specialisations in the top level hierarchy are inspired by these distinctions\footnote{We plan to reflect additional classifications as provided by official national or international standards.}. 

We report the definitions for the specific classes, according to ICCD standards.

\smalltt{:Demo\-Ethno\-An\-thro\-po\-log\-ical\-Her\-it\-age} can be either intangible or tangible, and is related to socially shared customs. Intangible demo-ethno-anthropological properties are unique and unrepeatable performances, transmitted orally or bodily, e.g. poems, traditional dances, music and performing arts, customary norms, centuries-old techniques, knowledge about ancient recipes, rituals.
Tangible demo-ethno-anthropological properties include physical objects such as body adornments, furnishings, means of transportation, ritual instruments.

\smalltt{:Ar\-chae\-o\-log\-i\-cal\-Pro\-per\-ty} includes tangible cultural properties that are signs of the ancient past, either movable or immovable: archaeological complexes consisting of several building units (e.g. inhabited areas, fortified centers), archaeological monuments as single building units (e.g. a tower, a Roman \emph{domus}, a temple) and archaeological sites, i.e. portions of territory that preserve archaeological evidence, are immovable, while anthropological materials, i.e. biological evidence related to archaeological contexts, and (batches of) archaeological objects (e.g. vases, jewelry, clothing, everyday items, masks) are movable.

\smalltt{:Ar\-chi\-te\-ctural\-Or\-Land\-scape\-Her\-it\-age} is by design immovable, such as monumental complexes, public or private, religious or rural, fortified or noble buildings of historical and/or artistic relevance. Landscape heritage includes green areas (parks, gardens) e.g. annexed to noble residences, botanical gardens, urban parks, cloisters.

\smalltt{:Hi\-sto\-ric\-Or\-Ar\-tis\-tic\-Pro\-per\-ty} refers to: handmade drawings on any support (e.g. paper, wood, stone) and different techniques (e.g. ink, pencil, charcoal); (contemporary) artworks (e.g. weapons and armors, paintings, statues); printing plates of various materials and related prints (e.g. lithographies); historic and contemporary garments, i.e. clothes to civil use, connected to private or social life.

\smalltt{:Mu\-sic\-Her\-it\-age} includes musical instruments: objects created specifically to produce sound according to different musical cultures, which can be of  archaeological, artistic, ethno-anthropological interest. 

\smalltt{:Nat\-u\-ral\-Her\-it\-age} represents objects related to botany (e.g. collections of dried plants), mineralogy (mineral specimens such as quartz), paleontology (fossils of animals, plants ichnofossils), petrology (specimens of rocks), planetology (specimens of meteorites), zoology (specimens from the animal kingdom such as butterfly collections).

\smalltt{:Nu\-mis\-matic\-Pro\-per\-ty} represents coins and other objects of numismatic interest (monetary punches, weights for monetary check, medals).

\smalltt{:Pho\-to\-graphic\-Her\-it\-age} refers to digital and analog photographs (e.g. negatives, positives, daguerreotypes), complex object like albums and artists' portfolios, photographic series as collections of photographs created or published as a unit.

\smalltt{:Sci\-en\-tif\-ic\-Or\-Tech\-no\-lo\-gi\-cal\-Her\-it\-age} includes instruments of interest to the history of science and technology and related to specific scientific disciplines (e.g. a telescope, a pendulum, an ancient sundial), machines, means of transport, etc.

%
\section{Matching requirements to Ontology Design Patterns}
\label{sec:odps}
In this section, we: (i) illustrate some of the main modelling issues that have emerged from ArCo's requirements, along with the modelling solutions adopted for addressing them; (ii) use (i) as driving examples to describe the process of matching requirements to Ontology Design Patterns (ODPs), introduced in Section \ref{sec:method}, as part of the XD methodology. 



\begin{figure}[!ht]
\centering
	\includegraphics[scale=0.25]{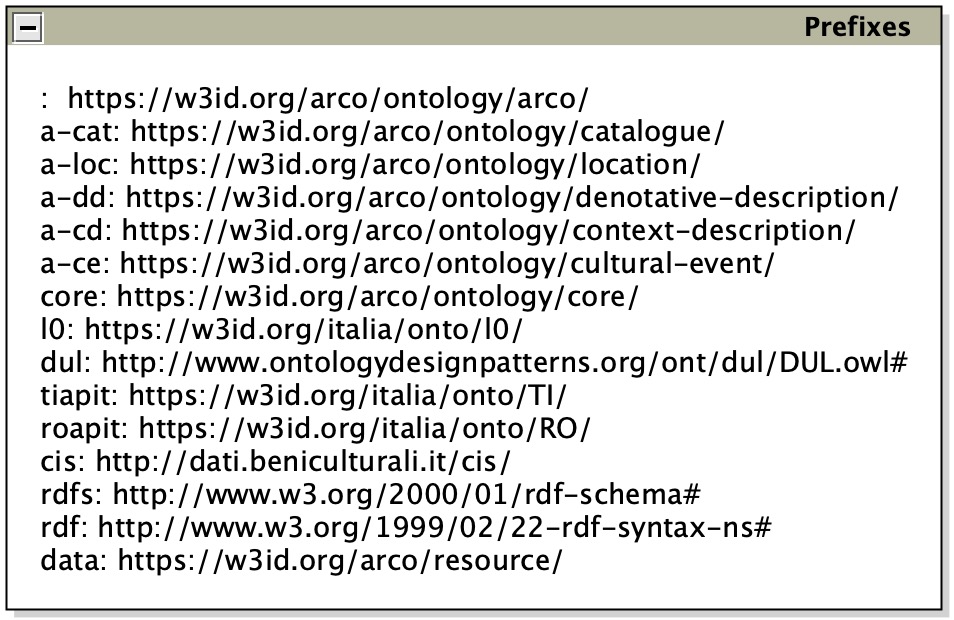}
	\caption{Prefixes used in the next figures.}
	\label{img:prefixes}
\end{figure}

Figure \ref{img:prefixes} shows all the prefixes used in the next diagrams.

\subsection{Representing dynamics}
\label{par:dynamics}

Dynamic concepts, such as situations that change over time, are present in almost every domain. There are different patterns that model dynamic situations: in this subsection, we exemplify ArCo approach to dynamicity with catalogue records and cultural property locations, which may both evolve over time.

\paragraph{A catalogue record as a fluent information object.}
\label{par:cat-rec}
A catalogue record is an entity that contains metadata about a cultural property. As it describes a real-word object, it can be defined as an \emph{information object}, i.e. a piece of information, independent from how it is concretely realised, describing something in the real word. This concept is defined in several ODPs, including \emph{Information Realization}\footnote{\url{http://www.ontologydesignpatterns.org/cp/owl/informationrealization.owl}} \cite{DBLP:books/ios/p/GangemiP16a}, which is reused in ArCo. The content of a catalogue record, i.e. the description of a cultural property, can change: \enquote{information about the creation of a catalogue record and possible following computerisation, update and corrections}. Indeed, different agents with different roles (e.g. the official in charge) can be recorded, and a date keeps track of the time interval associated with each action.

A catalogue record is then a fluent entity, an information object that changes as the description of its denoted cultural property changes\footnote{Cf. Section \ref{epist} about the difference between factual and reporting situations}. Possible corrections and updates implemented by a cataloguer can derive from (i) an ontological change of a cultural property's attributes (e.g. the conservation status from good becomes mediocre, in a badly preserved cultural property); (ii) an epistemological change, if the knowledge, which the catalogue record is based on, is either no longer complete, due to new knowledge acquired, or no longer valid, as a result of new research activities (e.g. after discovering new documentation, it turns out that another author played a role in creating the cultural property).

Every change of a catalogue record produces an information object, which is a new version of the catalogue record, including the reporting of a new situation involving a same persistent entity. Nevertheless, the catalogue record finds its persistence in describing the same real-word object, i.e. the same cultural property, independently from different versions of the content and the reported entity changes over time. Thus, the catalogue record is represented as a persistent information object, and is related to its versions, which are information objects reflecting changes of its content over time.

The \emph{Time Interval} ODP is used to represent the temporal validity of each version, and the pattern \emph{Sequence}\footnote{\label{ref:odp-sequence}\url{http://ontologydesignpatterns.org/cp/owl/sequence.owl}} to represent the sequence of consecutive information objects.

\begin{figure*}[!ht]
\label{img:catalogue-record}
\begin{subfigure}{\textwidth}
\centering
	\includegraphics[width=0.8\textwidth]{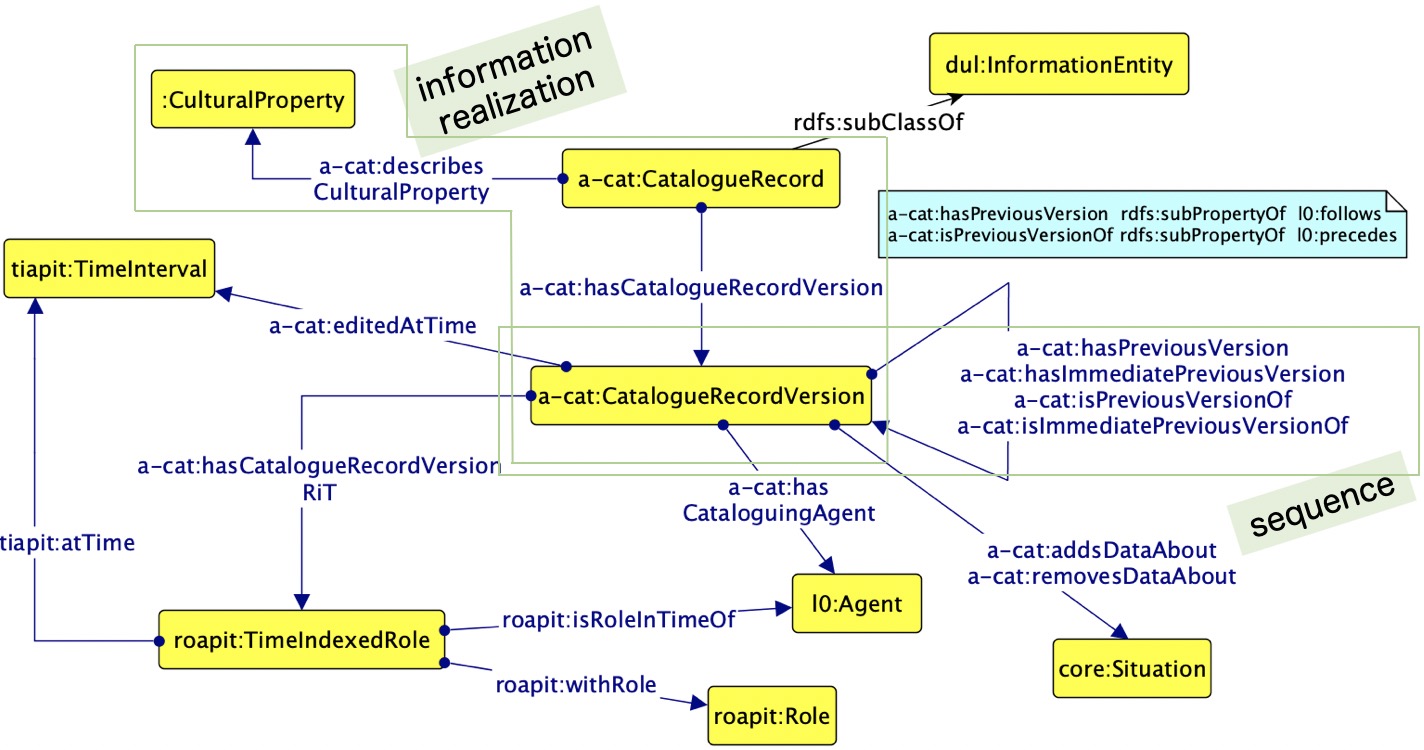}
	\caption{The model for catalogue records and their versions.}
	\label{img:catalogue-record-a}
\end{subfigure}
\begin{subfigure}{\textwidth}
\centering
	\includegraphics[width=0.9\textwidth]{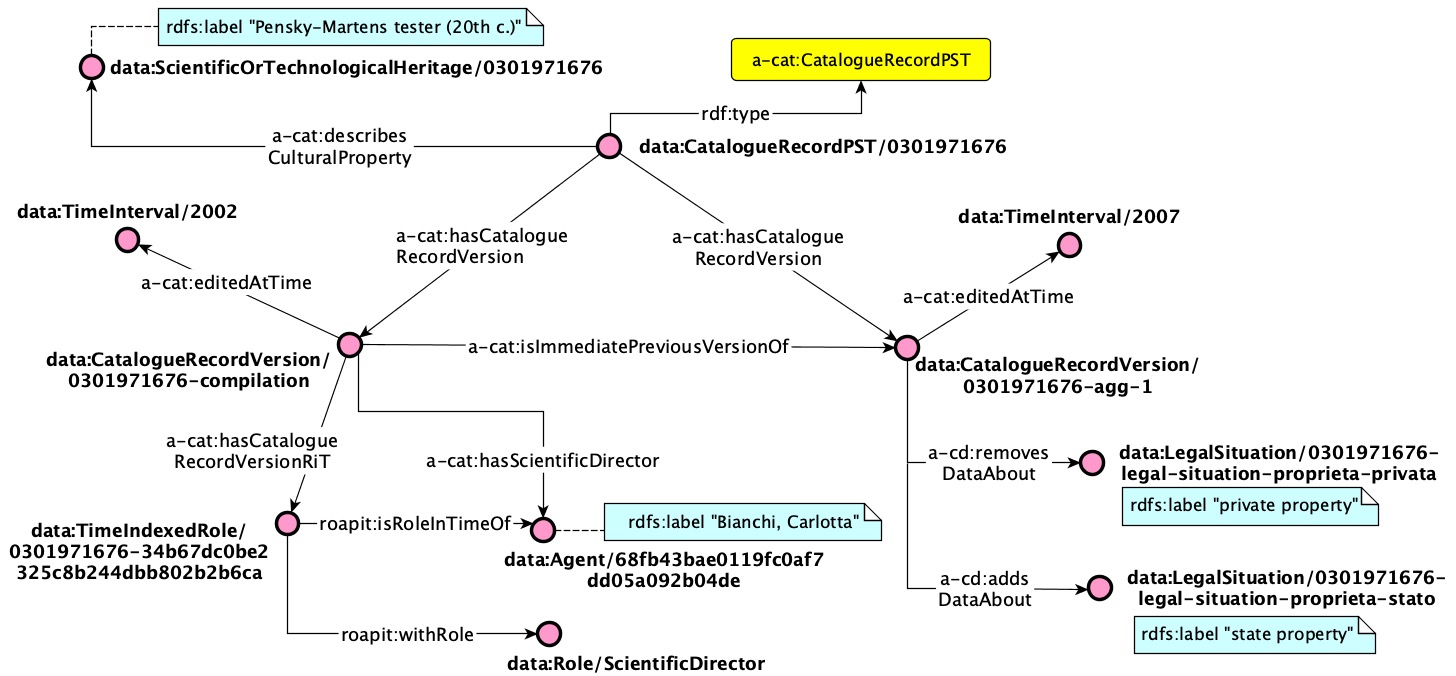}
	\caption{An instance of the model in Figure \ref{img:catalogue-record-a}, with two consecutive versions of a catalogue record about scientific or technological heritage.}
	\label{img:catalogue-record-b}
\end{subfigure}
\caption{Information Realization and Sequence ODPs reused for modeling catalogue records.}
\end{figure*}

Figure \ref{img:catalogue-record-a} depicts catalogue record modeling with the reused ODPs. A catalogue record is represented by the class \smalltt{a-cat:\-Cat\-a\-logue\-Rec\-ord}, which is aligned to \smalltt{dul:\-In\-for\-ma\-tionEn\-ti\-ty} with an \smalltt{rdfs:\-sub\-Class\-Of} axiom, and is related to the cultural property it describes.
Catalogue records have different \smalltt{a-cat:\-Cat\-a\-logue\-Rec\-ord\-Ver\-sion}s, which are linked to (new) data included (\smalltt{a-cat:\-adds\-Da\-ta\-About}), supported by a situation, or to deprecated data that were included in previous version(s) (\smalltt{a-cat:\-rem\-oves\-Da\-ta\-About}). Each version is associated with a time interval,  \smalltt{a-cat:ed\-it\-ed\-At\-Time}, and with agents involved in its creation (e.g. \smalltt{a-cat:hasCat\-a\-log\-uing\-A\-gent} with its subproperties). The agents involved in changing the catalogue record play some role, which has its own temporal validity, hence we reuse another pattern. \smalltt{roapit:Ti\-meInd\-ex\-ed\-Rol\-e}\footnote{\smalltt{roapit:} https://w3id.org/italia/onto/RO/} is modelled as a time-indexed situation (see Section \ref{sec:situation} for more details on situations) involving an agent, its role, and the temporal indexing of the agent-role relation. The object properties \smalltt{a-cat\-:has(Im\-me\-di\-ate)Pre\-vi\-ous\-Ver\-sion} and \smalltt{a-cat:is(Im\-me\-dia\-te)Pre\-vi\-ousVer\-sion\-Of} specialise the \emph{sequence} ODP, representing  (in)transitive \textit{previous} and \textit{next} versions of a catalogue record.

In Figure \ref{img:catalogue-record-b} we can see an instance of this model. The \smalltt{data:\-Cat\-alogue\-Record\-PST/0301971676}\footnote{\smalltt{data:} https://w3id.org/arco/resource/} is of type \smalltt{a-cat:\-Cat\-a\-logue\-Record\-PST}, i.e. a type of catalogue record describing scientific and technological heritage: indeed, this resource describes a \emph{Pensky-Martens tester}\footnote{\url{https://w3id.org/arco/resource/ScientificOrTechnologicalHeritage/0301971676}}. It has 2 versions: a first version is created when the cultural property is first catalogued (2002), and another version follows, as a result of editing and updating activities (2007). Thus, the first version, \smalltt{data:\-Cat\-a\-logue\-Record\-Ver\-sion\-/0301971676-com\-pi\-la\-tion}, is the immediate previous version of the second one: it has been encoded in 2002 and involved agents playing different roles in its compilation (e.g. the scientific director). This relation is expressed as both a binary relation and a $\ge2$-ary relation, whose range is a \smalltt{roapit:\-Time\-In\-dexed\-Role}, i.e. a situation involving an agent, its role and the time of its duration.

\paragraph{Multiple time-indexed and typed locations for one cultural property.}
A tangible cultural property, i.e. a physical object, is located in a physical place, which can be defined by a set of components: country, region, city, address, etc. For an immovable cultural property (e.g. a monumental park), this place overlaps with the area occupied by the cultural property, and to which it is fixed. Instead, for a movable cultural property (e.g. a photograph),
data about the address and the coordinates is referred to the building in which it is situated and preserved, and the related cultural institute. While an immovable cultural property, precisely because of its nature, will be related to a unique geographical place during its whole life cycle, a movable cultural property can be moved from a place to another. Different locations of a cultural property will hold in different time intervals.
As a consequence, the temporal indexing of the locations associated with a cultural property is represented, also promoting the reconstruction of the spatial trajectory of the cultural property over time. 
During its life cycle, a movable cultural property is involved at least in as many situations as the places in which it has been located, and each situation is associated with a time interval. The \emph{Time Indexed Situation}\footnote{\url{http://www.ontologydesignpatterns.org/cp/owl/timeindexedsituation.owl}} ODP \cite{DBLP:books/ios/p/GangemiP16}, which represents situations that have an explicit time parameter, is reused to satisfy this requirement. We also need to model the \textit{motivation} that links a cultural property to a location: it can be the place where it was found or created, an exhibition it was involved in, where it was temporarily stored, etc. A same place can play different \emph{roles} as location of one or different cultural properties, thus this \emph{role} should be valued in the time indexed situation.

\begin{figure*}[!ht]
\begin{subfigure}{\textwidth}
\centering
	\includegraphics[width=0.8\textwidth]{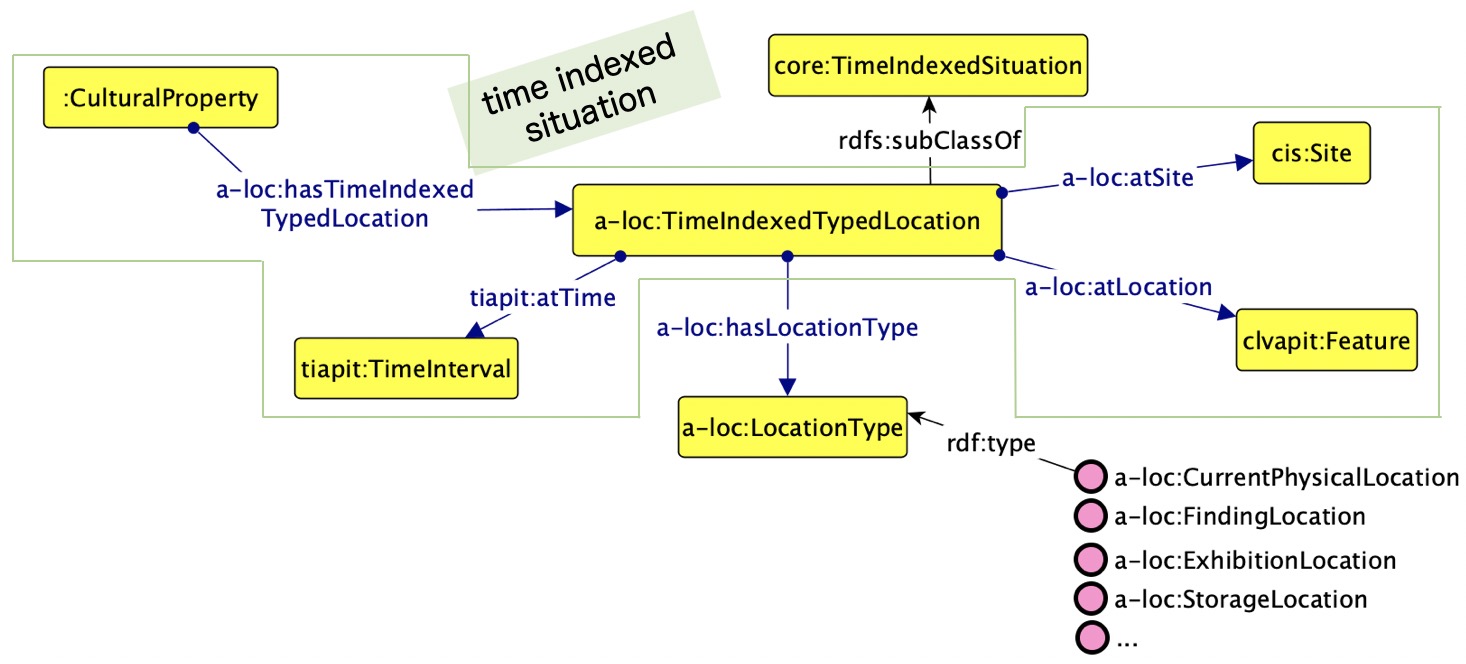}
	\caption{The model for time indexed typed locations.}
	\label{img:typed-location-a}
\end{subfigure}
\begin{subfigure}{\textwidth}
\centering
	\includegraphics[width=0.8\textwidth]{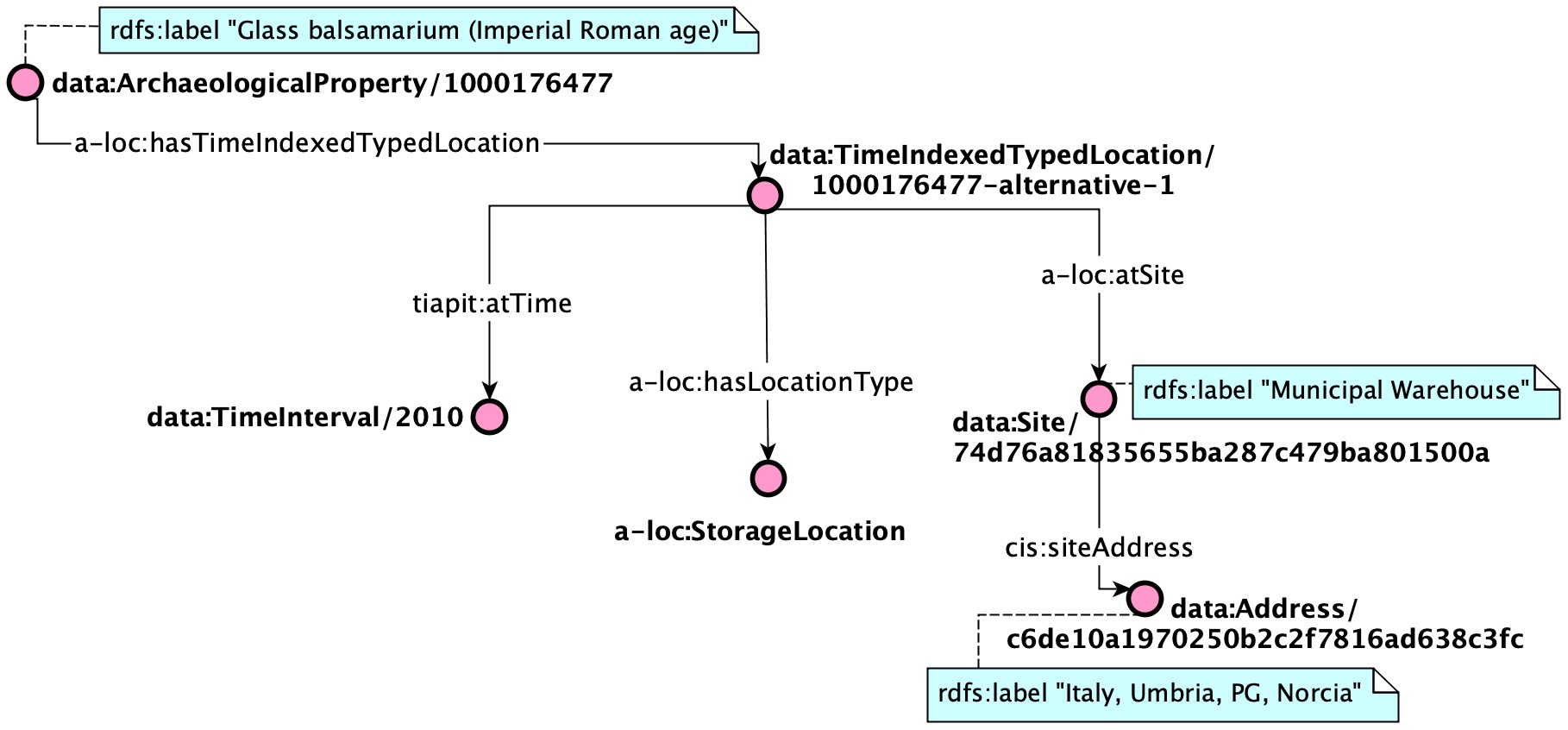}
	\caption{An instance of the model in Figure \ref{img:typed-location-a}, with the site where an archaeological property has been temporarily stored in 2010.}
	\label{img:typed-location-b}
\end{subfigure}
\caption{Time indexed situation ODP implemented for modelling different types of locations of a cultural property.}
\end{figure*}

Figure \ref{img:typed-location-a} shows the class \smalltt{a-loc\-:Time\-Ind\-ex\-ed\-Typed\-Lo\-ca\-tion} as the core class of the implementation of this pattern: it is a \emph{situation} of a cultural property that is \emph{located} in some place, at a certain point in \emph{time}, and with the location playing a specific \emph{role} in such situation, so providing a type to that situation. A time indexed typed location is therefore associated with a \smalltt{a-loc:\-Lo\-ca\-tion\-Type} (e.g. \smalltt{a-loc:\-Find\-ing\-Lo\-ca\-tion}, \smalltt{a-loc:\-Ex\-hi\-bi\-tion\-Lo\-ca\-tion}, etc.). \smalltt{tiapit:\-atTime} relates the situation to its temporal validity,
\smalltt{a-loc:\-at\-Lo\-ca\-tion} expresses the geographical entity involved in the situation, while its subproperty \smalltt{a-loc:\-at\-Site} is used when this entity is a physical building that hosts cultural properties, and is related to at least one cultural institute. For instance, the Pitti Palace is the \smalltt{cis:\-Si\-te} of the cultural institutes ``Palatine Galleries'', ``Museum of Custom and Fashion'' and others, and hosts many cultural properties.

Figure \ref{img:typed-location-b} depicts one of the time-indexed typed locations of a \emph{balsarium} glass from the Imperial Roman age\footnote{\url{https://w3id.org/arco/resource/ArchaeologicalProperty/1000176477}}: the place where it was temporarily stored. Indeed, \smalltt{data:\-Time\-Indexed\-Typed\-Lo\-ca\-tion/\-10\-001\-76477-al\-ter\-na\-tive-1} has \smalltt{a-loc:\-Stor\-age\-Lo\-ca\-tion} as location type, and the associated time interval and cultural site allow us to assert that this archaeological property has been stored in a Municipal Warehouse in the city of Norcia (Italy) in 2010.

\subsection{Situations and their descriptions}
\label{par:situations}
A cultural property can be involved in many different situations during its life: it can be commissioned, bought or obtained, used (e.g. a garment wore by one person), it can be part of a collection, photographic or numismatic series, can change its availability as a result of theft, destruction or rescue, etc. Each situation defines a contextual relation between the cultural property and the other entities involved. The  \emph{Situation}\footnote{\url{http://www.ontologydesignpatterns.org/cp/owl/situation.owl}} ODP  \cite{DBLP:books/ios/p/GangemiP16} models the concept of a contextual $\ge2$-ary (usually called $n$-ary) relation (see Section \ref{cdns}).

\begin{figure}[!ht]
\centering
	\includegraphics[scale=0.27]{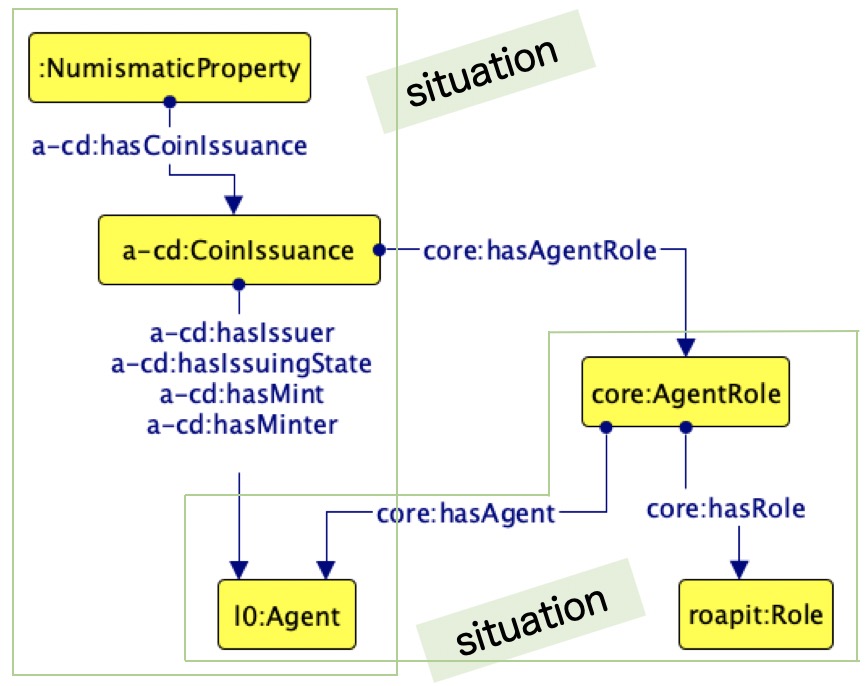}
	\caption{Situation ODP reused for representing the coin issuance.}
	\label{img:coin-issuance}
\end{figure}

For example, when a coin is issued, many entities play a role in such context: the cultural property itself, the issuer, the issuing State, the mint and the minter. The ``coin issuance'' is a situation representing the relation that keeps together all these entities for that purpose.
Figure \ref{img:coin-issuance} shows how we model the \emph{coin issuance} (\smalltt{a-cd:\-Coin\-Is\-su\-ance}) by implementing this ODP.

\begin{figure}[!ht]
\centering
	\includegraphics[scale=0.27]{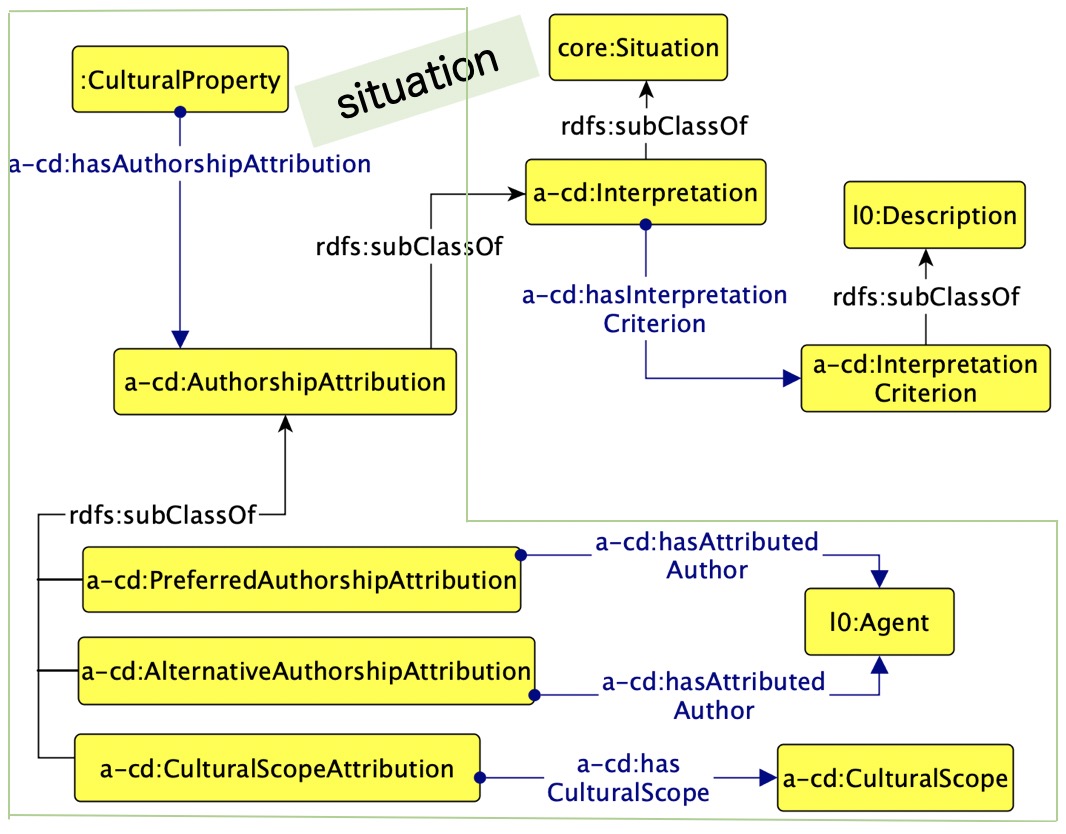}
	\caption{Situation ODP reused for the authorship attribution.}
	\label{img:authorship}
\end{figure}

A central situation in which a cultural property can be involved is the authorship attribution, a specific type of \smalltt{a-cd:\-Int\-er\-pret\-ation}, i.e. a situation in which pieces of information about a cultural property are processed by an agent, and produce explicit knowledge, based on a specific source or criterion (cf. Section \ref{epist} for the distinction between factual, interpretive, and reporting situations). 
As in Figure \ref{img:authorship}, an \smalltt{a-cd:\-Au\-thor\-ship\-At\-tri\-bu\-tion} is a situation in which one author is attributed to a cultural property, and this attribution is motivated by an \smalltt{a-cd:\-In\-ter\-pre\-ta\-tion\-Cri\-te\-ri\-on}, e.g. inscription, bibliography, documentation. Each cultural property has at least one preferred authorship attribution and/or a cultural scope attribution (e.g. Swedish workshop), and can have one or more alternative (i.e. previous and obsolete) authorship attributions.

\begin{figure*}[!ht]
\centering
	\includegraphics[width=0.9\textwidth]{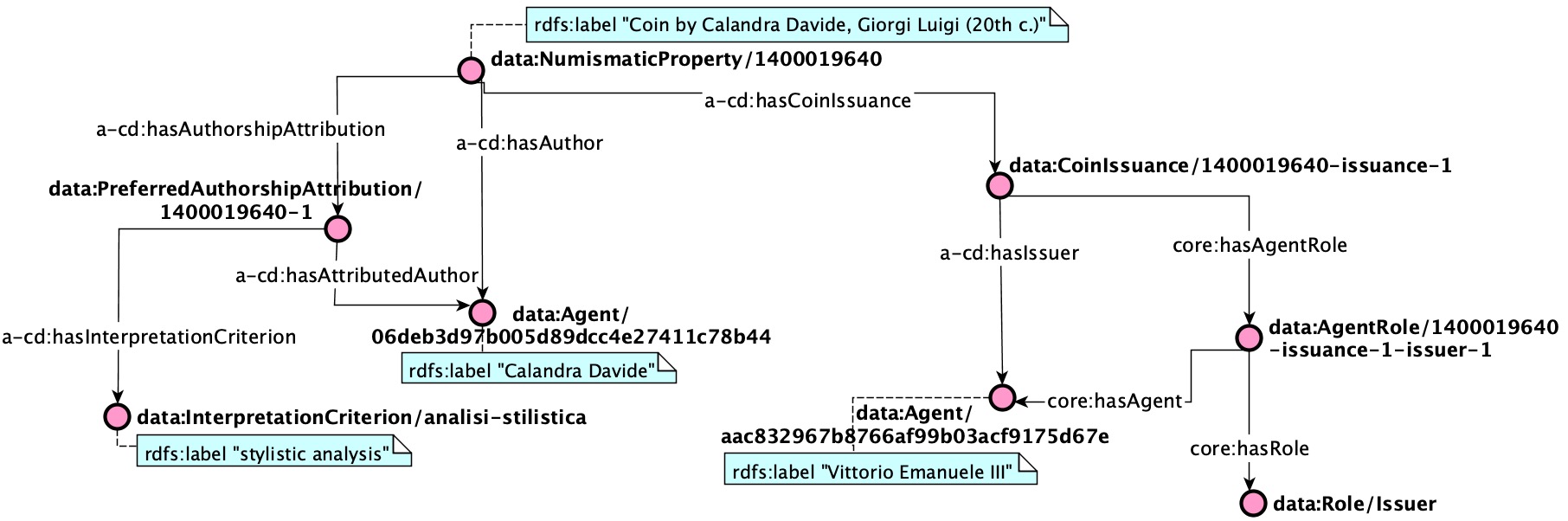}
	\caption{An instance of the model in Figures \ref{img:coin-issuance} and \ref{img:authorship}: a numismatic property involved in a coin issuance and a preferred authorship attribution.}
	\label{img:coin-author-ex}
\end{figure*}

Let us take as an example a coin\footnote{\url{https://w3id.org/arco/resource/NumismaticProperty/1400019640}} (Figure \ref{img:coin-author-ex}) from the 20th century, issued by Victor Emmanuel III of Italy: the issuer is expressed as both an \emph{n-ary} relation (\smalltt{a-cd:AgentRole}), involving the King and the role he played, and as an object property (\smalltt{a-cd:hasIssuer}). Moreover, this coin has two preferred authorship attributions, one of them is depicted in the Figure: the \smalltt{data:\-Pre\-fer\-red\-Au\-thor\-ship\-At\-tri\-bu\-tion/\-1400019640-1} has Calandra Davide as attributed author, which has been attributed based on \smalltt{data:\-In\-ter\-pre\-ta\-tion\-Cri\-te\-rion/\-a\-na\-li\-si-sti\-li\-sti\-ca}, i.e. \enquote{stylistic analysis}. The object property \smalltt{a-cd:\-has\-Au\-thor} works as a shortcut for relating the cultural property to the preferred author(s).

\paragraph{The technical status of a cultural property.}

Another example of situation involving a cultural property is its \textit{technical status}. In this case, a cultural property is related to a set of technical characteristics,
intended as its technical aspects, attributes or qualities.
For instance, \enquote{the archaeological cultural property realised with pottery material and cylindrical in shape}.
Technical status refers to physical features, since it involves characteristics of entities that are either physical objects or physical realisations of information objects.
These characteristics can change over time, thus modifying the technical status of the cultural property: for example, a new survey on an archaeological monument may discover new materials used for its foundation. The temporal validity of a technical status refers to the moment when the characteristics were observed (and recorded in the catalogue record), until when a new condition occurs.

Different technical characteristics of a cultural property can be specified, in order to describe its technical status: the constituting materials (e.g. wood, clay), the employed techniques (e.g. oil-painting, melting), the shape (e.g. square, octagon), the file format for a digital photograph (e.g. \enquote{.gif}, \enquote{.jpeg}), the prevalent colour of a garment, etc. All these concepts (i.e. material, techniques, shape) \emph{classify} the corresponding technical characteristics (i.e. wood, oil-painting, square). The \emph{Classification}\footnote{\label{note:classification}\url{http://www.ontologydesignpatterns.org/cp/owl/classification.owl}} ODP \cite{DBLP:journals/aamas/Gangemi08} defines a classification relation between a concept and an object, which exactly captures this circumstance.

\begin{figure*}[!ht]
\begin{subfigure}{\textwidth}
\centering
	\includegraphics[width=0.95\textwidth]{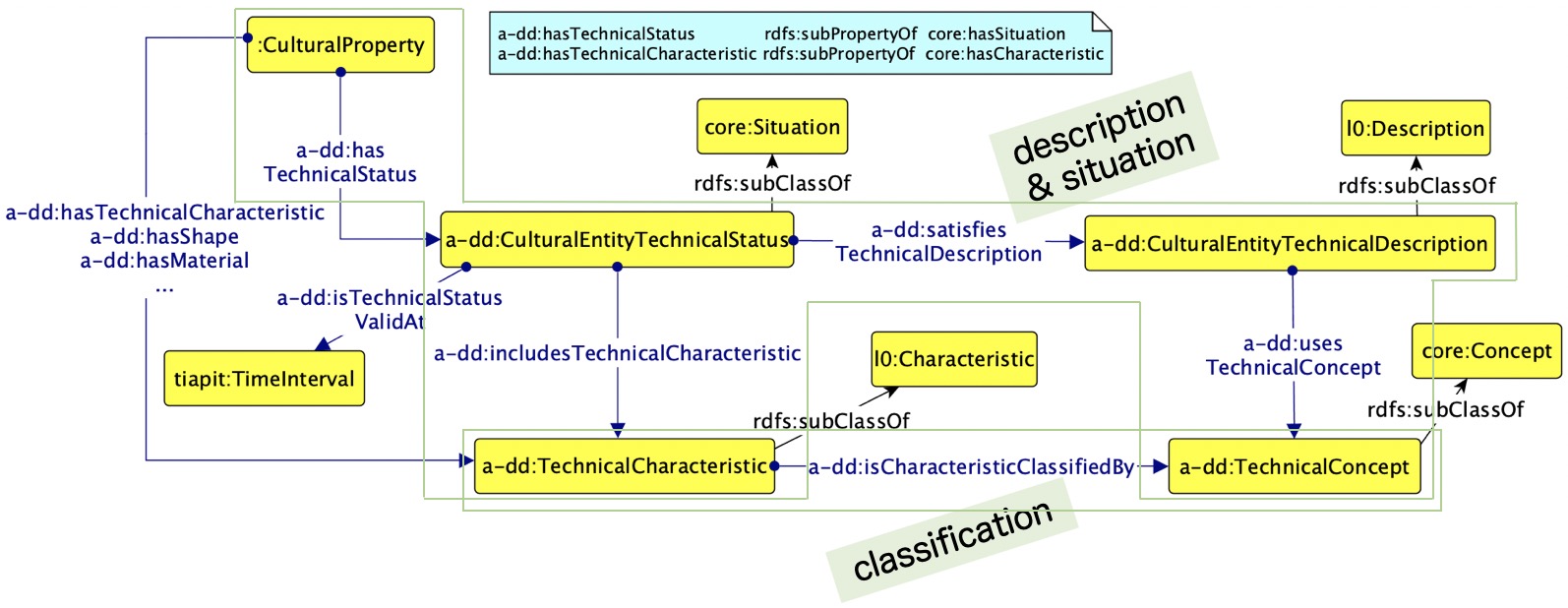}
	\caption{The model for cultural entity technical descriptions and status.}
	\label{img:technical-status-a}
\end{subfigure}
\begin{subfigure}{\textwidth}
\centering
	\includegraphics[width=0.8\textwidth]{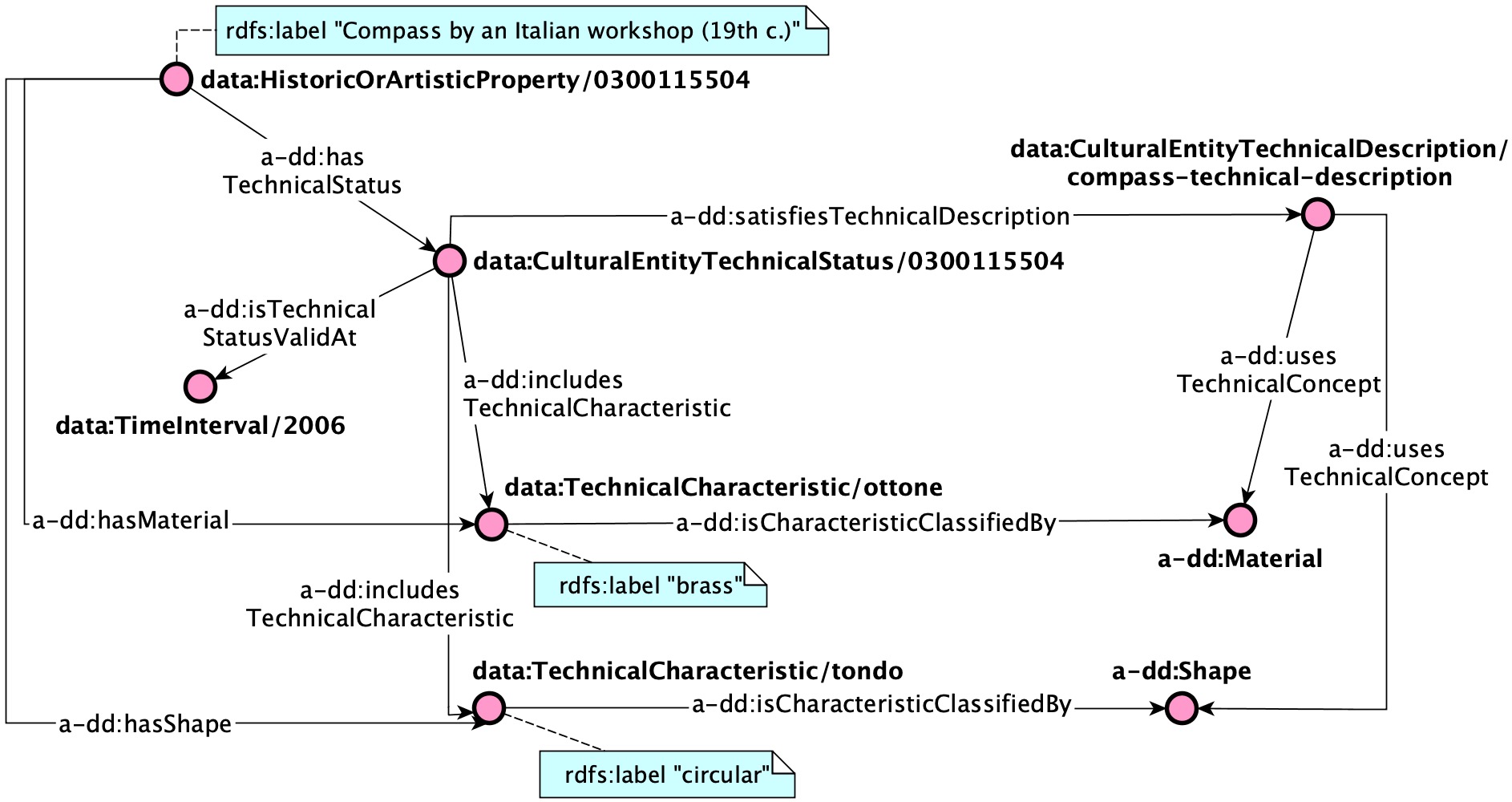}
	\caption{An instance of the models in Figure \ref{img:technical-status-a}, with the technical status of a compass made of brass (material) and circular (shape).}
	\label{img:technical-status-b}
\end{subfigure}
\caption{The D\&S pattern reused and specialised for modelling technical descriptions and status of a cultural entity.} 
\end{figure*}

A specific set of \emph{technical concepts} classifying the \emph{technical characteristics} of a cultural property type (e.g. an artwork), represents a way to conceptualise the technical status of a cultural property, hence they constitute a \emph{technical description} (cf. Section \ref{cdns} for the foundational description pattern reused here). For example, an \emph{artwork technical description} may be defined as the relation between \emph{constituting material}, \emph{employed technique}, and \emph{shape}. We say that such technical description \emph{uses} these concepts. The \emph{technical status} of artwork \emph{A1} could be wood, oil-painting, and square, while the technical status of artwork \emph{A2} could be clay, melting, and octagon. Both technical statuses are expressed according to the \emph{artwork technical description}: we say that they \emph{satisfy} it. The \emph{Description and Situation}\footnote{\url{http://www.ontologydesignpatterns.org/cp/owl/descriptionandsituation.owl}} ODP \cite{DBLP:books/ios/p/GangemiP16} models the satisfaction relation between situations and descriptions, and reuses the \emph{Classification} ODP to model the relation between objects of a situation and concepts of the corresponding description. 
%

Figure \ref{img:technical-status-a} shows how we model the \smalltt{a-dd:\-Cul\-tur\-al\-En\-tity\-Tech\-ni\-cal\-Sta\-tus}, which includes the \smalltt{a-dd:\-Tech\-ni\-cal\-Char\-ac\-ter\-is\-tic}s observed on a cultural property, as a subclass of \smalltt{core:\-Sit\-u\-a\-tion}.
The technical characteristic is modelled as a subclass of \smalltt{l0:\-Char\-ac\-ter\-is\-tic}, which is aligned to \smalltt{d0:\-Char\-ac\-ter\-is\-tic}\footnote{\smalltt{d0:} http://www.ontologydesignpatterns.org/ont/d0.owl\#}, i.e. a union of \smalltt{dul:\-Par\-am\-eter}\footnote{\smalltt{dul:} http://www.ontologydesignpatterns.org/ont/dul/DUL.owl\#}, \smalltt{dul:\-Qua\-li\-ty}, \smalltt{dul:\-Reg\-ion}.
Each characteristic is \emph{classified by} a \smalltt{a-dd:\-Tech\-ni\-cal\-Concept}, e.g. the \smalltt{a-dd:\-Shape}. For the most common values of technical concepts we provide a controlled vocabulary. These concepts are used in the \smalltt{a-dd:\-Cul\-tur\-al\-En\-ti\-ty\-Tech\-ni\-cal\-De\-scrip\-tion}, defined as a subclass of \smalltt{l0:\-De\-scrip\-tion}.

Let us take a compass by an Italian workshop of the 19th century\footnote{\url{https://w3id.org/arco/resource/HistoricOrArtisticProperty/0300115504}} as an example. Figure \ref{img:technical-status-b} represents the situation in which this compass has a technical status, which includes two technical characteristics: \enquote{ottone} (brass) and \enquote{tondo} (circular). The first one is classified by the \smalltt{a-dd:\-Tech\-ni\-cal\-Con\-cept} material, while the second one by the shape. Thus, through the technical status we know that this cultural property is made of brass and is circular.
Those technical characteristics have been observed and recorded in 2006 (\smalltt{a-dd:\-is\-Tech\-ni\-cal\-Sta\-tus\-Valid\-At}).

\subsection{Recurrence in cultural events and in intangible cultural heritage}
\label{sec:recurrent-events}

The involvement of a cultural property in an exhibition during its life cycle would be referred to, in everyday language, as a cultural event.
When we informally refer to the repetition of a cultural event (e.g. different editions of an annual painting award), we use the term \emph{recurrent event} as we are talking about an event that occurs more than once. Actually, we are implicitly referring to a \textit{series} of conceptually unified situations: for example, the Art Biennale\footnote{\label{ref:art-biennale}\url{https://www.labiennale.org/en/art/}} is a series of consecutive situations that can be somehow considered as part of a uniform collection. Cultural properties themselves can be recurrent: an intangible cultural heritage can have regular time intervals between its repetitive occurrences, such as a traditional ceremony related to the \emph{year cycle} (e.g. Carnival).

As these particular series of situations unfold, we can recognise a pattern in their iteration: an exhibition that has different editions over years usually follows a pattern in planning consecutive editions at regular time intervals (e.g. one edition per year). Moreover, it is possible to identify attributes that give all occurrences a unity: a general topic that does not change i.e. contemporary art, a place that host the situation i.e. Venice, etc.

Recurrent situations are usually modelled as a special type of events (cf. Wikidata\footnote{\url{https://www.wikidata.org/wiki/Q15275719}}), while their belonging to a series and the nature of such a unifying entity is neglected in literature or confused with the concept of event (cf. DBpedia resource for Venice Biennale\footnote{\url{http://dbpedia.org/page/Venice_Biennale}}). We believe that modelling both the unitary series of situations, e.g. the \emph{Art Biennale}\footref{ref:art-biennale} intended as something that occurs biennially under certain conditions, and its individual member situations, e.g. the \emph{Art Biennale 2019} intended as a particular edition of the series with a start date and an end date, is important in the CH domain context. We  introduce a new ODP for modelling \emph{Recurrent situation series}\footnote{\url{http://www.ontologydesignpatterns.org/cp/owl/recurrenteventseries.owl}} \cite{Carriero2019WOP}, which reuses other existing ODPs.

\begin{figure*}[!ht]
\begin{subfigure}{\textwidth}
\centering
	\includegraphics[width=0.7\textwidth]{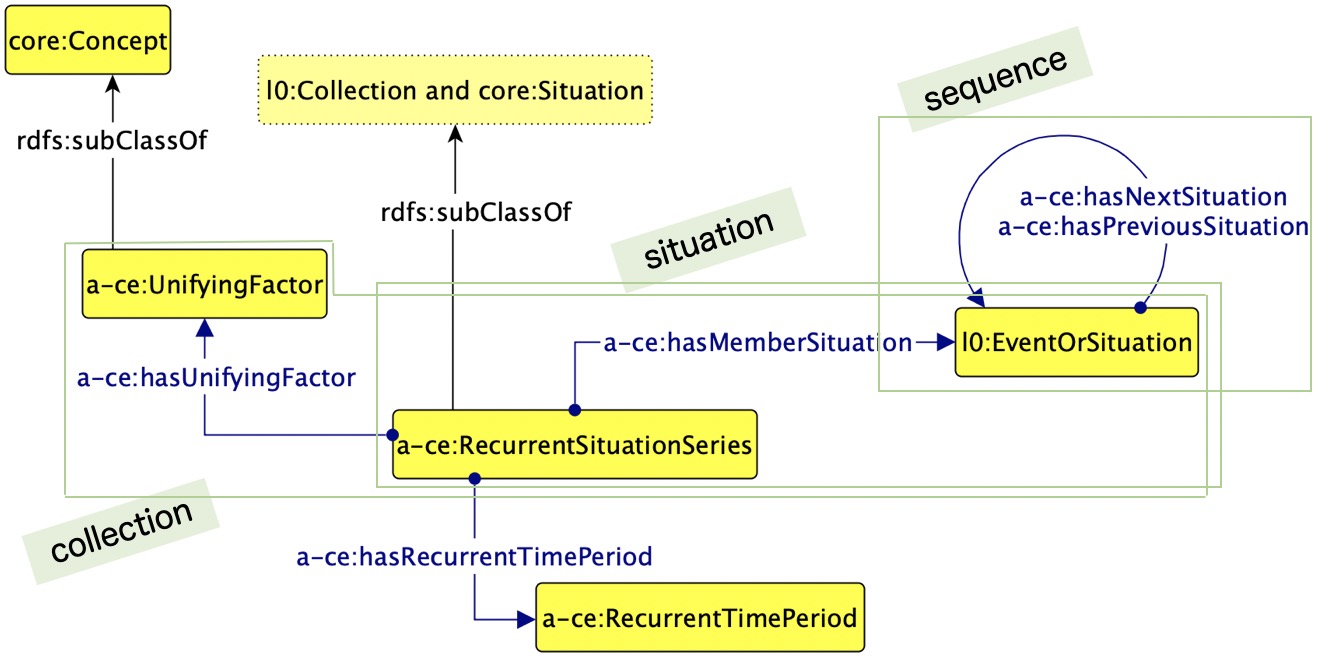}
	\caption{The model for recurrent situation series.}
	\label{img:recurrent-sit-a}
\end{subfigure}
\begin{subfigure}{\textwidth}
\centering
	\includegraphics[width=0.8\textwidth]{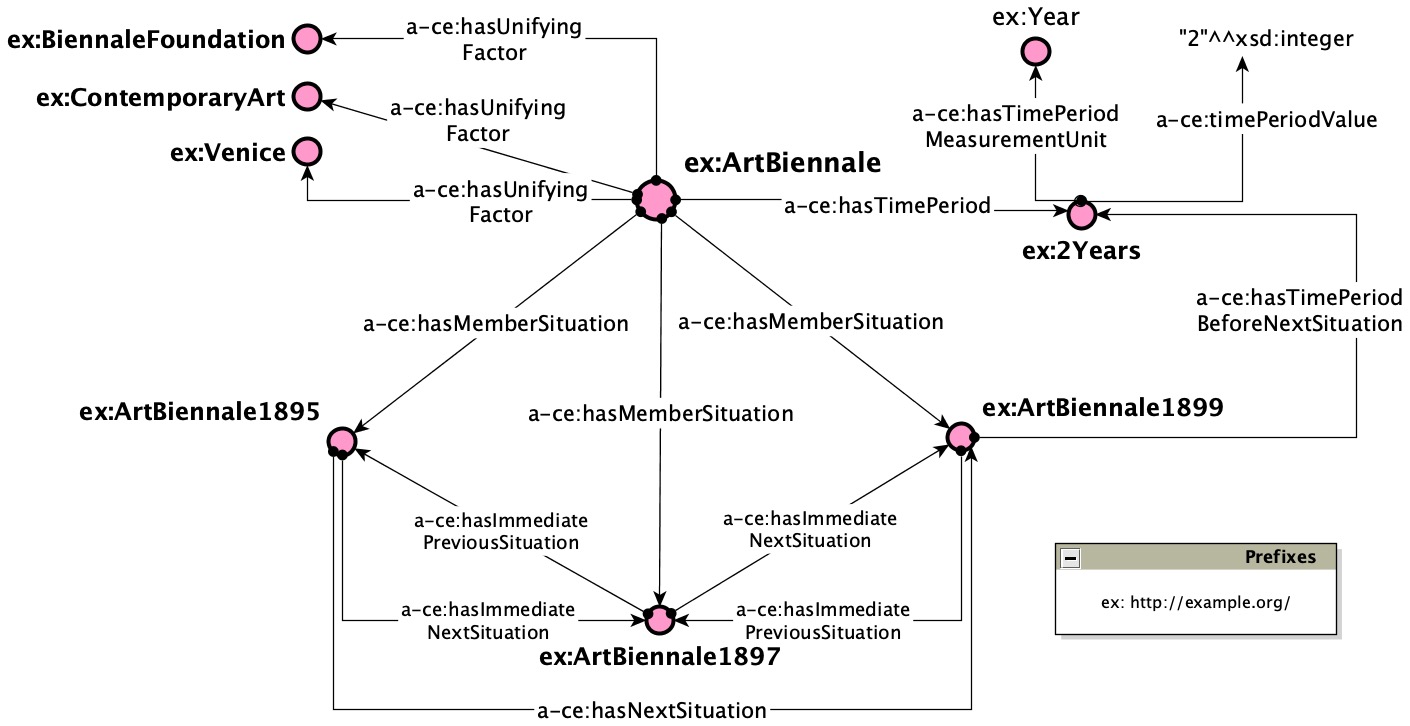}
	\caption{An instance of the model in Figure \ref{img:recurrent-sit-a}, with the 3 first situations member of the Art Biennale.}
	\label{img:recurrent-sit-b}
\end{subfigure}
\caption{The new pattern Recurrent Situation Series as implemented in ArCo.} 
\end{figure*}

We represent, as depicted in Figure \ref{img:recurrent-sit-a}, recurrent situation series (\smalltt{a-ce:\-Re\-cur\-rent\-Sit\-uat\-ion\-Series}) as collections of situations, their members (\smalltt{a-ce:\-has\-Mem\-ber\-Sit\-uat\-ion}). Member situations share at least one common property (e.g. the topic) and are conceptually unified by \emph{unifying factors} (\smalltt{a-ce:\-has\-U\-ni\-fy\-ing\-Fac\-tor}) that characterise the series. At the same time, a recurrent situation series is a situation, since it provides a relational context to all the member situations. Each member situation has its own time interval and is put in a \emph{sequence} that relates it to the other member events of the same series (e.g. \smalltt{a-ce:\-has\-Next\-Sit\-uat\-ion}). The time period that elapses between member situations is (approximately) regular and is an attribute of the series (\smalltt{a-ce:\-has\-Re\-cur\-rent\-Time\-Pe\-ri\-od}).

In Figure \ref{img:recurrent-sit-b} we can see an instance of this pattern. The \smalltt{ex:ArtBiennale} (\emph{Biennale d'arte di Venezia}) is a contemporary visual art exhibition, whose member situations are conceptually unified by: the Biennale Foundation as organiser, Venice as place, the promotion of new contemporary art trends as mission. The time period between two consecutive situations of the Art Biennale series is of about \smalltt{ex:2Years}. The first three situations member of the series are in a temporal sequence: for example, the \smalltt{ex:ArtBiennale1895} has as immediate next situation the \smalltt{ex:ArtBiennale1897}, while the object property \smalltt{a-cd:hasImmediatePreviousSituation} relates the 1899 edition to the one held in 1897.

\subsection{Direct and indirect reuse of patterns}
\label{sec:reuse}
Reusing ontologies and ODPs can be done by following two main different approaches, depending on the conditions and requirements of a project: direct and indirect reuse~\cite{DBLP:conf/er/PresuttiLNGPA16}.

\paragraph{Direct reuse.}
This approach consists in directly embedding individual entities or importing implementations of ODPs or other ontologies in the local ontology, thus making it highly dependent on them. This may jeopardize the stability of the ontology if the evolution of the imported ontologies is outside the control or monitoring of the team/organisation that is reusing them: even small changes in the reused ontologies could introduce inconsistencies in the local one, contrary to its original requirements. For this reason, ArCo directly reuses only two ontologies that are considered reference standards by the Italian Government and the evolving process of which is relatively slow and systematised, and involves ArCo's team. These ontologies are Cultural-ON\footnote{\url{http://dati.beniculturali.it/cis/}}, which is also directly maintained by MiBAC, and OntoPiA ontology network\footref{ref:ontopia}, which is recommended as a standard for open data of the Italian Public Administration (and that now includes ArCo). Examples of ontology modules directly reused from OntoPiA are: \emph{Address and Location}\footnote{\url{https://w3id.org/italia/onto/CLV}}, \emph{Time}\footnote{\url{https://w3id.org/italia/onto/TI}}, etc.

\paragraph{Indirect reuse.}
In this approach, relevant entities and patterns from external ontologies are used as \emph{templates}, by reproducing them in the local ontology and providing possible extensions. Alignments axioms (such as \smalltt{rdfs:\-sub\-Class\-Of} and \smalltt{owl:\-e\-quiv\-a\-lent\-Class}) are introduced to support interoperability with other ontologies and make it evident what parts have been reused. This practice decreases the dependency on external ontologies, and is widely adopted in ArCo. 

\subsection{Annotating reused ODPs} 
\begin{figure*}[!ht]
\begin{subfigure}{\textwidth}
\centering
	\includegraphics[width=0.9\textwidth]{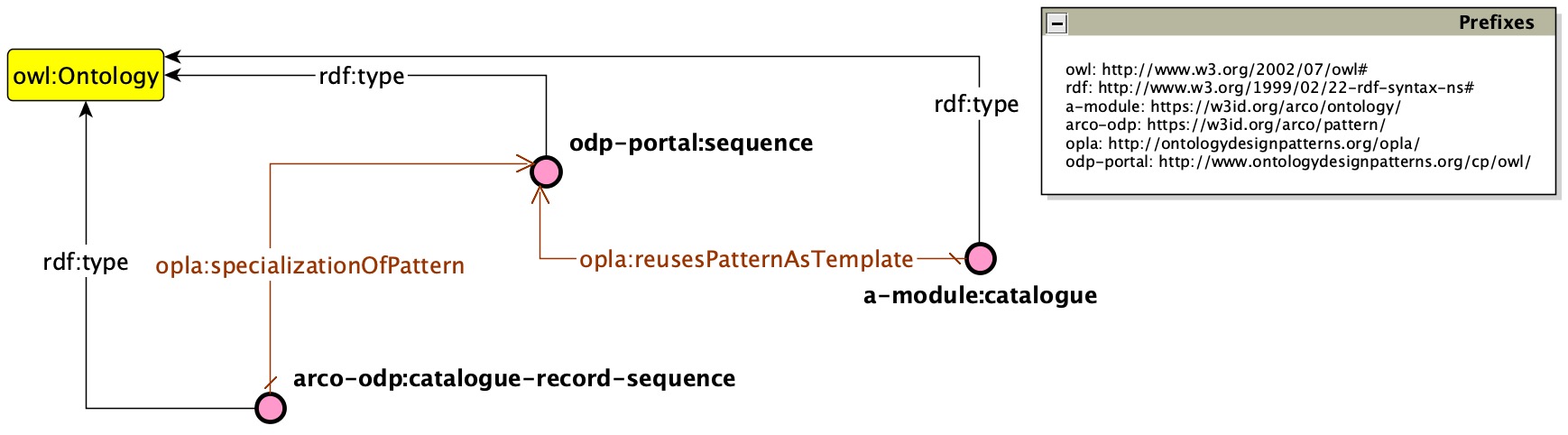}
	\caption{The annotation property \smalltt{opla:\-re\-uses\-Pat\-tern\-As\-Tem\-plate} relates the \emph{catalogue} module to the Sequence ODP that is reused over the module. The annotation property \smalltt{opla:\-spe\-cializa\-tion\-Of\-Pat\-tern} relates the pattern \emph{catalogue record sequence} implemented in the module to the \emph{Sequence} ODP that has been specialised.}
	\label{img:opla-example-onts}
\end{subfigure}
\begin{subfigure}{\textwidth}
\centering
	\includegraphics[width=0.9\textwidth]{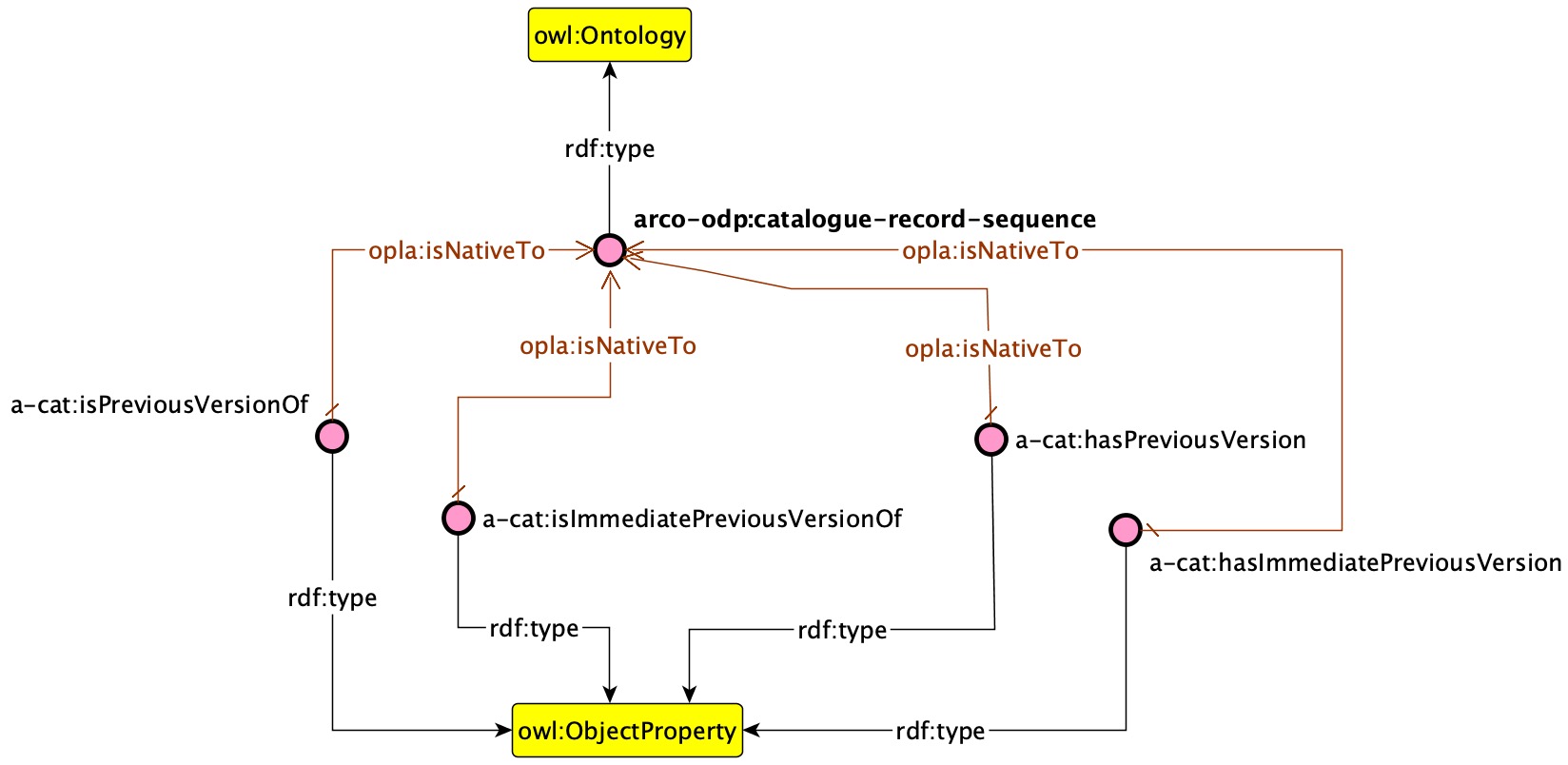}
	\caption{The annotation property \smalltt{opla:\-is\-Nat\-ive\-To} relates the object properties that belong to the \emph{catalogue record sequence} ODP to the ODP itself.}
	\label{img:opla-example-objprop}
\end{subfigure}
\caption{An example of a reused ODP annotated with OPLa ontology.}
\label{img:opla-example}
\end{figure*}

Annotating reused patterns supports the identification of ontology alignments, which is a tedious, non-trivial task. In fact, ODP annotations may ease the process to understand and explore an ontology. These assumptions have driven the development of the simple Ontology Pattern Language annotation (OPLa)\footnote{\smalltt{opla:} http://ontologydesignpatterns.org/opla/}~\cite{DBLP:conf/semweb/HitzlerGJKP17}. All (re)used ODPs in ArCo are annotated with OPLa. For instance, OPLa allows to express that a pattern in a local ontology is a specialisation or a generalisation of a more general ODP. Let us consider the \emph{catalogue} module (see Figure \ref{img:opla-example}). Over this module, two ODPs from the \emph{ODP portal}\footnote{\label{ref:odp-portal}\url{http://ontologydesignpatterns.org/}} have been (indirectly) reused: the module is therefore annotated with the property \smalltt{opla:\-reus\-es\-Pat\-tern\-As\-Tem\-plate} for representing the reuse of \emph{Sequence}\footref{ref:odp-sequence} (see Figure \ref{img:opla-example-onts}) and \emph{Classification}\footref{note:classification} ODPs. For example, the pattern \emph{Catalogue Record Sequence}\footnote{\url{https://w3id.org/arco/pattern/catalogue-record-sequence/}} is a specialisation of the pattern \emph{Sequence}, since it represents a sequence of catalogue records, hence it is annotated with the annotation property \smalltt{opla:\-spe\-cial\-i\-za\-tion\-Of\-Pat\-tern} (see Figure \ref{img:opla-example-onts}). For expressing that specific properties (e.g. \smalltt{a-cat:\-is\-Pre\-vious\-Vers\-ion\-Of}, \smalltt{a-cat:\-has\-Im\-med\-iate\-Pre\-vious\-Ver\-sion}, etc.) implemented in the catalogue module, belong to this ODP, the annotation property \smalltt{opla:\-is\-Nat\-ive\-To} is used, as in Figure \ref{img:opla-example-objprop}. 

%
%
\section{A formal evaluation of ArCo}
\label{sec:evaluation}
ArCo is evaluated along different dimensions: functional, logical, and structural dimensions as identified by~\cite{Gangemi2006}. The functional dimension is related to the intended use of a given knowledge graph (KG) and of its components, i.e. their function in a context. It is a core dimension for ontology testing. In fact, it allows ontology designers to assess the ability of an ontology to address requirements and cover the domain. The logical dimension measures whether an ontology can be successfully processed by a reasoner (inference engine, classifier, etc.).
Finally, the structural dimension of a KG\footnote{The authors of~\cite{Gangemi2006} refer to ontologies in their analysis. In the scope of this paper we generalise their results to knowledge graphs, since we also compute the distribution of the instances across classes.} focuses on its topological properties measured by means of context-free metrics that leverage its graph-based representation. The functional and logical dimensions are utmost important for assessing the quality of a KG. Nevertheless, the analysis of the structural dimension provides insights on design choices by means of indicators that might suggest quality weaknesses or strength.

\subsection{Functional and logical dimensions}
\label{sec:testing}
{\bf CQ verification, inference verification and error provocation.}
The logical dimension is addressed by running a reasoner on ArCo KG. This is a necessary step, but not sufficient. We regularly run a reasoner, but we perform additional tests at each iteration of the design methodology, i.e. every time new requirements are selected to be addressed. In doing so, we adopt the testing methodology described in \cite{Blomqvist2012}. This methodology focuses on evaluating the appropriateness of an ontology against its requirements intended as the ontological commitment expressed by means of CQs and domain constraints, i.e. functional dimension.
User stories are translated into one or more CQs and general constraints during the design phase. To each CQ and to each constraint corresponds a unit test, which contributes, when run, to validate the ontology. Thus, the core element of each unit test is either a CQ or a general constraint (e.g. disjointness axiom).
Accordingly, three different approaches are followed in the testing activity: \textit{CQ verification, inference verification, error provocation}.

\paragraph{CQ verification.} This approach consists in testing whether the ontology vocabulary allows to convert a CQ, reflecting an ontology requirement, to a SPARQL query. Let us consider the CQ ``When was a cultural property created, and which is the interpretation criterion which the dating is based on?'', which ArCo ontologies should answer, based on the collected requirements. The testing team starts verifying the completeness of the ontologies by translating this question from natural language to SPARQL, using classes and properties defined in ArCo ontologies (e.g. the entities defined for representing the date of creation of a cultural property). This step allows to detect any missing concept or gap in the vocabulary, e.g. whether the concept of interpretation criterion has been modeled. If the CQ can be successfully converted, the testers run the resulting SPARQL query over the actual RDF data or, when missing, over test data generated using Fuseki\footnote{\url{https://jena.apache.org/documentation/fuseki2/}}, and complete the test by comparing the expected result (i.e. the output they expect from the query) to the actual result.
 

\paragraph{Inference verification.} This step focuses on checking the inferences over the ontologies, by comparing the expected inferences to the actual ones. Let us consider a complex cultural property, which is a cultural property with one or more components, as proper parts. If a \smalltt{:Com\-plex\-Cul\-tur\-al\-Prop\-er\-ty} is defined as a  \smalltt{:Cul\-tur\-al\-Prop\-er\-ty} that has one or more \smalltt{:Cul\-tur\-al\-Prop\-er\-ty\-Com\-po\-nent}s, an axiom stating that a \smalltt{:Cul\-tur\-al\-Prop\-er\-ty} has a \smalltt{:Cul\-tur\-al\-Prop\-er\-ty\-Com\-po\-nent} would suffice to infer that the property is complex, even if it is not explicitly asserted.
For comparing this expected inference with the actual one, the testing team injects the necessary data in the knowledge graph -- e.g. an instance of the class \smalltt{:Cul\-tur\-al\-Prop\-er\-ty} related to an instance of the class \smalltt{:Cul\-tur\-al\-Prop\-er\-ty\-Com\-po\-nent} through the object property \smalltt{:has\-Cul\-tur\-al\-Prop\-er\-ty\-Com\-po\-nent} -- and runs the reasoner. If the reasoner does not infer that the first instance is \smalltt{rdf:\-type} \smalltt{:Com\-plex\-Cul\-tur\-al\-Pro\-per\-ty}, this means that the appropriate axiom is missing from the ontology, i.e. an equivalent axiom between \smalltt{:Com\-plex\-Cul\-tur\-al\-Pro\-per\-ty} and (\smalltt{:has\-Cul\-tur\-al\-Pro\-per\-ty\-Com\-po\-nent} some \smalltt{:Cul\-tur\-al\-Pro\-per\-ty\-Com\-po\-nent}).

\paragraph{Error provocation.} This third testing activity is intended to ``stress'' the knowledge graph by injecting inconsistent data that violate our requirements.

For instance, the entities representing the concepts of dating and attributing an author to a cultural property should be disjoint, since there can be no individuals that are dating and authorship attributions at the same time. For validating the ontology regarding this requirement, the testers inject in the KG an individual belonging to both \smalltt{a-cd:\-Au\-thor\-ship\-At\-tri\-bu\-tion} and \smalltt{a-cd:\-Dat\-ing} classes, and run the reasoner. The expected result is an inconsistency: if this is not detected by the reasoner, it means that the appropriate (disjointness) axiom is missing.

\paragraph{Refactoring and integration.}
Problems spotted during the testing phase are passed back to the design team as issues. The design team refactors the modules and updates the ontology after performing a consistency checking. The result of this step is validated again by the testing team before including the model in the next release.

\begin{figure}[ht!] 
\centering
\includegraphics[scale=0.27]{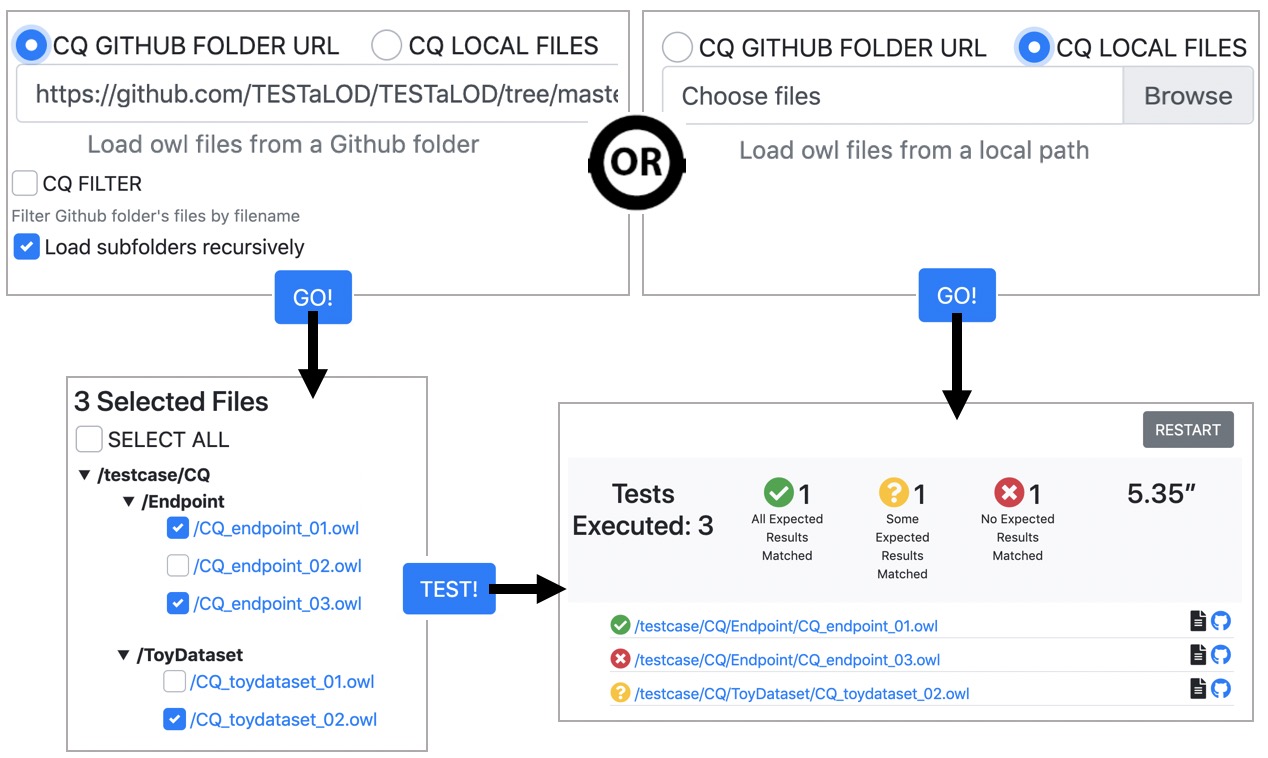}	\caption{Workflow implemented by TESTaLOD based on the user interface.}
	\label{img:testalod}
\end{figure}

\paragraph{Evaluation tool.}
We rely on TESTaLOD~\cite{Carriero2019TESTaLOD} for dealing with testing activities associated with CQ verification, inference verification, error provocation. TESTaLOD is a tool designed and implemented in the context of ArCo's project for supporting not only the testing team of ArCo KG, but in general any testing team of projects adopting XD methodology or other test-driven methodologies. TESTaLOD is developed as a Web application\footnote{Demo: \url{https://w3id.org/testalod} \\ Source code: \url{https://github.com/TESTaLOD/TESTaLOD}} that provides a knowledge graph testing toolbox: as presented in Figure \ref{img:testalod}, it implements a two-step workflow, allowing the user to select and automatically run defined test cases aiming at verifying CQs. The test cases are OWL files, and are modelled by using the TestCase OWL meta model introduced in~\cite{Blomqvist2016}, thus containing: a Competency Question and its corresponding SPARQL query, the expected (correct) result and data sample. The test cases can be either retrieved from a GitHub repository or uploaded from a local file system. Once the tests have been automatically executed, the expected result is compared to the actual result, and three possible outputs can be displayed to the user: successful, partially successful, unsuccessful.

Let us consider as an example the competency question \enquote{Which archival set (fonds, series, subseries) a cultural property is member of?}, and that we want to verify if our ontology models information on membership of cultural properties to archival record sets. The test case for running this test will be an OWL file, annotated with the following properties\footnote{\smalltt{test:} \url{http://www.ontologydesignpatterns.org/schemas/testannotationschema.owl\#}\label{fn:testannotationschema}}: \smalltt{test:\-has\-CQ} has the competency question expressed in natural language as a value; \smalltt{test:\-has\-SPARQL\-Query\-Unit\-Test} the translation of the CQ to SPARQL, using the ontology entities; \smalltt{test:\-has\-Input\-Test\-Data} points out the test data used as input for running the test; \smalltt{test:\-has\-Ex\-pec\-ted\-Re\-sult} stores a set of expected results of running the query over a certain set of test data; \smalltt{test:\-has\-Actual\-Re\-sult} stores the actual outcome of a test run. Other properties are used in order to annotate who run the test and when.

In order to allow TESTaLOD to automatically run this test, two new annotation properties\footnote{\smalltt{testalod:} \url{https://raw.githubusercontent.com/TESTaLOD/TESTaLOD/master/ontology/testalod.owl\#}} have been defined. \smalltt{testalod:\-has\-Input\-Test\-Data\-Cat\-e\-go\-ry} annotates if the input data are available at a SPARQL endpoint (\smalltt{testalod:\-SPARQL\-end\-point}) or in a file with test data (\smalltt{testalod:\-Toy\-Data\-set}); \smalltt{testalod:\-has\-Input\-Test\-Data\-Uri} annotates the URI of the SPARQL endpoint or the file, which is used by TESTaLOD to run the query.

{\bf Terminological coverage.}
Additionally, we further analyse the functional dimension by setting up an experiment aimed at assessing ArCo ontologies with regards to their ability in capturing and conveying domain-specific terminology. This is of utmost important to assess whether ArCo addresses its intended use, i.e. compliance to expertise. Inasmuch as only measuring the terminological coverage for ArCo ontologies might not be informative, we set up this experiment as a comparative analysis. For the comparison we select EDM\footnote{\label{ref:edm}\url{https://pro.europeana.eu/page/edm-documentation}} and CIDOC CRM\footnote{\label{ref:cidoc}\url{http://www.cidoc-crm.org/}} as they are two well known and widely used ontologies in the same domain of ArCo. The terminological coverage is modelled as an ontology alignment problem between the vocabulary that represent the domain-specific terminology and the target ontology (ArCo, EDM, and CIDOC CRM, respectively). The vocabulary is automatically extracted with Rapid Automatic Keyword Extraction~\cite{Rose2010} (RAKE) from a corpus composed of the ICCD cataloguing standards, their associated guidelines, and ArCo's competency questions. The resulting vocabulary counts of 55 terms and is publicly available as RDF\footnote{\url{https://doi.org/10.6084/m9.figshare.7926599.v1}}.

{\bf Experiments execution and results.} The results recorded are the following.
\label{sec:testalod}
\paragraph{Inference and CQ verification, and error provocation.}
We define 18 test cases for inference verification, 29 test cases for error provocation, and 55 test cases for competency question verification. Each test case is publicly available on GitHub\footnote{\url{https://github.com/ICCD-MiBACT/ArCo/tree/master/ArCo-release/test}} and it is modelled by using the {\em testannotationschema}\footref{fn:testannotationschema} ontology. For both inference verification and error provocation we define data samples to use with the HermiT reasoner\footnote{\url{http://www.hermit-reasoner.com/}} for checking (i) the soundness of ArCo ontologies in inferring correct axioms (i.e. inference verification) and (ii) producing expected {\em in vitro} logical inconsistencies (i.e. error provocation). We rely on automatic reasoning as inference verification and error provocation provide an indication about the computational integrity and efficiency. \cite{Gangemi2006} defines computational integrity and efficiency as the property that prospects an ontology that can be successfully processed by a reasoner. We use TESTaLOD for competency question verification by providing the corresponding test cases as input to the tool. 
The results obtained by using TESTaLOD record all test cases as successful.

\begin{figure}[!ht]
\centering
	\includegraphics[width=0.45\textwidth]{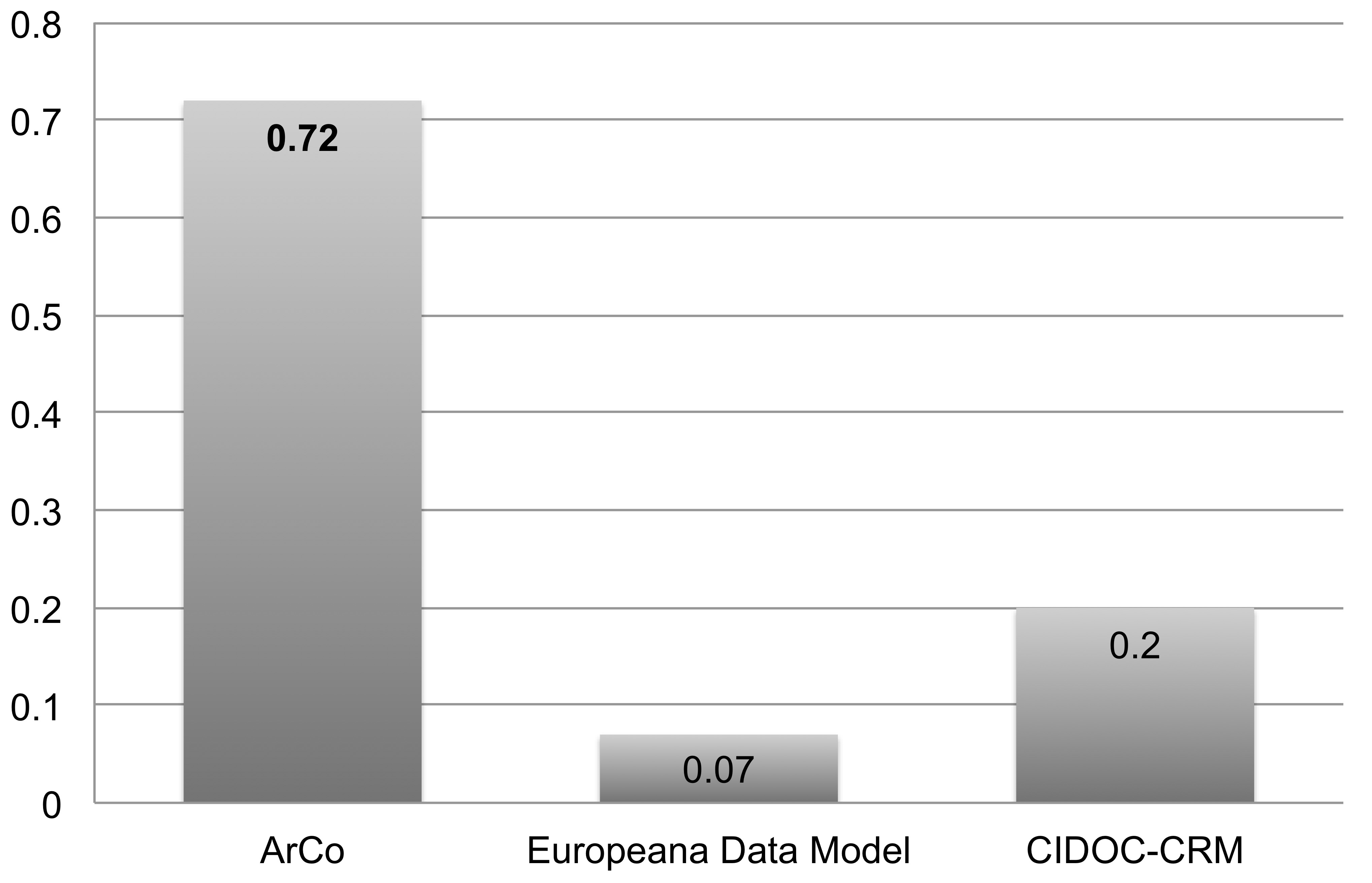}
	\caption{The terminological coverage as recorded for ArCo, EDM, and CIDOC CRM.}
	\label{img:terminological-coverage}
\end{figure}

\paragraph{Terminological coverage.}
The ontology alignment is computed with Silk~\cite{Volz2009} by using the {\em substring} metric with 0.5 as threshold. The alignment with the vocabulary is executed three times, i.e. once for each ontology involved in the comparison. 
The configuration files provided as input to Silk are available on FigShare\footnote{The link specification files for ArCo, CIDOC CRM, and EDM are published with the DOIs \url{https://doi.org/10.6084/m9.figshare.7925555.v1}, \url{https://doi.org/10.6084/m9.figshare.7925573}, and \url{https://doi.org/10.6084/m9.figshare.7925867}, respectively.}.
Figure~\ref{img:terminological-coverage} reports the results of the terminological coverage for ArCo, EDM, and CIDOC CRM.


\subsection{Structural dimension}
For assessing the structural dimension of ArCo KG we use different metrics that have been defined and used in literature~\cite{Tartir2010,Yao2005,Gangemi2006,Orme2006,Schlicht06,dAquin09,Khan16}. First, we compute base metrics that record quantitative aspects of ArCo knowledge graph: classes and their instances, properties, axioms, etc. Then, we compute schema and graph metrics aimed at assessing (i) the richness, width, depth, and inheritance at the schema level and (ii) the cohesion, coupling, multihierarchical degree, and extensional coverage of the ontologies. Those parameters are used for understanding the quality of ArCo expressed in terms of (i) flexibility, (ii) transparency, (iii) cognitive ergonomics, and (iv) compliance to expertise. These quality properties have been defined in~\cite{Gangemi2006}: (i) flexibility is the property of an ontology to be easily adapted to multiple views; (ii) transparency is the property of an ontology to be analysed in detail, with a rich formalisation of conceptual choices and motivation; (iii) cognitive ergonomics is the property of an ontology to be easily understood, manipulated, and exploited by its consumers; and (iv) compliance to expertise is the property of an ontology to be compliant with the knowledge it is supposed to model.

Table~\ref{tab:arco-base-om}\footnote{The number of triples and individuals for EDM were retrieved by querying the SPARQL endpoint of Europeana (i.e. \url{http://sparql.europeana.eu/}) on June 1st 2020.} and Table~\ref{tab:arco-sl-om} describe the metrics used along with their corresponding results as recorded for ArCo, CIDOC CRM, and EDM\footnote{We use the CIDOC CRM v6.2.1 and EDM v5.2.4.}. The results for ArCo KG have been reported for three of its different versions resulting by as many iterations of the design methodology. That is, ArCo v0.1\footnote{Available at \url{http://doi.org/10.5281/zenodo.3872004}.}, which is the first development release, ArCo v0.5\footnote{Available at \url{http://doi.org/10.5281/zenodo.2630447}.}, which is an intermediate development release, and ArCo v1.0\footnote{Available at \url{http://doi.org/10.5281/zenodo.3242580}.}, which is the latest and current stable release. The comparison of the different versions of ArCo provides indicators about the structural evolution of the knowledge graph. Instead, the structural analysis of CIDOC CRM and EDM provides a comparative grid that allows us to assess the indicators computed for ArCo by means of comparison. Table~\ref{tab:arco-sl-om} also reports, for the same ontologies, the quality properties that the metrics are an indicator of. We use the association between metrics and quality property defined by~\cite{Gangemi2006}. The metrics are computed by using OntoMetrics\footnote{\url{https://ontometrics.informatik.uni-rostock.de/ontologymetrics/index.jsp}}, a web-based tool aimed at computing statistics about an ontology. 

\begin{table*}[h!]
\begin{center}
	\caption{Comparison of base knowledge graph metrics as computed for ArCo v0.1, ArCo v0.5, ArCo v1.0, CIDOC-CRM, and EDM, respectively. ArCo v1.0 is the latest release of the knowledge graph.}
	\label{tab:arco-base-om}
	\resizebox{0.85\textwidth}{!}{ 
	\begin{tabular}{p{1.8cm}|p{5.5cm}||c|c|c|c|c}
		{ \bf Metric } & { \bf Description } & {\bf ArCo 0.1} & {\bf ArCo 0.5} & {\bf ArCo 1.0}& {\bf CIDOC-CRM} & {\bf EDM}\\\hline 
		\# of axioms & The total number of axioms defined for classes, properties, datatype definitions, assertions and annotations. & 715 & 9,564 & 13,792 & 3,503 & 299 \\\hline
		\# of logical axioms & The axioms which affect the logical meaning the ontology network. & 180 & 2,210 & 3,416 & 830 & 130 \\\hline
		\# of classes & The total number of classes defined in the ontology network. & 54 & 329 & 340 & 84 & 41 \\\hline
		\# of object properties & The total number of object properties defined in the ontology network. & 38 & 332 & 616 & 275 & 51 \\\hline
		\# of datatype properties & The total number of datatype properties defined in the ontology network. & 8 & 153 & 154 & 12 & 12 \\\hline
		\# of annotation assertions & The total number of annotations in the ontology network. & 429 & 6,357 & 8,734 & 2,589 & 125 \\\hline
		DL expressivity & The description logics expressivity of the ontology (network). & SROIF(D) & SROIQ(D) & SROIQ(D) & ALH(D) & ALCHIN(D)\\\hline
		\# of individuals & The total number of individuals instantiated in the knowledge graph. & 6,656,408 & 22,651,078 & 20,030,941 & n.a. & 415,410,190 \\\hline
		\# of triples & The total number of triples available in the knowledge graph. & 35,993,563 & 169,147,193 & 172,580,211 & n.a. & 2,836,270,332
		\\\hline
	\end{tabular}
	}
	\end{center}
\end{table*}

\begin{table*}[htp]
\begin{center}
	\caption{Schema and graph metrics with corresponding quality properties addressed and values recorded. Values are reported for ArCo v0.1, ArCo v0.5, ArCo v1.0, CIDOC-CRM, and EDM. ArCo v1.0 is the latest release of the knowledge graph.}
	\label{tab:arco-sl-om}
	\resizebox{.9\textwidth}{!}{
	\begin{tabular}{p{1.5cm}|p{9.5cm}|p{1.7cm}||c|c|c|c|c}
		{ \bf Metric } & { \bf Description } & {\bf Quality property} & {\bf ArCo 0.1} & {\bf ArCo 0.5} & {\bf ArCo 1.0} & {\bf CIDOC-CRM} & {\bf EDM} \\\hline 
		Relationship Richness & The ratio between non-inheritance relations and the total number of relations defined in the ontology as proposed by~\cite{Tartir2010}. Inheritance relations are \smalltt{rdfs:subClassOf} axioms. Values range from 0 (i.e. the ontology contains inheritance relationships only) to 1 (i.e. the ontology contains non-inheritance relationships only). & Transparency & 0.43 & 0.34 & 0.44 & 0.74 & 0.84 \\\hline
		Inheritance Richness & The average number of subclasses per class computed as proposed by~\cite{Tartir2010}. Inheritance Richness (IR) is expressed on ordinal scale. Its values should be interpreted relatively to the number of classes. If IR is much smaller than the number of classes, then the value is low. On the contrary, if IR tends to equalise the number of classes, the value is high. A low value indicates a deep (or vertical) ontology, while a high value indicates a shallow (or horizontal) ontology. & Transparency & 1.1 & 2.9 & 2.48 & 1.17 & 0.32 \\\hline
		Axiom/class ratio & The ratio between axioms and classes computed as the average amount of axioms per class. Its values should be interpreted relatively to the number of classes and axioms. If the ratio is much smaller than the number of classes, then the value is low. On the contrary, if the ratio is much greater than the number of classes, the the value is high. Low values (i.e. $\sim0$) indicate poorly axiomatised ontologies. On the contrary, higher values indicate better axiomatisations. Extremely high values might indicate over axiomatisation. & Transparency & 13.24 & 29.7 & 39.55 & 41.7 & 7.3 \\\hline
		Class/property ratio & The ratio between the number of classes and the number of properties. Typically good values are in the range $[0.3, 0.8]$ indicating a sufficient number of properties connecting things with other things (i.e. object properties) and values (datatype properties). Low values (i.e $\sim0$) indicate an ontology with many properties connecting few concepts. On the contrary, high values indicate an ontology with many concepts connected by few properties. Nevertheless, the interpretation of the ratio always depends of the ontology size. & Cognitive ergonomics & 0.52 & 0.31 & 0.44 & 0.23 & 0.5 \\\hline
		NoR & The number of root classes as defined by~\cite{Yao2005}. A root class is a class that is not subclass of any other class in the ontology. NoR values are on ordinal scale and provide an indication of cohesion, i.e. the degree of relatedness between the different ontological entities. The interpretation of NoR values depends on the number of classes in the ontology. For example, 8 as NoR value might be low or high if the number of classes is 300 or 10, respectively. & Flexibility, Transparency & 11 & 17 & 16 & 1 & 31 \\\hline
		NoL & The number of leaf classes as defined by~\cite{Yao2005}. A leaf class is a class that has no sub-class in the ontology. NoL values are on ordinal scale and provide an indicator of cohesion, i.e. the degree of relatedness between the different ontological entities. Again, the interpretation of NoL values depends on the number of classes in the ontology. For example, 8 as NoL value might be low or high if the number of classes is 300 or 10, respectively. & Flexibility, Transparency & 44 & 270 & 277 & 48 & 34 \\\hline
    	NoC & The number of external classes as defined by~\cite{Orme2006}. An external class is a class defined in a different ontology. Values for NoC are on ordinal scale. A low value of NoC suggests self-containment and semantic independence of an ontology. On the contrary, a high value suggests strong semantic dependency of an ontology with concepts defined in external ontologies. As for other metrics on ordinal scale the interpretation of good or negative NoC values is relative. For example, if the NoC is comparable to the number of internal classes then self-containment and semantic independence might not be guaranteed. In fact, a large portion of the ontology relies on concepts defined elsewhere. Accordingly, a change in an external ontology might affect the intended semantics deeply. & Flexibility, Transparency & 11 & 35 & 38 & 0 & 3 \\\hline
		Average breadth & The average breadth~\cite{Gangemi2006} computed on the graph whose nodes are ontology classes and edges are \smalltt{rdfs:subClassOf} axioms. The metric suggests the degree of horizontal modelling (i.e. flatness) of the hierarchies of an ontology. Values are on ordinal scale. The value should be interpreted relatively to the number of classes. For example, average breadth values of 10 and 100 in an ontology consisting of 600 classes are low and high, respectively. & Cognitive ergonomics & 5 & 5.7 & 5.75 & 2.57 & 6.0 \\\hline
		Max breadth & The maximal cardinality recorded on ordinal scale over the graph constructed as for the average breadth~\cite{Gangemi2006}. The interpretation of max breadth is similar to that suggested for the average breadth. & Cognitive ergonomics & 30 & 35 & 34 & 10 & 31 \\\hline
		ADIT-LN & It records the average depth of the graph constructed as for the average breadth. The average is computed as the sum of the depth of all paths divided by the total number of paths~\cite{Yao2005}. ADIT-LN values are on ordinal scale and are indicators of cohesion. The interpretation of the values depends on the size of the ontology. Accordingly, low values occur when ADIT-LN is significantly lower then the number of classes. On the contrary, high values occur when the difference between ADIT-LN and the number of classes is less significant. & Transparency, Cognitive ergonomics & 2.45 & 3.23 & 3.93 & 6.4 & 1.33 \\\hline
		Max depth & The maximal depth obtained by traversing \smalltt{rdfs:subClassOf} axioms in the graph constructed as for the average breadth. The interpretation of max depth is similar to that suggested for ADIT-LN & Cognitive ergonomics & 4 & 6 & 5 & 10 & 4 \\\hline
		Tangledness & The degree of multihierarchical nodes in the class hierarchy computed according to the formula provided by~\cite{Gangemi2006}. A multihierarchical node is a class having multiple super classes. Values for tangledness range from 0 (i.e. no tangledness) to 1 (i.e. each concept in the ontology has multiple super classes) & Cognitive ergonomics & 0.09 & 0.37 & 0.56 & 0.18 & 0.07 \\
	\end{tabular}
	}
	\end{center}
\end{table*}

\begin{figure*}[!ht]
\centering
	\includegraphics[width=1\textwidth]{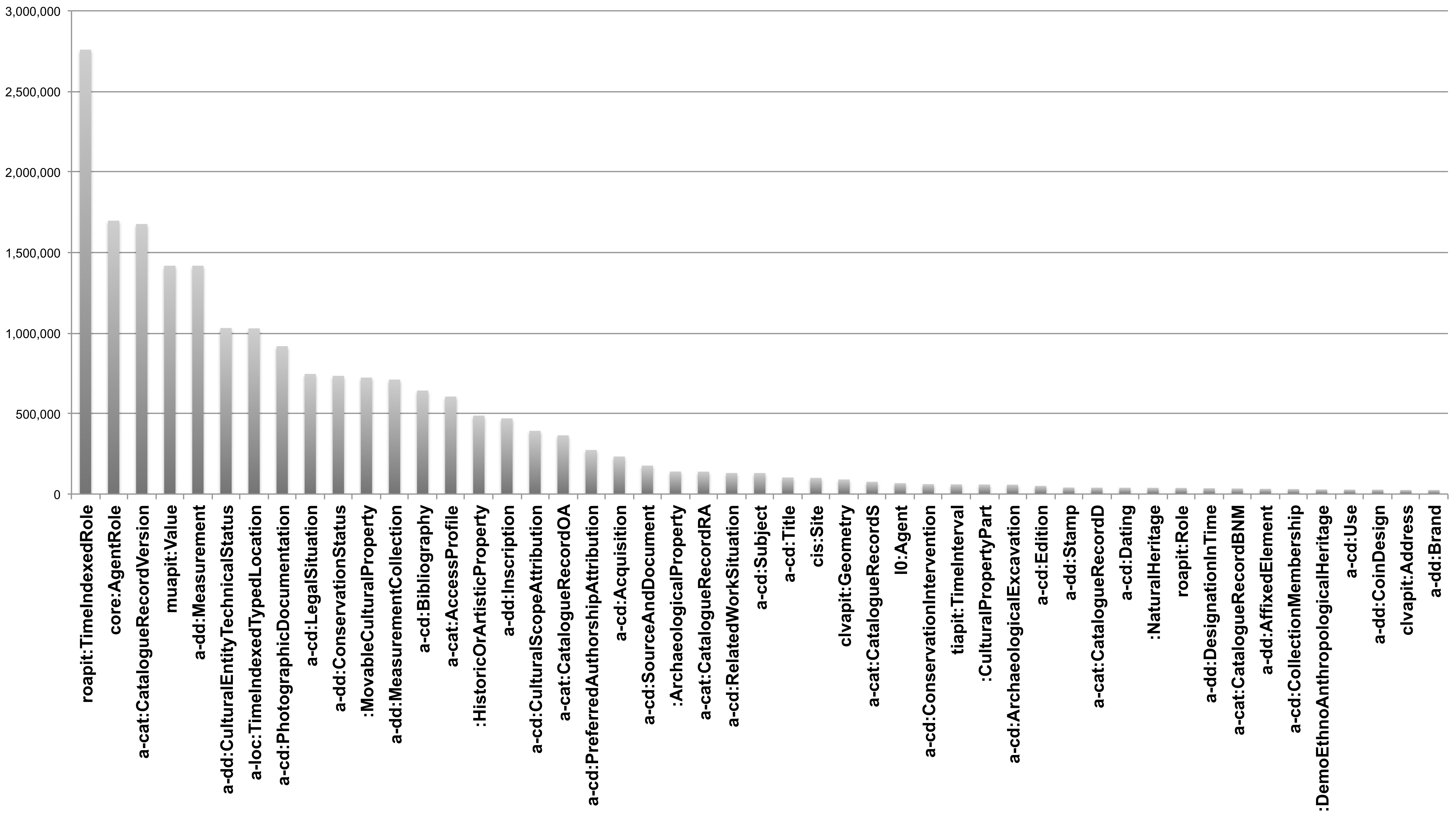}
	\caption{Top-50 ranked classes according to the number of individuals they have in the knowledge graph.}
	\label{img:top50-ranked-classes}
\end{figure*}

We focus on ArCo KG v1.0, which counts 20,030,941 individuals, for analysing the the distribution of those individuals across classes. Such an analysis allows us to understand how individuals are organised in the knowledge graph with respect to concepts. This suggests possible compliance to expertise. In fact, it provides an indication about the recall of classes over the entities of the domain (i.e. the individuals). In this case the recall is meant as extensional coverage computed as the average number of entities captured by ontology classes. It is worth saying that compliance to expertise has a strong functional characterisation that we investigate further by analysing the functional dimension. Notwithstanding, the distribution of the instances across classes is a fair structural metric as it provides us a tool for empirically validating if dense areas (most populated parts of the ontology) correspond to ontology design patterns. The use of patterns is among the indicators suggested by~\cite{Gangemi2006} for measuring the quality properties of transparency and cognitive ergonomics. 
Figure~\ref{img:top50-ranked-classes} shows the top-50 ranked classes based on the number of individuals they have in the knowledge graph. The ranking including all the classes can be retrieved by querying the knowledge graph\footnote{The result set with the ranking of all classes is available at \url{https://bit.ly/2ORiqnM}.}.

A high degree of modularity in an ontology is an indicator of transparency and flexibility. ArCo ontology network is highly modularised, however addressing transparency and flexibility meaningfully requires appropriate design of ontology modules. We compute the following metrics to assess the quality of ArCo modules.

\begin{itemize}
    \item {\em Atomic size}: the average size of a group of interdependent axioms in a module;
    \item {\em Appropriateness of module size}: computed with the Schlicht and Stuckenschmidt function~\cite{Schlicht06} that determines the appropriateness of an ontology module. The appropriateness value ranges from 0 (i.e. no appropriateness) to 1 (i.e. fully appropriateness). According to the Schlicht and Stuckenschmidt function a module size is as much more appropriate as the number of axioms defined in such a module is close to 250;
    \item {\em Encapsulation}: the measure of knowledge preservation within the given module computed as defined by~\cite{Khan16}. Encapsulation values range from 0 (poor encapsulation) to 1 (good encapsulation);
    \item {\em Coupling}: the measure of the degree of interdependence of a module computed as proposed by~\cite{Khan16}. Possible values range from 0 (high interdependence) to 1 (low interdependence).
\end{itemize}

\begin{table*}[!htp]
\begin{center}
	\caption{Results of the module metrics.}
	\label{tab:arco-ml-om}
	\begin{tabular}{p{3cm}||c|c|c|c|c}
	{\bf Ontology Module } & {\bf Atomic size} & {\bf Appropriateness} & 
	{\bf Encapsulation} & {\bf Coupling} \\\hline\hline
	Denotative Description & 6.37 & 1 & 
	0.99 & 0 \\\hline
    Catalogue & 5.64 & 1 & 
    0.99 & 0 \\\hline
    Context Description & 6.63 & 1 & 
    1 & 0 \\\hline
    Core & 4.85 & 1 & 
    0.96 & 0 \\\hline
    Cultural Event & 5.78 & 1 & 
    0.96 & 0 \\\hline
    Location & 5.50 & 1 & 
    0.99 & 0 \\
	\end{tabular}
	\end{center}
\end{table*}

Table~\ref{tab:arco-ml-om} reports the values recorded for the aforementioned module metrics computed for each ontology module of ArCo. Module metrics are obtained by using the Tool for Ontology Module Metrics\footnote{The specific version of the tool we used can be downloaded from \url{https://bit.ly/2nSS2yD}.} (TOMM)~\cite{Khan16}. We do not report module metrics for CIDOC CRM and EDM as they are monolithic ontologies, thus they do not have any modules to assess.

\subsection{Discussion}
In the context of ArCo's project, performing the testing activities initially resulted in a significant manual effort, for both annotating and running the unit tests. For this reason, TESTaLOD has been designed and implemented. The successful execution of inference verification, error provocation, and competency question verification is an indicator of (i) computational integrity and efficiency, and (ii) compliance to expertise. The former suggests that the ontology can be successful processed by a reasoner. The latter suggests that ArCo KG is compliant with its collected requirements. Finally, the terminological coverage measured for ArCo (i.e. 0.72) shows very good results. In fact, the comparison with the results obtained for the Europeana Data Model (EDM) (i.e. 0.07) and  for CIDOC CRM (i.e. 0.2) support the claim that the expressiveness provided by such existing reference ontologies is not completely suitable for addressing ArCo's requirements.

The analysis of the structural dimension shows that ArCo KG provides a larger terminological component than CIDOC CRM and EDM with 3,416 logical axioms, 340 classes, etc. ArCo is a massive knowledge graph counting of 172,580,211 triples describing 20,030,941 individuals. Nevertheless, ArCo is smaller than Europeana, which in turns counts of 2,836,270,332 triples describing 415,410,190 individuals. This finding is fair, first because ArCo is much younger than Europeana. Additionally, we remark that ArCo organises knowledge about Italian cultural properties only; on the contrary, Europeana contains structured knowledge about digital artifacts provided by 28 EU countries. 
Then, if we analyse the indicators obtained, we record that they suggest good transparency. In fact, we record: 
\begin{itemize}
    \item 39.55 axioms per class (i.e. axiom/class ratio), which is similar to that recorded for CIDOC CRM (41.7) and much higher than the number recorded for EDM (7.3); 
    \item an inheritance richness (2.48) comparable to CIDOC CRM (1.17) and EDM (0.32). This is a good indication of how well knowledge is grouped into different categories and subcategories in the ontology. Hence, it suggests a deep (or vertical) ontology, which, in turns, may indicate that the ontology covers a specific domain in a detailed manner; 
    \item a higher NoC (i.e. number of external classes) value for ArCo (38) than for CIDOC CRM (0) and EDM (3). However, this result should be contextualised with respect to the total number of classes (340, 84, and 41 for ArCo, CIDOC CRM, and EDM, respectively). Accordingly, we record 0.1, 0 and 0.07 external classes on average for ArCo, CIDOC CRM and EDM. This indicator suggests, besides transparency, low coupling;
    \item a high degree of relatedness among the different classes, i.e. strong cohesion. In fact, the classes are organised in a hierarchy with (i) a low depth (i.e. ADIT-LN=3.93), (ii) a limited number of root classes if compared to the total number of classes (i.e. NoR=16), and (iii) a high number of leaf classes if compared to the total number of classes (i.e. NoL=277).
\end{itemize}

Low coupling (i.e. NoC) and high cohesion (NoR, NoL, and ADIT-LN) also suggest flexibility, i.e. the property of adapting or changing the ontology with limited side-effects.
The property of cognitive ergonomics (i.e. property of a knowledge graph to be easily understood, manipulated, and exploited by final users) is suggested by:
\begin{itemize}
    \item a lower class/property ratio for ArCo 0.44 (on a scale ranging from 0 to 1) than for CIDOC CRM (0.74) and EDM (0.84);
    \item a low depth and breadth of the inheritance tree (i.e. 3.93 as ADIT-LN, 5 as max depth, 5.75 as average breadth, and 34 as max breadth). According to this indicators ArCo has a similar inheritance tree as EDM. Instead, the inheritance tree of CIDOC CRM is slightly different as it results in higher values for depth and lower for breadth. This means that ArCo has a more compact inheritance tree than CIDOC CRM;
    \item a moderate tangledness (i.e. 0.56 on a scale ranging from 0 to 1) if compared to CIDOC CRM (0.18) and EDM (0.07). This suggests that the inheritance tree is more complex (a number of classes have multiple superclasses, i.e. a multihirarchiy nodes) than that of CIDOC CRM and EDM. We remind that this moderate complexity is only structural and derived from functional requirements. Nevertheless, the high number of annotation axioms (8,374) facilitates user readability. This value is much higher than (i) the total number of classes and properties in the ontology and (ii) that recorded for CIDOC CRM (2,589) and EDM (125);
    \item the use of patterns. With regards to this it is worth noticing that patterns identify dense areas within the knowledge graph. In fact, most of the top-ranked classes among the most instantiated (cf. Figure~\ref{img:top50-ranked-classes}) identifies patterns, such as those described in Section~\ref{sec:odps}. Significant examples are \smalltt{roapit:Time\-Indexed\-Role}, \smalltt{a-cat:Ca\-ta\-lo\-gue\-Re\-cord\-Ver\-sion}, and \smalltt{a-dd:Cul\-tural\-En\-ti\-ty\-Techni\-cal\-Sta\-tus} that count 2,758,760, 1,676,180, and 1,030,566 individuals, respectively.
\end{itemize}

With respect to the evolution of ArCo KG during the design process, we record a significant growth of the knowledge graph from v0.1 to v0.5. For example the number of axioms, classes, and object properties changes from 715, 54, and 38 to 9,564, 329, and 332, for v0.1 and v0.5, respectively. This observation is confirmed by the fact the knowledge graph counts ${\sim}133M$ more triples in v0.5 than in v0.1. On the contrary, we observe a smoothening of the growth curve from v0.5 to v1.0. As a matter of fact, the number of classes varies from 329 to 340 and the knowledge graph increases of ${\sim}3K$ triples from v0.5 to v1.0. This means that: (i) from v0.1 to v0.5 the design process is more focused on the building of the knowledge graph from scratch by means of iterative development cycles; (ii) instead, from v0.5 to v1.0 we observe the change of the design approach towards iterative refinements of the knowledge graph. The refinement activities are fairly evident if we take into account the number of individuals. In fact, such a number, between v0.5 and v1.0, goes down from ${\sim}22.5M$ to ${\sim}20M$. This is due to data cleansing operations aimed, for instance, at collapsing duplicate entries (e.g. a same author associated with multiples IRIs). If we focus on the metrics reported in Table~\ref{tab:arco-sl-om}, we observe comparable values among ArCo v0.1, v0.5, and v1.0. This is an indicator of the fact that the design process has followed an homogeneous strategy imposed by the pattern-based approach. As an example, we record similar among the three versions for relationship richness, inheritance richness, class/property ratio, average breadth, and ADIT-LN.

Module metrics suggest that all modules are modelled by following a similar design principles: identifying small and highly cohesive partitions as basic building blocks for ontology design. This result is fully compliant with the pattern-based approach adopted for modelling ArCo. As a matter of fact, the atomic size values we record are low and they differ only slightly from one module to another, i.e. ranging from 4.85 (core module) to 6.63 (context description module). The appropriateness values recorded are optimal (=1 for all modules). In fact, the appropriateness value for a module ranges from 0 (i.e. no appropriateness) to 1 (i.e. complete appropriateness)~\cite{Khan16}. We record excellent values for encapsulation ($\sim1$ for all modules). We remark that encapsulation values range from 0 (i.e. no encapsulation) to 1 (complete encapsulation). According to ~\cite{dAquin09} a high encapsulation value is a good indication of the quality of a module. In fact, it suggests that such a module can be easily exchanged for another, or internally modified, without side-effects. The extremely low value for coupling (=0 for all modules) is excellent. Again, coupling values range from 0 (i.e. low coupling) to 1 (i.e. high coupling). Low coupling for an ontology module means that its entities do not have strong relations to entities in other modules. Accordingly, it is easy to modify and update such modules independently. Furthermore, the high encapsulation values along with the low coupling values suggest a high degree of independence of a module. This indicates that ArCo modules are self-contained and can be updated and reused separately. Thus, ArCo modules address the flexibility property identified by~\cite{Gangemi2006}, which prospects an ontology/module that can be easily adapted to multiple views.

\section{Developing a KG using XD: lessons learned}
\label{sec:lessons}
This project led us to reflect on both strong and weak points of the methodology applied, thus suggesting possible improvements for the future. In particular, in this section we want to focus on two key aspects of eXtreme Design methodology: (re)using patterns and test-driven design. Finally, we discuss how involving the community let us collect a wider set of requirements.

\subsection{Reusing existing ontologies and patterns}
\label{sec:lesson-reuse}
eXtreme Design is a methodology that encourages the reuse of Ontology Design Patterns (ODPs), as common modelling solutions to classes of problems recurring in ontology design. Patterns to be reused can be both selected from dedicated catalogues (such as the \emph{ODP Portal}\footref{ref:odp-portal}) and extracted from state-of-the-art ontologies. In Section \ref{sec:reuse} we briefly explained the two main practices for ontology reuse: \emph{direct} and \emph{indirect}~\cite{DBLP:conf/er/PresuttiLNGPA16}.

Even if ODP catalogues represent a relevant support for pattern-based ontology design, there is lack of well-documented and well-maintained high-quality ontology design patterns, as well as of tools for supporting ODP-driven ontology-engineering \cite{blomqvist2016considerations}, which could guide the user in the selection of ODPs, e.g. by recommending possible ODPs to be reused for a certain modelling requirement.
Additionally, using available ontologies as input to generate new ontologies is a difficult process, far from being automated \cite{uschold1998ontology, katsumi2016reuse}, and can be hampered by scarsly documented ontologies, ontologies big in size and with a high number of classes, properties and axioms. Moreover, there is a need to carefully (thus time-consuming) consider the context, intended usage and semantic meaning of ontology entities. Issues in reusing existing ontologies seem to be confirmed by \cite{Asprino2019}, which observes a lack of explicit alignments between ontological entities in Linked Open Data, while the high number of top level classes may suggest a high number of conceptual duplicates.

Ontology reuse would benefit from annotations about the ODPs implemented by ontologies: \cite{DBLP:conf/semweb/HitzlerGJKP17} proposes a simple representation language for ontology design patterns (OPLa ontology), which makes use of OWL annotation properties for documenting ODPs. OPLa certainly contributes to fill a gap but its expressiveness requires an improvement. ArCo ontologies have been annotated with OPLa, but we soon realised that we were missing many relevant attributes of, and relations between, patterns that could be annotated and therefore possibly later detected from other parties. 

As described in Section \ref{sec:odps}, during ArCo KG development we incrementally selected a CQ from the available list and then match it with one or more existing ODPs. We also inspected state-of-the-art ontologies, such as CIDOC CRM\footref{ref:cidoc}, EDM\footref{ref:edm}, BIBFRAME\footnote{\label{ref:bibframe}\url{http://id.loc.gov/ontologies/bibframe.html}}, FRBR\footnote{\label{ref:frbr}\url{http://vocab.org/frbr/core}}, etc., in this process. In all cases, this matching activity was incremental and manual, and a significant effort has been made to look for reusable fragments in big ontologies such as CIDOC CRM. 
We believe there is a urgency in developing methods for automatically detecting ODPs used in ontologies as well as in building tools able to provide a modularised ODP-based visualisation of ontologies. These tools would help making the inspection of ontologies clearer and more understandable, hence easing ontology reuse, and contributing in supporting automatic matching procedures. Some work have considered detection of Ontology Design Patterns, e.g. \cite{khan2010ontology} and \cite{lawrynowicz2018discovery}. Nevertheless, to the best of our knowledge, there is no automatic procedure able to recognise ODPs in knowledge graphs nor for annotating and reusing them yet. 
\subsection{Support for test-driven methodologies}
\label{sec:lesson-testing}
Testing an ontology network, which is periodically released in unstable and incremental versions, can be a time-consuming and repetitive activity, and, if performed manually, error-prone. Tests need to be run in order to validate our ontology, by translating competency questions into SPARQL queries, verifying expected inferences and provoking expected errors. Each time there are changes over the ontologies (e.g. a new version which models new information), new tests are created, and all previous tests must be executed again and, if needed, updated, in order to identify new possible bugs.

While performing testing in the context of ArCo KG, we realised that tools automatising it would have been of great support for the testing team. Building TESTaLOD (described in Section \ref{sec:testalod}) helped us executing tests over new versions of the ontology network, allowing for automatic regression tests. At the moment TESTaLOD only addresses CQs-based testing and their corresponding SPARQL queries. Tests for inference verification and error provocation are executed externally. Moreover, the creation and annotation of test cases is not automatised. We believe that developing tools supporting (semi-) automatic creation of unit tests is of paramount importance to push the overall quality of released knowledge graphs. TESTaLOD is just a scratch on the surface of a possible tool suite for automatising many activities of ODP-based and test-driven methodologies such as XD.

\subsection{Extended customer team for Cultural Heritage LOD projects}
\label{sec:lesson-customer}
In ontology engineering methodologies, domain experts are the main actor and input source of requirements and validation tests: they give a crucial contribution, especially in defining domain and task requirements that guide the ontology design and testing phases \cite{Lodi2017}. User stories (then translated into Competency Questions) were used as a \emph{lingua franca} for making communication effective between ontology designers and ICCD domain experts, during the development of ArCo KG.

Whilst not denying the key role played by ICCD domain experts in eliciting requirements, by means of both cataloguing standards, catalogue records and discussions on specific topics and issues, we believe that the Cultural Heritage (CH) domain has a specificity in its users: the community interested in CH data for different purposes is wide and diverse, involving domain experts, researchers, art critics, students, simple citizens, institutions and companies owning and managing CH data or data on related domains (e.g. tourism), public administrations and private companies offering services related to the CH domain, etc.

Cultural Heritage is usually managed with a top-down approach, where professionals and data owners (Galleries, Libraries, Archives, Museums, etc.) are in charge of defining standards and means for describing, representing and making available data on cultural heritage. More rarely, end-users are involved in this process. Instead, institutions aiming at enhancing cultural heritage would benefit from a bottom-up approach, alongside a top-down one, for collecting requirements from the community that consumes their data.

Linked Open Data projects can help in getting domain experts closer to their potential wide and diverse audience, and in promoting interactions between them. In carrying out the ArCo's project, considering the characteristics of the CH domain and CH users, we involved a wider community in the requirements and feedback collection phase. Launching an Early Adoption Program, and involving the community in the unstable and incremental phases of the project, allowed us to capture a wider range of perspectives and requirements.
For example, Synapta\footnote{\url{https://synapta.it/}}, which reuses ArCo ontologies for representing musical instruments belonging to Sound Archives \& Musical Instruments Collection\footnote{\url{http://museopaesaggiosonoro.org/sound-archives-musical-instruments-collection-samic/}} (SAMIC), needed information on musical heritage to be prioritised in the design of the network, while, based on ICCD requirements, this was previously given a lower priority (due to lack of data).

With ArCo's EAP, we experimented the involvement of private and public organisations, and extended XD to this purpose by identifying a set of tools (web forms, mailing lists, GitHub issue tracker), and practices that could support collecting requirements from such a diverse community (webinars, meetups). We believe that collecting requirements from a very diverse community is relevant for the CH domain but can apply also in other contexts, hence methodologies and possible supporting tools shall consider this aspect, so far neglected to the best of our knowledge, among their key requirements. 


%
\section{LOD, ontologies and methodologies in the cultural heritage domain}
\label{sec:related}
The Semantic Web and LOD principles have changed how cultural institutions manage and publish their data, how machines and users can access linked and enriched data on Cultural Heritage (CH), and have widened the possibility of reuse and generation of new knowledge starting from existing data.
Ontologies make it possible to go beyond traditional CH data production and publication, providing users with new, more intelligent and eventually personalised Web applications and services, and with more and richer data~\cite{DBLP:series/ihis/Hyvonen09}.

\subsection{LOD and ontologies for Cultural Heritage}
\label{sec:cidoc-edm}
Projects such as LODLAM\footnote{\url{http://lodlam.net}} and OpenGLAM\footnote{\url{http://openglam.org/}} give evidence of a growing community interested in these themes.
Many cultural institutions are now making cultural properties they safeguard accessible online, by releasing their datasets as Linked Open Data~\cite{DBLP:journals/ao/DijkshoornAOS18}. In the Italian context, notable examples are the Zeri\&Lode\footnote{\url{http://data.fondazionezeri.unibo.it/}} project, which publishes LOD of metadata collections from the Zeri Photo Archive~\cite{DBLP:journals/jocch/DaquinoMPTV17}, and the LOD published by the Institute of Artistic, Cultural and Natural heritage of the Emilia-Romagna region\footnote{\url{https://ibc.regione.emilia-romagna.it/servizi-online/lod/}} (IBC-ER), which include data about libraries, museums, historic castles, butterflies, monumental trees, etc. 
Other noteworthy examples of Linked Data projects include the Rijksmuseum Amsterdam collection\footnote{\url{http://datahub.io/dataset/rijksmuseum}}~\cite{DBLP:journals/semweb/DijkshoornJAOSW18}, the Smithsonian Art Museum\footnote{\url{http://americanart.si.edu/collections/search/lod/about/}}~\cite{DBLP:conf/esws/SzekelyKYZFAG13}, and the German National Library. Effort are also being made in organising knowledge on CH through the publication of controlled vocabularies, as in the case of Getty Vocabularies\footnote{\url{http://www.getty.edu/research/tools/vocabularies/index.html}}, which contain structured terminology for art, architecture, archival and bibliographic materials (e.g. ULAN for artist names, TGN for places relevant to the CH domain, etc.).

Publishing and interconnecting data is leading to the creation of international CH portals \cite{DBLP:series/ihis/Hyvonen09}, such as Europeana\footnote{\url{https://www.europeana.eu/portal/en}}, Google Arts \& Culture\footnote{\url{https://artsandculture.google.com/}}, and MuseumFinland \cite{hyvonen2005museumfinland}, which aggregate content from various publishers into a single site as a point of access of heterogeneous collections; they are referred as \emph{aggregators}.

Along with the publication of LOD collections, ontologies representing the CH domain are being developed, and some of them are becoming widely adopted standards, e.g. the Europeana Data Model\footref{ref:edm} (EDM)~\cite{DBLP:journals/semweb/IsaacH13} and CIDOC Conceptual Reference Model (CRM)\footref{ref:cidoc}~\cite{DBLP:journals/aim/Doerr03}. In addition to them, many other ontologies model specific domains that can be relevant to CH (e.g. the PRO ontology\footnote{\url{http://purl.org/spar/pro/}} for agentive roles).

ArCo substantially contributes to the existing LOD CH cloud with a huge amount of invaluable data of Italian cultural properties, and an ontology network  tackling overlooked modelling issues.


There is quite a variety of knowledge associated with works of art and their dynamics, including their dating, authorship attribution, history, maintenance, symbolic and cultural interpretation, catalog reporting, etc. The design of CH knowledge dynamics requires then sufficient flexibility, which ArCo gathers from its constructive stance, described in Sections \ref{cdns} and \ref{epist}. 

We provide here some cases, where ArCo supplies modelling solutions that are not easily obtained in other widely adopted CH ontologies (see also some related foundational differences in Section \ref{other-event-notions}).

\begin{figure}[!ht]
\centering
	\includegraphics[scale=0.15]{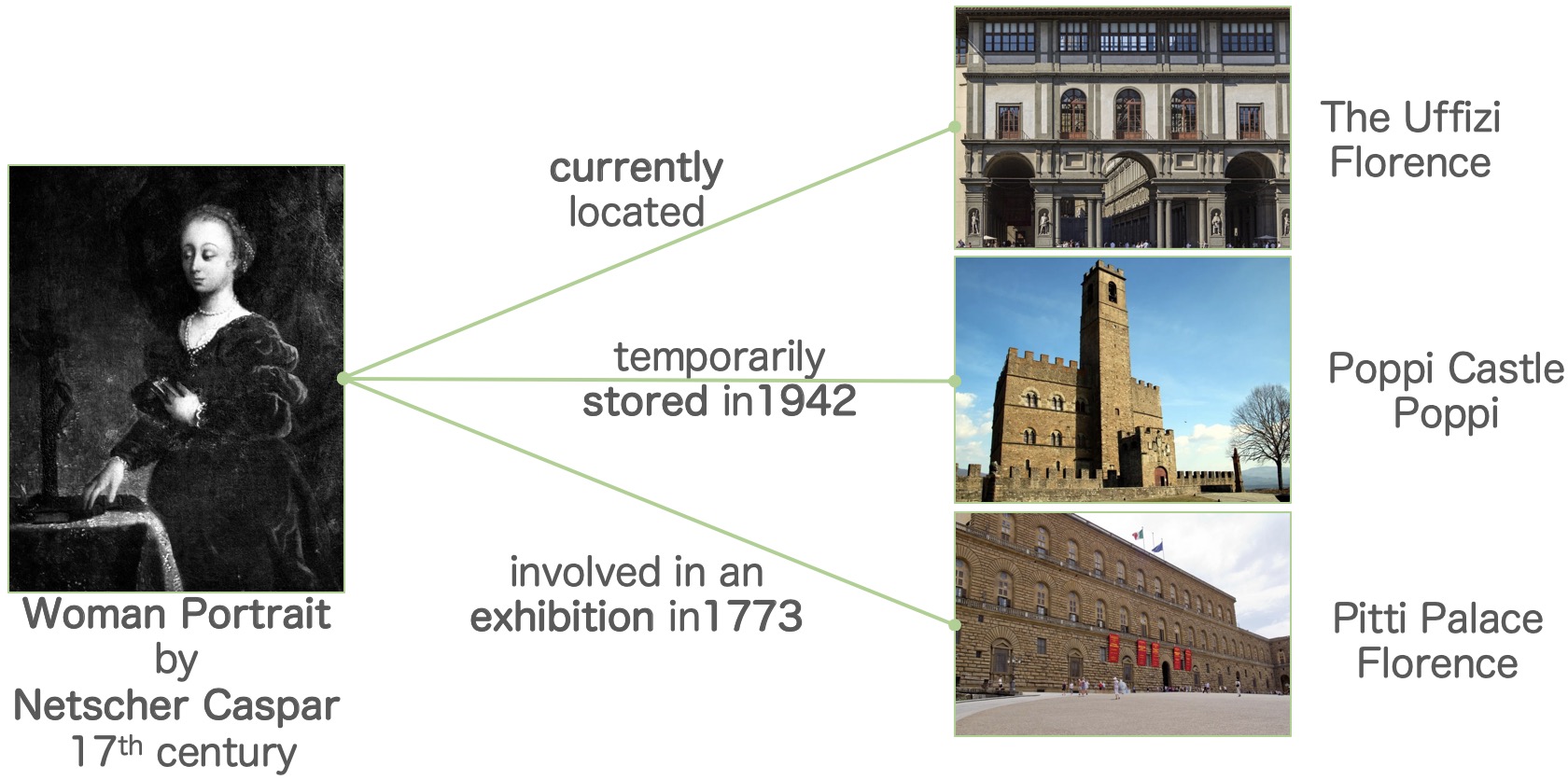}
	\caption{The painting ``Woman Portrait'' by Caspar Netscher (17th century) and the different (types of) locations it is associated with.}
	\label{img:woman}
\end{figure}

As an example, the painting ``Woman Portrait'' by Caspar Netscher (17th century) is associated with several types of locations: it is now located at the Uffizi in Florence, it was stored in 1942 at Poppi Castle, it was involved (hence temporarily moved) in an exhibition at Pitti Palace in Florence in 1773 (cf. Figure~\ref{img:woman}). 

CIDOC CRM allows us to encode the data about all these locations by means of a ``move'' event that is both temporally and geographically indexed. This representation lacks the means to express (i) the knowledge about the type or motivation of a specific location, e.g. production, exhibition, storage, etc.; (ii) the temporal validity of that location, which is different from the time at which a moving event occurs; (iii) in addition, location types can be further characterised by other data or entities specific to them. 
In ArCo it is possible to represent the moving event with its temporal and spatial indexing, as well as the (functional type) of its locations with their own temporal validity, and their specificities. 
We also remark that CIDOC CRM events are defined as \textit{state changes}, hence they should not be applicable to generalised situations or eventualities. However, CIDOC CRM \textit{condition states}, which should cover the rest of phenomena (cf. Section \ref{other-event-notions}), are not supposed to have participants.

While in principle one could possibly extend CIDOC CRM to supply missing modelling solutions, in ArCo we stress the importance of having an explicit constructive stance (Section \ref{cdns}) providing patterns for joint modelling of events, situations, interpretations, etc.

Other examples of ArCo design patterns motivated by its constructive stance include: the modelling of catalogue records as entities of the CH domain representing \textit{interpretative} perspectives on cultural properties, the recurrence of cultural events such as periodic festivals, as well as other situations in which cultural properties are involved (observations, issuances, etc).

\paragraph{\bf Denotation, cataloguing, interpretation.}
When modelling the CH domain, there are at least three levels of representation that might need to be investigated and modelled: (i) the cultural property as such, with its inherent attributes, (ii) the process of systematically gathering and recording pieces of information related to the cultural property, (iii) the process of interpreting these pieces of information in order to extract knowledge about the cultural property.
These levels are separate, but interrelated. Cataloguing is a preliminary step for being able to get the attributes of a cultural property into the universe of discourse: for example, a tangible cultural property is made of a certain material, but without observing it, and collecting and recording data about the material of that specific cultural property, this information may be lost: this activity is essential for preserving cultural heritage. Moreover, cataloguing supports interpretations over cultural properties and their history, and interpretations can be in turn recorded by cataloguers.

We will call the first of these levels \emph{denotation}, as the set of attributes of a cultural property that refer specifically and ``literally'' to that cultural property, without considering its context and the associated meanings that these attributes can have. The second level, \emph{cataloguing}, is based on the observation and recording of the directly observable attributes of the cultural property, while the \emph{interpretation level} discovers new knowledge on the cultural property by interpreting the meaning of one or more attributes within the context, e.g. the observable style of an artwork, or a bibliographic resource about it, can reveal its author.

ArCo distinguishes these three levels (cf. Section \ref{epist}), by providing models for: the denotative representation of a cultural property (materials, conservation status, inscriptions, measurements, etc.); the process of cataloguing (catalogue records as documents describing cultural properties, cataloguing agents, etc.); the interpretation process, with agents involved and criteria guiding it (\smalltt{a-cd:\-Interpretation}).

Both EDM and CIDOC CRM distinguish between the cultural object and possible information resources related to it: \smalltt{crm:\-E89\_Propositional\_\-Object}\footnote{\smalltt{crm:} \url{http://www.cidoc-crm.org/cidoc-crm/}}, to which \smalltt{edm:\-Information\-Resource}\footnote{\smalltt{edm:} \url{http://www.europeana.eu/schemas/edm/}} is aligned, is an item with a set of propositions about real things, and is specialised by e.g. \smalltt{crm:\-E31\_Document}, which documents an entity. Moreover, Europeana represents the set (\smalltt{edm:\-Europeana\-Aggregation}) of resources related to a cultural property, e.g. visual items that are representation of it. However, the concept of cataloguing, fundamental when talking about documentation and preservation activities involving a cultural property, is not explicitly modelled. 
Moreover, it is not possible, by using EDM or CIDOC CRM, to represent how attributes and pieces of information about a cultural property have been interpreted by different agents (e.g. researchers), and which new pieces of information have been generated.


\paragraph{\bf High-level vs. specific.}
ArCo's requirements are primarily based on the ICCD cataloguing standards: as explained in Section \ref{sec:catalogue}, catalogue records can describe 30 different types of cultural properties, each one with distinguishing features, not shared with other types. 
Specificity is therefore a key characteristic of ArCo ontologies, which provide cultural property-specific models to be reused, extending by specialisation the taxonomy offered by EDM and CIDOC CRM. 
As explained in~\cite{DBLP:journals/ao/DijkshoornAOS18}, EDM has been developed for integrating and making interoperable various metadata standards from a multitude of Galleries, Libraries, Archives and Museums (GLAM) across Europe, as a \enquote{common denominator} model to use for the Europeana portal. 
Therefore, it is by design that EDM does not invest into a fine level of granularity or rich axiomatisation. It provides a basic set of classes and properties, which also include many Dublin Core\footnote{\url{http://dublincore.org}} constructs, hence its axiomatic detail is underspecified. 
CIDOC CRM is a richer model than EDM, but remains a high-level model, which needs to be specialised for specific types of cultural entities.

For example, a musical instrument would be of type \smalltt{edm:\-Physical\-Thing}, \smalltt{edm:ProvidedCHO} and \smalltt{crm:\-E22\_Man\_Made\_Object} by using without extensions EDM and CIDOC CRM, and of type \smalltt{:Music\-Heritage} by using ArCo. The same need to specialise concepts can be found in other projects reusing and extending CIDOC, e.g. Zeri\&Lode, where FEntry ontology\footnote{\label{ref:fentry}\smalltt{fentry:} \url{https://w3id.org/people/essepuntato/fentry}} has been developed for modelling concepts related to photographs, and DOREMUS\footnote{\url{https://www.doremus.org/}}, which extends many classes and properties for musical data.


Specific types of cultural properties share some features (e.g. location, dating, author), but ArCo also needs to satisfy more modelling issues, overlooked by other ontologies so far, related to specific cultural properties, such as the diagnosis of a paleopathology and the interpretation of sex and age of death in anthropological material, other types of surveys on archaeological objects (e.g. laboratory tests), the coin issuance, the Hornbostel-Sachs classification of musical instruments, musicians and musical ensemble, recurrent art exhibitions, etc.

For example, let us take a modelling problem that a cultural institution publishing data on anthropological materials may need to address: it has been estimated that discovered anthropological materials (teeth, mandible and radius) are from a female  individual dead at young age.
The assignments of the attributes ``female'' and ``young age'' would be modelled in CIDOC as \smalltt{crm:E13\_Attribute\_Assignment}, where the \smalltt{crm:E1\_CRM\_Entity} ``female'' can be typed (\smalltt{crm:\-P2\_has\_type}) as sex and ``young age'' as age of death. There is no means to represent that the sex is attributed based on the size and thickness of radius, and the age of death on the basis of the light dental wear. In ArCo, specific classes and properties, with an explicit semantics, have been created for solving this modelling issue: \smalltt{a-cd:Sex\-Interpreta\-tion} and \smalltt{a-cd:Age\-Of\-Death\-Interpreta\-tion}, that can be related to the criterion that motivated that interpretation \smalltt{a-cd:\-Interpreta\-tion\-Criterion}. Axioms on the class \smalltt{:Ar\-chae\-o\-log\-i\-cal\-Pro\-per\-ty} allow to specify that this type of cultural property can have sex and age of death interpretations. 

Even in the case of features shared by most cultural properties, in many cases CIDOC lacks the expressiveness needed for modelling ArCo data without missing information. For example, according to CIDOC, changes of the physical location of a cultural property are represented by move events, and we can only know from and to where the cultural property was moved, and when the move happened, while there is no means to express e.g. the role that a specific location played during a time interval, with respect to a specific cultural property. EDM specialises \smalltt{dct:spatial}\footnote{\smalltt{dct:} \url{http://purl.org/dc/terms/}} for representing the current location, while CIDOC distinguishes, by means of relations between a cultural property and a place, 3 types of locations: current, current or former, current permanent. In order to satisfy our requirements in representing physical locations associated with a cultural property, we need more expressiveness, e.g. for distinguishing between different types of locations with a temporal validity: the place were a cultural property was exhibited, the place where it was found, etc.

\paragraph{\bf Alignments.} ArCo is aligned to both EDM and CIDOC CRM\footnote{For each module of ArCo, a separate file stores its alignments, e.g. \url{https://w3id.org/arco/ontology/context-description-aligns/}.}. Few ontology entities have been aligned to EDM, since (i) EDM reuses many external properties e.g. from Dublin Core, (ii) EDM models some top-level concepts (e.g. \smalltt{edm:\-Place}, \smalltt{edm:\-TimeSpan}) that ArCo reuses from other top-level models (see Subsection \ref{sec:reuse} for more details), (iii) many EDM concepts (e.g. \smalltt{edm:\-Europeana\-Aggregation}) and relations (e.g. \smalltt{edm:\-aggregated\-CHO}) are explicitly related to Europeana as an aggregator. When possible, patterns have been aligned to CIDOC CRM, e.g. the ones for modelling acquisition, copyright, previous and current owners, conservation status. In most cases, ArCo specialises CIDOC's concepts, as in the case of the subclasses of \smalltt{a-cd:\-Affixed\-Element} (for modelling e.g. coat of arms, emblems, coin legend, logo), which is aligned to \smalltt{crm:\-E37\_Mark}.

Other alignments include: BIBFRAME\footref{ref:bibframe}, FRBR\footref{ref:frbr}, FaBiO\footnote{\label{ref:fabio}\url{http://purl.org/spar/fabio}} (for bibliographic data), FEntry\footref{ref:fentry} and OAEntry\footnote{\label{ref:oaentry}\url{http://purl.org/emmedi/oaentry}} (dedicated to photographs and artworks). 
More alignments are planned as ArCo evolves. The richness and high level of detail of ArCo requirements though led us to perform a consistent modelling effort and to release to the community a number of useful ontology patterns for representing the CH domain, which integrates existing ontologies modelling cultural heritage.

\subsection{Methodologies for CH LOD modelling and publishing} 
As discussed in~\cite{DBLP:journals/ao/DijkshoornAOS18}, when building a knowledge graph (KG) for publishing its data, a cultural institution makes a first relevant choice: it can publish Linked Open Data by building and using its own infrastructure, give its data to a cultural heritage data aggregator such as Europeana, or invest in infrastructure for publishing its data as well as in the whole process for producing them, by using the ontology model of an aggregator. 

In making this choice, a cultural heritage administrator is influenced by different aspects both political, economical and technical. An aggregator provides a single point of access to different collections from many cultural institutions, giving visibility and guaranteeing respective enrichment and interoperability. Nevertheless, the adopted ontologies only capture a subset and a simplified encoding of the available information about a cultural property because they prefer a \emph{lightweight} modelling i.e. based on binary relations, as opposed to more complex predicates, e.g. \emph{n}-ary relations. Many existing CH institutions provide data to Europeana and/or use Europeana Data Model, along with Dublin Core, for representing their collections, such as the Rijksmuseum dataset~\cite{DBLP:journals/semweb/DijkshoornJAOSW18}, possibly extending it, as in the case of the VVV ontology~\cite{dragoni2016enriching}. In each of these projects, the institution intentionally chooses to publish only a subset of all the features characterising cultural properties, which are instead present in the original dataset, in order to reuse EDM and avoid a more significant effort in mapping between the input data and the ontology model. In our opinion, when possible, it is preferable for an institution to carry out the whole process of data production and publication and to release as much rich data as possible, while guaranteeing the interoperability with and the publication (of simplified or subsets of its data) through aggregators: this is the approach followed by ArCo. This choice allows a cultural institution to clearly define its requirements regardless of which data is possible to publish through an aggregator, and by using which ontology. Moreover, such an approach better supports an open requirements collection, not limiting the commitment of the developed ontologies to the institutional guidelines. Having full control on the ontological commitment minimises loss of information contained in input data, for reaching a wide audience of diverse users.



%
\section{Conclusion}
\label{sec:conclusion}
This paper presents how ArCo, a knowledge graph of Italian Cultural Heritage (CH), has been designed, following the principles of the XD methodology. There are other valuable LOD resources containing and describing the Italian CH. Nevertheless, ArCo KG has a prominent role in this domain, not only because it injects in LOD a huge amount of high-quality data, extracted from the official institutional database of Italian Cultural Heritage (General Catalogue), but also because the expressiveness of its ontologies facilitates the adoption of its LOD by scholars and researchers in humanities and beyond, to make discoveries and find new patterns. The expected impact of ArCo KG on the general CH domain is motivated by a set of new requirements, addressed by its ontologies, which have been overlooked so far. These requirements emerged both from the richness of details provided by the General Catalogue records as well as from a growing community of consumers and producers of CH LOD.

ArCo KG can have an impact on the general Semantic Web community as well, since it is designed by following a robust methodology, based on the reuse of ontology design patterns, including extensive testing, detailed documentation and tutorial material, and formal evaluation: thus, it is a well-documented case study of the application of a methodology of ontology engineering (eXtreme Design), and can be used as a reference example by other researchers that are approaching knowledge graph engineering.

ArCo KG is still evolving and growing, and can be further improved and enriched. We plan to extend our ontologies, in order to model other aspects not addressed by the current version, e.g. some specific characteristics of naturalistic heritage, like slides and phials associated to an \emph{herbarium}, the optical properties of a stone, etc. Being ArCo KG developed in the context of an evolving project, we keep encouraging our community to give us new requirements, in addition to continuous feedback, that we aim at addressing in the future. Moreover, in future requirement collection iterations, we want to extend our customer team to interested citizens, and further investigate how to best capture requirements from such a diverse audience.

ArCo KG will be enriched by extracting structured data from many textual metadata contained in the catalogue records (e.g. generic narrative descriptions of the cultural properties, historical biographical data about authors, etc.), using NLP techniques.
Additional effort is being put to complete the translation of the data to other languages, starting from English, with an automated bootstrap to be refined by the community. Finally, ArCo's project has highlighted the need for tools for facilitating reuse and testing, in general but also specific to the CH domain.
%





\bibliographystyle{ios1}           
\bibliography{bibliography}        

%

\end{document}